\documentclass{article}

\PassOptionsToPackage{numbers, sort&compress}{natbib}

\usepackage[preprint]{neurips_data_2022}

\usepackage[utf8]{inputenc} 
\usepackage[T1]{fontenc}    
\usepackage{hyperref}       
\usepackage{url}            
\usepackage{booktabs}       
\usepackage{amsfonts}       
\usepackage{nicefrac}       
\usepackage{microtype}      
\usepackage{xcolor}         

\usepackage{times}
\usepackage{soul}
\usepackage{url}
\usepackage[utf8]{inputenc}
\usepackage[small]{caption}
\usepackage{graphicx}
\usepackage{amsmath}
\usepackage{amsthm}
\usepackage{booktabs}
\usepackage{algorithm}
\usepackage{algorithmic}
\usepackage{cite}
\usepackage{natbib}
\usepackage{subfig}
\usepackage{float}
\usepackage{amsfonts}
\usepackage{enumitem}
\usepackage{threeparttable}
\usepackage{booktabs}
\usepackage{diagbox}
\usepackage{multirow}
\usepackage{makecell}
\usepackage{xcolor}

\newcommand{\D}{{\mathcal D}}

\newcommand{\bx}{\mathbf{x}}

\newcommand{\eg}{\textit{e.g.}}
\newcommand{\ie}{\textit{i.e.}}

\newcommand{\zxy}[1]{\textcolor{black}{#1}}

\newcommand{\topa}[1]{\textcolor{red}{#1}}
\newcommand{\topb}[1]{\textcolor{teal}{#1}}
\newcommand{\topc}[1]{\textcolor{blue}{#1}}
\title{A Comparative Survey of Deep Active Learning}

\author{
Xueying Zhan\thanks{Work was completed while the first author was at Baidu Research.} \\
City University of Hong Kong\\
\texttt{\fontsize{7.9pt}{\baselineskip}\selectfont{xyzhan2-c@my.cityu.edu.hk} }\\
\And
Qingzhong Wang \\
Baidu Research\\
\texttt{\fontsize{7.9pt}{\baselineskip}\selectfont{wangqingzhong@baidu.com}} \\
\And
Kuan-Hao Huang \\
University of California \\
\texttt{\fontsize{7.9pt}{\baselineskip}\selectfont{khhuang@cs.ucla.edu}}\\
\AND
Haoyi Xiong \\
Baidu Research \\
\texttt{\fontsize{7.8pt}{\baselineskip}\selectfont{xionghaoyi@baidu.com}}\\
\And
Dejing Dou \\
Baidu Research \\
\texttt{\fontsize{7.8pt}{\baselineskip}\selectfont{doudejing@baidu.com}}\\
\And
Antoni B. Chan \\
City University of Hong Kong \\
\texttt{\fontsize{7.8pt}{\baselineskip}\selectfont{abchan@cityu.edu.hk}} \\
}

\begin{document}

\maketitle
\begin{abstract}
While deep learning (DL) is data-hungry and usually relies on extensive labeled data to deliver good performance, Active Learning (AL) reduces labeling costs by selecting a small proportion of samples from unlabeled data for labeling and training. Therefore, Deep Active Learning (DAL) has risen as a feasible solution for maximizing model performance under a limited labeling cost/budget in recent years. Although abundant methods of DAL have been developed and various literature reviews conducted, the performance evaluation of DAL methods under fair comparison settings is not yet available. Our work intends to fill this gap.
In this work, We construct a DAL toolkit, \emph{$\text{DeepAL}^+$}, by re-implementing 19 highly-cited DAL methods. We survey and categorize DAL-related works and construct comparative experiments across frequently used datasets and DAL algorithms. Additionally, we explore some factors (e.g., batch size, number of epochs in the training process) that influence the efficacy of DAL, which provides better references for researchers to design their DAL experiments or carry out DAL-related applications. 
\end{abstract}
\section{Introduction}
Blessed by the capacity of representation learning in an over-parameterized architecture, Deep Neural Networks (DNNs) have been used as significant workhorses in various machine learning (ML) tasks.
While DNNs can work with extensive training datasets and deliver decent performance, collecting and annotating data to feed DNNs training becomes extremely expensive and time-consuming.
On the other hand, given a large pool of unlabeled data, AL improves learning efficiency by selecting small subsets of samples for annotating and training  \citep{xie2021towards}. In this way, a sweet spot appears at the intersection of DNNs and AL, where representation learning can be achieved with reduced labeling costs.
Deep Active Learning (DAL) has been employed in various tasks, e.g., named entity recognition \citep{chen2015study,shen2017deep}, semantic parsing \citep{duong2018active}, object detection \citep{roy2018deep, haussmann2020scalable}, image segmentation \citep{casanova2020reinforced,saidu2021active}, counting \citep{zhao2020active}, etc. Besides these applications,  
 multiple unified DAL frameworks have been designed and perform well on various tasks  \citep{sener2017active,ash2019deep,pinsler2019bayesian,shui2020deep}.

DAL originated from AL for classical ML tasks, which has been well studied in past years. The application of AL to classical ML tasks appear in a wealth of literature surveys \citep{settles2009active, wang2011active, fu2013survey, aggarwal2014active, elahi2016survey, kumar2020active, wu2020multi} and comparative studies \citep{korner2006multi, schein2007active, settles2008analysis, tuia2011survey, cawley2011baseline, sivaraman2014active, ramirez2017active, pereira2019empirical, naseem2021comparative, zhan2021comp}. Some traditional AL methods for classical ML have been generalized to DL tasks \citep{wang2014new, gal2017deep, beluch2018power}. Adapting AL methods to work well on classical ML tasks has several issues to overcome \citep{ren2021survey}: 1) different from traditional AL methods that use fixed pre-processed features to calculate uncertainty/representativeness, in DL tasks, feature representations are jointly learned with DNNs. Therefore, feature representations are dynamically changing during DAL processes, and 
thus pairwise distances/similarities used by representativeness-based measures need to be re-computed in every stage, 
whereas for AL with classical ML tasks, these pairwise terms can be pre-computed.
2) DNNs are typically over-confident with their predictions and thus evaluating the uncertainty of unlabeled data might be unreliable.
\citet{ren2021survey} conducted a comprehensive review of DAL, which systematically summarizes and categorizes $189$ existing works. Indeed it is a comprehensive study of DAL and guides new and experienced researchers who want to use it. However, due to the lack of experimental comparisons among various branches of DAL algorithms across different datasets/tasks, it is difficult for researchers to distinguish which DAL algorithms are suitable for which task. Our work aims to fill this gap. 

In this work, we construct a DAL toolkit, called \emph{$\text{DeepAL}^+$}, by re-implementing 19 DAL methods surveyed in this paper. \emph{$\text{DeepAL}^+$} is sequel to our previous work \emph{DeepAL} \citep{huang2021deepal}. Compared to \emph{DeepAL}, which includes \zxy{11} highly-cited DAL approaches prior to 2018, in \emph{$\text{DeepAL}^+$}, 1) we upgraded and optimized some algorithms that already were implemented in \emph{DeepAL}; 2) we re-implemented more highly-cited DAL algorithms, most of which are proposed after 2018; 3) besides well-studied datasets adopted in \emph{DeepAL} like \emph{MNIST} \citep{deng2012mnist}, \emph{CIFAR} \citep{krizhevsky2009learning}  and \emph{SVHN} \citep{netzer2011reading}, we integrated more complicated tasks in \emph{$\text{DeepAL}^+$} like medical image analysis \citep{spanhol2015dataset, kermany2018identifying} and object recognition with correlated backgrounds (containing spurious correlations) \citep{sagawa2019distributionally}.
We conduct comparative experiments between a variety of DAL approaches based on \emph{$\text{DeepAL}^+$} on multiple tasks and also explore factors of interest to researchers, such as the influence of batch size and the number of training epochs in each AL iteration, and timing-cost comparison. More descriptions of \emph{$\text{DeepAL}^+$} are in Section B in Appendix.

We hope that our comparative study/benchmarking test brings authentic comparative evaluation for DAL, provides a quick look at which DAL models are more effective and what are the challenges and possible research directions in DAL, as well as offering guidelines for conducting fair comparative experiments for future DAL methods. More importantly, we expect that our \emph{$\text{DeepAL}^+$} 
will contribute to the development of DAL since \emph{$\text{DeepAL}^+$} is extensible, allowing easy incorporation of new basic tasks/datasets, new DAL algorithms, and new basic learned models. This makes the application of DAL to downstream tasks, and designing new DAL algorithms becomes easier. \emph{$\text{DeepAL}^+$} is an ongoing process. We will keep expanding it by incorporating more basic tasks, models, and DAL algorithms. Our \emph{$\text{DeepAL}^+$} is available on \url{https://github.com/SineZHAN/deepALplus}.

\section{DAL Approaches}
This section provides an overview of highly-cited DAL methods in recent years, including the perspectives of querying strategies and techniques for enhancing DAL methods.
\paragraph{Problem Definition.}
We only discuss pool-based AL, since most DAL approaches belong to this category. Pool-based AL selects most informative data iteratively from a large pool of unlabeled $i.i.d.$ data samples until either the basic learner(s) reaches a certain level of performance or a fixed budget is exhausted \citep{chu2011unbiased}. We consider a general process of DAL, taking classification tasks as example, where other tasks (\eg, image segmentation) follow the common definition of their tasks domain. We have an initial labeled set $\D_{l}=\{(\bx_j,y_j)\}_{j=1}^M$ and a large unlabeled data pool $\D_{u}=\{\bx_i,\}_{i=1}^N$, where $M \ll N$,  $y_i \in \{0,1\}$ is the class label of $\bx_i$ for binary classification, or $y_i \in \{1,...,k\}$ for multi-class classification. In each iteration, we select batch of samples $\D_q$ with batch size $b$ from $\D_u$ based on basic learned model $\mathcal{M}$ and an acquisition function $\alpha(\bx, \mathcal{M})$, and query their labels from the oracle. Data samples are selected by $\D_q^* = \arg\max\nolimits_{\bx \in \D_u}^{b} \alpha(\bx, \mathcal{M})$,
where the superscript $b$ indicates selection of the top $b$ points. $\D_l$ and $\D_u$ are then updated, and $\mathcal{M}$ is retrained on $\D_l$. DAL terminates when the budget $Q$ is exhausted or a desired model performance is reached. 
\subsection{Querying Strategies}
\label{query_strategy}
\begin{figure}[H] 
\centering 
\includegraphics[width=1.0\linewidth]{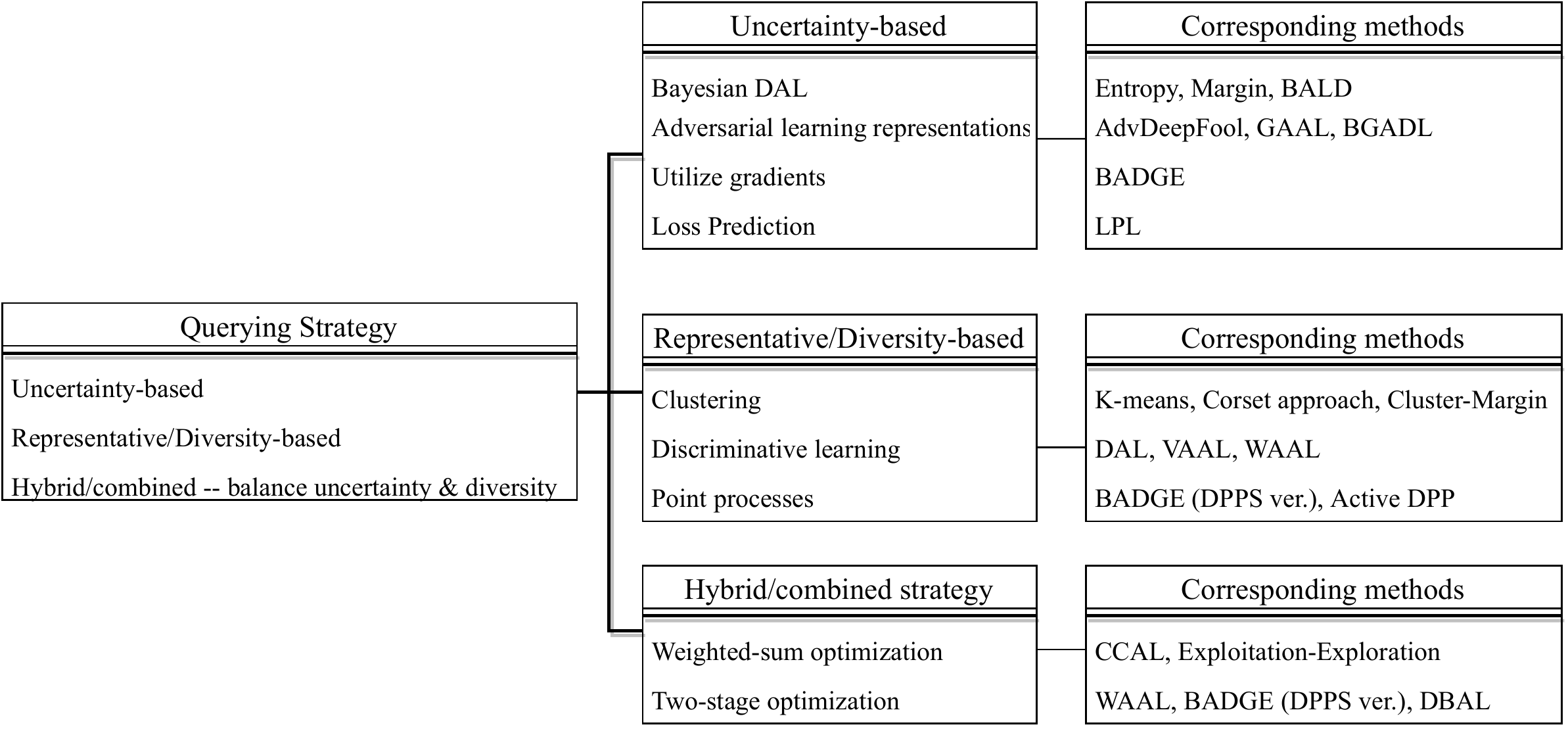} 
\caption{Categorization of DAL sampling/querying strategies.}
\label{category_query_strategy}
\end{figure}

DAL can be categorized into 3 branches from the perspective of querying strategy: uncertainty-based, representativeness/diversity-based and combined strategies, as shown in Figure~\ref{category_query_strategy}.

\subsubsection{Uncertainty-based Querying Strategies} 
Uncertainty-based DAL selects data samples with high aleatoric uncertainty or epistemic uncertainty, where aleatoric uncertainty refers to the natural uncertainty in data due to influences on data generation processes that are inherently random. Epistemic uncertainty comes from the modeling/learning process and is caused by a lack of knowledge \citep{settles2009active,senge2014reliable, nguyen2019epistemic, karamcheti2021mind}. Many uncertainty-based DAL measures are adapted from pool-based AL techniques for classical ML tasks. Typical methods include:
\begin{enumerate}[leftmargin=*]
\item Maximum Entropy (\textbf{Entropy}) \citep{shannon2001mathematical} selects data $\bx$ that maximize the predictive entropy $H_{\mathcal{M}}[y|\bx]$: $\alpha_{\textbf{entropy}} (\bx, \mathcal{M}) = H_{\mathcal{M}}[y|\bx] = -\sum\nolimits_k p_{\mathcal{M}}(y = k|\bx) \log p_{\mathcal{M}}(y = k|\bx)$, 
where $p_{\mathcal{M}}(y|\bx)$ is the posterior label probability from the classifier $\mathcal{M}$.
\item \textbf{Margin} \citep{netzer2011reading} selects data $\bx$ whose two most likely labels $(\hat{y}_1,\hat{y}_2)$ have smallest difference in posterior probabilities: $\alpha_{\textbf{margin}} (\bx, \mathcal{M}) = - [p_{\mathcal{M}}(\hat{y}_1|\bx) - p_{\mathcal{M}}(\hat{y}_2|\bx)]$.
\item Least Confidence (\textbf{LeastConf}) \citep{wang2014new} selects data $\bx$  whose most likely label $\hat{y}$ has lowest posterior probability: $\alpha_{\textbf{LeastConf}} (\bx, \mathcal{M}) = - p_{\mathcal{M}}(\hat{y}|\bx)$.
A similar method is Variation Ratios (\textbf{VarRatio}) \citep{freeman1965elementary}, which measures the lack of confidence like \textbf{LeastConf}: $\alpha_{\textbf{VarRatio}} (\bx, \mathcal{M}) =  1 - p_{\mathcal{M}}(\hat{y}|\bx)$.
\item Bayesian Active Learning by Disagreements (\textbf{BALD}) \citep{houlsby2011bayesian, gal2017deep} chooses data points that are expected to maximize the information gained from the model parameters $\mathbf{\omega}$, i.e. the mutual information between predictions and model posterior: $\alpha_{\textbf{BALD}} (\bx, \mathcal{M}) = H_{\mathcal{M}}[y|\bx] - \mathbb{E}_{p(\mathbf{\omega}|D_l)}[H_{\mathcal{M}}[y|\bx,\mathbf{\omega}]]$.
\item  Mean Standard Deviation (\textbf{MeanSTD}) \citep{kampffmeyer2016semantic} maximizes the mean standard deviation of the predicted probabilities over all $k$ classes: $\alpha_{\textbf{MeanSTD}} (\bx, \mathcal{M}) = \frac{1}{k} \sum\nolimits_k \sqrt{\text{Var}_{q(\omega)}[p(y=k|\bx, \omega)]}$.
\end{enumerate}

Inspired by recent advances in generating adversarial examples,
some DAL methods utilize adversarial attacks to rank the uncertainty/informativeness of each unlabeled data sample.
The DeepFool Active Learning method (\textbf{AdvDeepFool})  \citep{ducoffe2018adversarial} queries the unlabeled samples that are closest to their adversarial attacks (DeepFool). 
 Specifically, $\alpha_{\textbf{AdvDeepFool}}(\bx, \mathcal{M}) = \mathbf{r}_{\bx}$,
 where $\mathbf{r}_{\bx}$ is the minimal perturbation that causes the changing of labels, \eg, for binary classification, $\mathbf{r}_{\bx} = \mathop{\mathrm{argmin}}_{\mathbf{r}, ~ \mathcal{M}(\bx) \neq \mathcal{M}(\bx+\mathbf{r})} -\tfrac{\mathcal{M}(\bx+\mathbf{r})}{||\nabla \mathcal{M}(\bx+\mathbf{r})||^2_2} \nabla \mathcal{M}(\bx+\mathbf{r})$.
DeepFool attack can be replaced by other attack methods, \eg, Basic Interactive Method (BIM) \citep{kurakin2016adversarial}, called \textbf{AdvBIM}. 

Generative Adversarial Active Learning (\textbf{GAAL}) \citep{zhu2017generative}  synthesizes queries via Generative Adversarial Networks (GANs). In contrast to regular AL that selects points from the unlabeled data pool, \textbf{GAAL} generates images from GAN for querying human annotators. 
However, the generated data very close to the classifier decision boundary may be meaningless, and even human annotators could not distinguish its category.
An improved approach called Bayesian Generative Active Deep Learning (\textbf{BGADL}) \citep{tran2019bayesian} combines active learning with data augmentation. \textbf{BGADL} utilizes typical Bayesian DAL approaches for its acquisition function (e.g., $\alpha_{\textbf{BALD}}$) and then trains a VAE-ACGAN to generate synthetic data samples to enlarge the training set. 
Other practical uncertainty-based measures include i) utilizing gradient: 
\citet{wang2021boosting} found that gradient norm can effectively guide unlabeled data selection; that is, selecting unlabeled data of higher gradient norm can reduce the upper bound of the test loss. Another work that utilizes gradient is 
Batch Active learning by Diverse Gradient Embeddings (\textbf{BADGE})\citep{ash2019deep} measures uncertainty as the gradient magnitude with respect to parameters in the output layer since DNNs are optimized using gradient-based methods like SGD. ii) Loss Prediction Loss  (\textbf{LPL}) \citep{yoo2019learning} uses a loss prediction strategy by attaching a small parametric module that is trained to predict the loss of unlabeled inputs with respect to the target model, by minimizing the loss prediction loss between predicted loss and target loss. 
\textbf{LPL} picks the top $b$ data samples with the highest predicted loss.

\subsubsection{Representative/Diversity-based Querying Strategies} 
Representative/diversity-based strategies select batches of samples representative of the unlabeled set and are based on the intuition that the selected representative examples, once labeled, can act as a surrogate for the entire dataset \citep{ash2019deep}. 
Clustering methods are widely used in representative-based strategies. A typical method is \textbf{KMeans}, which selects centroids by iteratively sampling points in proportion to their squared distances from the nearest previously selected centroid. Another widely adopted approach \citep{geifman2017deep, sener2017active} selects a batch of representative points based on a core set, which is a sub-sample of a dataset that can be used as a proxy for the full set. \textbf{CoreSet} is measured in the penultimate layer space $h(\bx)$  of the current model. Firstly, given $\D_l$, an example $\bx_u$ is selected with the greatest distance to its nearest neighbor in the hidden space $u = \arg\max\nolimits_{\bx_i \in \D_u} \min\nolimits_{\bx_j \in \D_l} \Delta(h(\bx_i, \bx_j))$. Sampling is then repeated until batch size $b$ is reached. In another method,  \textbf{Cluster-Margin} \citep{citovsky2021batch} selects a diverse set of examples on which the model is least confident. It first runs hierarchical agglomerate clustering with average-linkage as pre-processing, and then selects an unlabeled subset with lowest margin scores (\textbf{Margin}), which is then filtered down to a diverse set with $b$ samples. In contrast to \textbf{CoreSet}, \textbf{Cluster-Margin} only runs clustering once as pre-processing.

Point Processes are also adopted in representative-based DAL, \eg, 
\textbf{Active-DPP} \citep{biyik2019batch}.
A determinantal point process (DPP) captures diversity by constructing a pair-wise (dis)similarity matrix and calculating its determinant.
\textbf{BADGE} also utilizes DPPs as a representative measure.
 Discriminative AL (\textbf{DiscAL}) \citep{gissin2019discriminative} is a representative measure 
that, reminiscent of GANs, attempts to fool a discriminator that tries to distinguish between data coming from two different distributions (unlabeled/labeled). 
 Variational Adversarial AL (\textbf{VAAL}) \citep{sinha2019variational} learns a distribution of labeled data in latent space using a VAE and an adversarial network trained to discriminate between unlabeled and labeled data. The network is optimized using both reconstruction and adversarial losses. 
 $\alpha_{\textbf{VAAL}}$ is formed with the discriminator that estimates the probability that the data comes from the unlabeled data.
Wasserstein Adversarial AL (\textbf{WAAL}) \citep{shui2020deep} searches the diverse unlabeled batch that also has larger diversity than the labeled samples through adversarial training by $\mathcal{H}$-divergence. 
\subsubsection{Combined Querying Strategies}
Due to the demand for larger batch size (representative/diversity) and more precise decision boundaries for higher model performance (uncertainty) in DAL, combined strategies have become the dominant approaches to DAL. It aims to achieve a trade-off between uncertainty and representativeness/diversity in query selection. We mainly discuss the optimization methods with respect to multiple objectives (uncertainty, diversity, etc.) in this paper, including weighted-sum and two-stage optimization.

Weighted-sum optimization is both simple and flexible, where the objective functions are summed up with weight $\beta$: $\alpha_{\textbf{weighted-sum}} = \alpha_{\textbf{uncertainty}} + \beta \alpha_{\textbf{representative}}$. However, two factors limit its usage in Combined DAL: 1) it introduces extra hyper-parameter $\beta$ for tuning; 2) unlike uncertainty-based measure that provide a single score per sample, representativeness is usually expressed in matrix form, which is not straightforward to convert into a single per-sample score.
A example of weighted-sum optimization is \textbf{Exploitation-Exploration} \citep{yin2017deep} selects samples that are most uncertain and least redundant (exploitation), as well as most diverse (exploration). Specifically, in the exploitation step, $\alpha_{\textbf{exploitation}} = \alpha_{\textbf{entropy}}(\D_q, \mathcal{M}) - \tfrac{\beta}{|\D_q|}\alpha_{\textbf{similarity}}(\D_q)$.
Using DPPs is a natural way to balance uncertainty score and pairwise diversity well without introducing additional hyper-parameters \citep{biyik2019batch, ash2019deep, zhan2021multiple}. However, sampling from DPPs in DAL is not trivial since DPPs have a time complexity of $O(N^3)$.

Two-stage (multi-stage) optimization is a popular combined strategy, 
Each stage refines the previous stage's selections using different criteria. 
E.g., stage 1 selects an informative subset with a size larger than $b$, and then stage 2 selects  $b$ samples with maximum diversity.
\textbf{WAAL} uses two stage optimization for implementing discriminative learning via training a DNN for discriminative features in stage 1, and making batch selections in stage 2 \citep{shui2020deep}. \textbf{BADGE} computes gradient embeddings for each unlabeled data samples in stage 1, then clusters by \textbf{KMeans++} in stage 2 \citep{ash2019deep}. Diverse mini-Batch Active Learning (\textbf{DBAL}) \citep{zhdanov2019diverse} first pre-filters unlabeled data pool to the top $\rho b$ most informative/uncertain examples ($\rho$ is pre-filter factor), then clusters these samples to $b$ clusters with (weighted) \textbf{KMeans} and selects $b$ samples closest to the cluster centers.
\subsection{Enhancing of DAL Methods}
\label{tweak_dal}
In Section~\ref{query_strategy}, many highly-cited DAL methods
have designed acquisition functions, e.g.,  \textbf{Entropy}, \textbf{CoreSet} and \textbf{BADGE}. These methods are easily adapted to various tasks since they only involve the data selection process, not the training process of the backbone. However, how well these DAL methods can perform is limited, \eg, one might not exceed the performance of training on full data. Some DAL models are proposed for enhancing DAL methods that can break the limitation, which can be categorized into two branches: data and model aspect. The data aspect includes data augmentation and pseudo labeling, while the model aspect includes attaching extra networks, modifying loss functions, and ensemble. Due to limited space, we exclude related joint tasks that modify DAL methods like semi-/ self-/un-/supervised, transfer, or reinforcement learning.
\paragraph{Data aspect.}

Pseudo-labeling utilizes large-scale unlabeled data for training. Cost-Effective AL (\textbf{CEAL}) \citep{wang2016cost} assigns high-confident (low entropy $H_{\mathcal{M}}[y|\bx]$) pseudo labels predicted by $\mathcal{M}$ for training in the next iteration. However, this introduces new hyperparameters to threshold the prediction confidence, which, if badly tuned, can corrupt the training set with wrong labels \citep{ducoffe2018adversarial}.
Data augmentation uses labeled samples for enlarging the training set. However, data augmentation might waste computational resources because it indiscriminately generates samples that are not guaranteed to be informative. 
\textbf{AdvDeepFool} adds adversarial samples to the training set \citep{ducoffe2018adversarial}, while \textbf{BGADL} employs ACGAN and Bayesian Data Augmentation for producing new artificial samples that are as informative as the selected samples \citep{tran2019bayesian}.

\paragraph{Model aspect.}
Some researchers utilize extra modules to improve target model performance and make selections in DAL. For instance, \textbf{LPL} jointly learns the target backbone model and loss prediction model, which can predict when the target model is likely to produce a wrong prediction.
\citet{choi2021active} constructs mixture density networks to estimate a probability distribution for each localization and classification head's output for the object detection task. Revising the loss function of the target model is also promising. \textbf{WAAL} adopts min-max loss by leveraging the unlabeled data for better distinguishing labeled and unlabeled samples \citep{shui2020deep}. Another approach is ensemble learning. DNNs use a softmax layer to obtain the label's posterior probability and tend to be overconfident when calculating the uncertainty. To increase uncertainty, \citet{gal2017deep} leverages Monte-Carlo (MC) Dropout, where uncertainty in the weights $\omega$ induces prediction uncertainty by marginalizing over the approximate posterior using MC integration. It can be viewed as an ensemble of models sampled with dropouts. \citet{beluch2018power} found that ensembles of multiple classifiers perform better than MC Dropout for calculating uncertainty scores.
\section{Comparative Experiments of DAL}

We conduct comparisons on $19$ methods across $10$ public available datasets, in which these datasets are selected with reference of \citep{ren2021survey} (see Table 2 in \citep{ren2021survey}) and highly-cited DAL papers. 
\subsection{Experimental Settings}
\paragraph{Datasets.}
Considering some DAL approaches currently only support computer vision tasks like \textbf{VAAL}, for consistency and fairness of our experiments, we adopt the image classification tasks, similar to most DAL papers. We use: 
\emph{MNIST} \citep{deng2012mnist}, \emph{FashionMNIST} \citep{xiao2017fashion}, \emph{EMNIST} \citep{cohen2017emnist}, \emph{SVHN} \citep{netzer2011reading}, \emph{CIFAR10}, \emph{CIFAR100} \citep{krizhevsky2009learning} and \emph{TinyImageNet} \citep{le2015tiny}. 
We construct an imbalanced dataset based on \emph{CIFAR10}, called \emph{CIFAR10-imb}, which sub-samples the training set 
with ratios of 1:2:$\cdots$:10 for classes 0 through 9. We also consider medical imaging analysis tasks, including Breast Cancer Histopathological Image Classification (\emph{BreakHis}) \citep{spanhol2015dataset} and Chest X-Ray Pneuomonia classification (Pneumonia-MNIST) \citep{kermany2018identifying}. Additionally, we adopted an object recognition dataset with correlated backgrounds (\emph{Waterbird}) \citep{sagawa2019distributionally, koh2021wilds}. This dataset contains waterbird and landbird classes, which are manually mixed to water and land backgrounds.
It is challenging since DNNs might spuriously rely on the background instead of learning to recognize the object semantics. 
\paragraph{DAL methods.} We test Random Sampling (\textbf{Random}), \textbf{Entropy}, \textbf{Margin}, \textbf{LeastConf} and their MC Dropout versions \citep{beluch2018power} (denoted as \textbf{EntropyD}, \textbf{MarginD}, \textbf{LeastConfD}, respectively), \textbf{BALD}, \textbf{MeanSTD}, \textbf{VarRatio}, \textbf{CEAL(Entropy)}, \textbf{KMeans}, the greedy version of \textbf{CoreSet} (denoted as \textbf{KCenter}), \textbf{BADGE}, \textbf{AdversarialBIM}, \textbf{WAAL}, \textbf{VAAL}, and \textbf{LPL}. For \textbf{KMeans}, considering that we need to cluster large amounts of data, the original  \textbf{KMeans} implementation based on the scikit-learn library \citep{pedregosa2011scikit} will be too time-consuming on large-scale unlabeled data pools. Thus, to save the time cost and let our \emph{$\text{DeepAL}^+$} be more adaptable to DL tasks, we implemented \textbf{KMeans} with GPU (\textbf{KMeans (GPU)}) based on the faiss library \citep{johnson2019billion}. 

For all AL methods, we employed \textbf{ResNet18} (w/o pre-training) \citep{he2016deep} as the basic learner. For a fair comparison, consistent experimental settings of the basic classifier are used across all DAL methods. We run these experiments using \emph{$\text{DeepAL}^+$} toolkit.

\paragraph{Experimental protocol.}
We repeat each experiment for $3$ trials with random splits of the initial labeled and unlabeled pool (using the same random seed) and report average testing performance. For evaluation metrics, for brevity, we report \emph{overall performance} using \emph{area under the budget curve (AUBC)} \citep{zhan2021comp, zhan2021multiple}, where the performance-budget curve is generated by evaluating the DAL method for varying budgets (\eg, accuracy vs. budget). Higher AUBC values indicate better overall performance.
We also report the final accuracy (F-acc), which is the accuracy after the budget $Q$ is exhausted.
More details of experimental settings (\ie, datasets, implementations) are in Section D in Appendix.

\begin{table*}[!htb]
\centering
\resizebox{14cm}{!}{
\begin{tabular}{ll|cc|cc|cc|cc|cc}
\hline 
&& \multicolumn{2}{c|}{\emph{MNIST}}  & \multicolumn{2}{c|}{\emph{FashionMNIST}} & \multicolumn{2}{c|}{\emph{EMNIST}} & \multicolumn{2}{c|}{\emph{SVHN}}  & \multicolumn{2}{c}{\emph{PneumoniaMNIST}} \\
&Model & AUBC & F-acc & AUBC & F-acc & AUBC & F-acc & AUBC & F-acc & AUBC & F-acc \\
\hline 
&Full & $-$ & $0.9916$ & $-$ & $0.9120$ & $-$ & $0.8684$ & $-$ & $0.9190$  & $-$ & $0.9039$ \\
&\textbf{Random} & $0.9570$ & $0.9738$ & $0.8313$ & $0.8434$ & $0.8057$ & $0.8377$ & $0.8110$ & $0.8806$ & $0.8283$ & $\mathbf{0.9077}$ \\
\Xhline{0.6pt}
\multirow{10}{*}{\rotatebox{90}{Unc}} &\textbf{LeastConf} & $0.9677$ & $0.9892$ & $0.8377$ & $0.8820$ & $0.8113$ & $0.8479$ & $0.8350$ & $0.9094$ & $0.8520$ & $\mathbf{0.9097}$ \\
&\textbf{LeastConfD} & \topc{$0.9750$} & \topb{$0.9915$} & $0.8450$ & $0.8744$ & $0.8117$ & \topc{$0.8483$} & $0.8320$ & $0.9083$ &  $0.8243$ & $0.8654$ \\
&\textbf{Margin} & $0.9733$ & $0.9881$ & $0.8427$ & $0.8772$ & $0.8103$ & $0.8468$ & $0.8373$  & \topc{$0.9138$} &  \topc{$0.8580$} & $0.8859$  \\
&\textbf{MarginD} & $0.9703$& $0.9899$ & $0.8417$ & $0.8756$ &  \topb{$0.8197$} & $0.8472$ & $0.8357$ & $0.9104$ &  $0.8230$ & $\mathbf{0.9149}$  \\
&\textbf{Entropy} & $0.9723$ & $0.9883$  & $0. 8397$ & $0.8660$ & $0. 8090$ & $0.8458$ & $0.8297$ & $0.9099$ &  $0.8570$ & $\mathbf{0.9132}$   \\
&\textbf{EntropyD} & $0.9683$ & $0.9887$ & $0.8417$ & $0.8784$ & $0.8167$ &  \topa{$0.8507$}  & $0.8290$ & $0. 9091$ &  $0.8177$ & $0.8710$  \\
&\textbf{BALD} & $0.9697$ & $0.9885$ & $0. 8423$ & \topb{$0.8888$} &  \topb{$0.8197$} & $0.8448$  & $0.8333$ & $0.9020$ &  $0.8270$ & \topc{$\mathbf{0.9204}$}  \\
&\textbf{MeanSTD} & $0.9713$ & $0.9735$ & \topc{$0.8457$} & $0.8766$ & $0.8110$ & $0.8426$ & $0.8323$ & $0.9087$ &  $0.7827$ & $0.8802$    \\
&\textbf{VarRatio} & $0.9717$ & $0.9841$ & $0.8410$ & $0.8754$ & $0.8107$ & \topb{$0.8497$} & $0.8357$ & $0.9079$ &  $0.8530$ & $0.8672$   \\
&\textbf{CEAL(Entropy)} & \topb{$0.9787$} & $0.9889$ & \topb{$0.8477$} & \topc{$0.8826$} & $0.8163$ & $0.8459$ & \topc{$0.8430$} & \topb{$0.9142$} &  $0.8543$ & $\mathbf{0.9179}$   \\
\hline 
\multirow{4}{*}{\rotatebox{90}{Repr/Div}}
&\textbf{KMeans} & $0.9640$ & $0.9813$ & $0.8260$ & $0.8525$ & $0.7903$ & $0.8264$ & $0.8027$ & $0.8671$ &   $0.8243$ & $\mathbf{0.9044}$   \\
&\textbf{KMeans (GPU)} & $0.9637$ & $0.9747$ &  $0.8343$ & $0.8657$ &  $0.7990$ & $0.8362$ &  $0.8120$ & $0.8688$  &  $0.8333$ & $\mathbf{0.9155}$  \\
&\textbf{KCenter} & $0.9740$ & $0.9877$ & $0.8353$ & $0.8466$ & $*$ & $*$ & $0.8283$ & $0.9000$ & $0.8130$ & $\mathbf{0.9189}$  \\
&\textbf{VAAL} & $0.9623$ & $0.9573$ & $0.8297$ & $0.8535$ &$0.8027$ & $0.8363$ & $0.8117$ & $0.8813$ & $0.8393$ & \textbf{$0.9064$}  \\
\hline 
&\textbf{BADGE(KMeans++)} & $0.9707$ & \topc{$0.9904$} & $0.8437$ & $0.8662$ & $*$ & $*$  & $0.8377$ & $0.9057$ &  $0.8340$ & $\mathbf{0.9066}$  \\
\hline 
\multirow{3}{*}{\rotatebox{90}{Enhance}}
&\textbf{AdvBIM} & $0.9680$ & $0.9840$  & $0.8437$ & $0. 8729$  & \#&\# &\# &\# &  $0.8297$ & \textbf{$0.9197$}  \\
&\textbf{LPL} & $0.8913$ & $0.9732$ & $0.7600$ & $0.8471$ & $0.5640$ & $0.6474$ &  \topa{$0.8737$} &  \topa{$\mathbf{0.9452}$} &  \topb{$0.8593$} & \topb{$\mathbf{0.9346}$}  \\
&\textbf{WAAL} & \topa{$0.9890$} &  \topa{$\mathbf{0.9946}$} &  \topa{$0.8703$} &  \topa{$0.8984$} & \topa{$0.8293$} &$0.8423$ & \topb{$0.8603$} & $0.9135$&  \topa{$0.9663$} &\topa{$\mathbf{0.9564}$}\\
\hline 
\hline
 && \multicolumn{2}{c|}{\emph{CIFAR10}}  & \multicolumn{2}{c|}{\emph{CIFAR100}} & \multicolumn{2}{c|}{\emph{CIFAR10-imb}} & \multicolumn{2}{c|}{\emph{Tiny ImageNet}}  & \multicolumn{2}{c}{\emph{BreakHis}} \\
\hline 
&Full & $-$ & $0.8793$ & $-$ & $0.6062$ & $-$ & $0.8036$ & $-$& $0.4583$ & $-$ & $0.8306$ \\
&\textbf{Random} & $0.7967$ & $0.8679$ & $0.4667$ & $0.5903$ & $0. 7103$ & $\mathbf{0.8105}$ & $0. 2577$ & $0. 3544$ &  $0.8010$ & $0.8150$  \\
\Xhline{0.6pt}
\multirow{10}{*}{\rotatebox{90}{Unc}}
&\textbf{LeastConf} & $0.8150$ & $0.8785$ & $0.4747$ & \topb{$\mathbf{0.6072}$} & $0.7330$ & $0.8022$ & $0.2417$ & $0.3470$ & $0.8213$ & $0.8302$ \\
&\textbf{LeastConfD} & $0.8137$ & $\mathbf{0.8825}$ & $0.4730$ & $0.5997$ & $0.7323$ & $\mathbf{0.8065}$  & \topc{$0.2620$} & \topb{$0.3698$} &  $0.8140$ & $\mathbf{0.8313}$  \\
&\textbf{Margin} & \topc{$0.8153$} &  \topc{$\mathbf{0.8834}$} & \topb{$0.4790$} & $0.6010$ & \topc{$0.7367$} & $0.8029$ & $0.2557$ & $0.3611$ &  $0.8217$ & $0.8289$  \\
&\textbf{MarginD} & $0.8140$ &  \topb{$\mathbf{0.8837}$} & \topc{$0.4777$} & $0.6000$ & $0.7260$ & $\mathbf{0.8128}$ & $0.2607$ & $0.3541$ &  $0.8253$ & \topc{$\mathbf{0.8364}$} \\
&\textbf{Entropy} & $0.8130$ & $0.8784$ & $0.4693$ & $0.6048$ & $0.7320$ & \topb{$\mathbf{0.8187}$} & $0.2343$ & $0.3346$ &  $0.8213$ & $0.8251$   \\
&\textbf{EntropyD} & $0.8140$ & $0.8787$ & $0.4677$ & $0.6004$ & $0.7317$ & $0.7963$ & \topa{$0.2627$} &  \topa{$0.3716$} &  $0.8017$ & $0.8115$   \\
&\textbf{BALD} & $0.8103$ & $0.8762$ & $0.4760$ & $0.5942$ & $0.7210$ & $0.7927$ & \topb{$0.2623$} & \topc{$0.3648$} &  $0.8147$ & $0.8296$  \\
&\textbf{MeanSTD} & $0.8087$ & $\mathbf{0.8821}$ & $0.4717$ & $0.5963$ & $0.7203$ & $0.7996$ & $0.2510$ & $0.3551$  &  $0.8053$ & $0.8202$  \\
&\textbf{VarRatio} & $0.8150$ & $0 8780$ & $0.4747$ & $0.5959$ & $0.7353$ & $\mathbf{0.8165}$ & $0.2407$ & $0.3426$ &  $0.8197$ & $0.8264$  \\
&\textbf{CEAL(Entropy)} & $0.8150$ & $\mathbf{0.8794}$ & $0.4693$ & \topc{$0.6043$} & $0.7327$ & \topb{$\mathbf{0.8187}$} & $0. 2347$ & $0.3400$ &  $0.8163$ & $0.8181$  \\
\Xhline{0.6pt}
\multirow{4}{*}{\rotatebox{90}{Repr/Div}}
&\textbf{KMeans} & $0.7910$ & $0.8713$ & $0.4570$ & $0.5834$ & $0.7070$ & $0.7908$ & $0.2447$ & $0.3385$ &  $0.8203$ & \topb{$\mathbf{0.8394}$}   \\
&\textbf{KMeans (GPU)} &  $0.7977$ & $0.8718$ &  $0.4687$ & $0.5842$ &  $0.7140$ & $0.7921$ &  $0.1340$ & $0.2288$ &  $0.8140$ & $\mathbf{0.8323}$   \\
&\textbf{KCenter} & $0.8047$ & $0.8741$ & $0. 4770$ & $0.5993$ & $0.7233$ & $0.7826$ & $0.2540$ &$0.346$ &  $0.8027$ & $0.8289$   \\
&\textbf{VAAL} & $0.7973$ & $0.8679$ & $0.4693$ & $0.5870$ & $0.7113$ & $0.7950$ & $0.1313$ & $0.2191$ &  $0.8197$ & $\mathbf{0.8344}$   \\
\hline 
&\textbf{BADGE(KMeans++)} & $0.8143$ & $\mathbf{0. 8794}$ & \topa{$0.4803$} & $0.6034$ & $0.7347$ & $\mathbf{0.8126}$ & \#& \#& \topb{$0.8343$} & \topb{$\mathbf{0.8470}$} \\
\hline 
\multirow{3}{*}{\rotatebox{90}{Enhance}}
&\textbf{AdvBIM} & $0.7997$ & $0. 8750$ & $0.4713$ & $0.5855$ &\# &\# &\# &\# &  $0.8240$ & $\mathbf{0.8337}$  \\
&\textbf{LPL} &  \topb{$0.8220$} &  \topa{$\mathbf{0.9028}$} & $0.4640$ &  \topa{$\mathbf{0.6369}$} & \topb{$0.7477$} &  \topa{$\mathbf{0.8478}$} & $0.0090$ & $0.0051$ & \topc{$0.8277$} & $\mathbf{0.8316}$   \\
&\textbf{WAAL} & \topa{$0.8253$} & $0.8717$ & $0.4277$ & $0.5560$ &  \topa{$0.7523$} & $0.7993$ & $0.0157$ & $0.0050$ &  \topa{$0.8620$} & \topa{$\mathbf{0.8698}$}   \\
\hline 
\end{tabular}
}
\caption{Overall results of DAL comparative experiments. 
We \textbf{bold} F-acc values that are higher than full performance. We rank F-acc and AUBC of each task with top \topa{1st}, \topb{2nd} and \topc{3rd} with \topa{red}, \topb{teal} and \topc{blue} respectively. 
``$*$'' refers to the experiment needed too much memory, \eg, \textbf{KCenter} on \emph{EMNIST}. ``\#'' refers to the experiment that has not been completed yet. Completed tables of all tasks are shown in Tables 4, 5, 6, 7, and 8 in Appendix.
} 
\label{performance}
\end{table*}

\subsection{Analysis of Comparative Experiments}
\label{analysis_exp}
For analyzing the experiment results, we roughly divided our tasks into three groups: 1) standard image classification, including \emph{MNIST}, \emph{FashionMNIST}, \emph{EMNIST}, \emph{SVHN}, \emph{CIFAR10}, \emph{CIFAR10-imb}, \emph{CIFAR100} and \emph{TinyImageNet}; 2) medical image analysis, including \emph{BreakHis} and \emph{PneumoniaMNIST}; 3) comparative studies, including \emph{MNIST} and \emph{Waterbird}, which would be introduced in Section~\ref{pretrain-sec}.
We report \emph{AUBC (acc)} and F-acc performance in Table~\ref{performance}. 
We provided overall accuracy-budget curves and summarizing tables, as shown in Figure 5 and Tables 4 to 9 in Appendix. 

The typical uncertainty-based strategies on group 1, standard image classification tasks (in Table~\ref{performance}, from \textbf{LeastConf} to \textbf{CEAL}) are generally $1\%\sim3\%$ higher than \textbf{Random} on average performance across the whole AL process (\emph{AUBC}). Among these uncertainty-based methods, we have conducted paired t-tests of each method with the other methods comparing AUBCs across group 1, standard image classification tasks, and no method performs significantly better than the others (all $p$-value are larger than $0.05$). 
Considering dropout, there are only negligible effects (or even counter-intuitive) compared with the original versions (\eg, \textbf{EntropyD} vs.~\textbf{Entropy}) on the normal image classification task, which is consistent with the observations in \citep{beluch2018power}, except for \emph{TinyImageNet}. On \emph{TinyImageNet}, dropout versions are generally $1\%\sim3\%$  higher than the original versions. One possible explanation is that it is not accurate to use the feature representations provided by a single backbone model for calculating uncertainty score, while the dropout technique could help increase the uncertainty, and the differences among the uncertainty scores of unlabeled data samples will be increased. Therefore dropout versions provide better acquisition functions on \emph{TinyImageNet}. Another comparison group is \textbf{CEAL(Entropy)} and \textbf{Entropy}, \textbf{CEAL} improved \textbf{Entropy} by an average of $0.5\%$ on $8$ datasets with threshold of confidence/entropy 1e-5. This idea seems effective, but the threshold must be carefully tuned to get better performance. On medical image analysis tasks (\eg, \emph{PneumoniaMNIST}), performances are slightly different; \textbf{VarRatio} is even $4.5\%$ lower than \textbf{Random}. Additionally, we observed the F-acc of many DAL algorithms are higher than full performance (\eg, F-acc is $0.9189$ on \textbf{KCenter} and $0.9039$ on full data), and the performances of dropout versions are worse than normal methods, \eg, $0.857$ on \textbf{Entropy} and $0.8177$ on \textbf{EntropyD}. These abnormal phenomena could be caused by the distribution shift between training/test sets and data redundancy in AL processes. The detailed explanations are in Section E.1 in Appendix.

Compared with uncertainty-based measures, the performances of representativeness/diversity-based methods (\textbf{KMeans}, \textbf{KCenter} and \textbf{VAAL}) do not show much advantage.
Furthermore, they have relatively high time and memory costs since the pairwise distance matrix used by \textbf{KMeans} and \textbf{KCenter} need to be re-computed in each iteration with current feature representations, while \textbf{VAAL} requires re-training a VAE. Also, a high memory load is needed for storing the pairwise distance matrix for large unlabeled data pools like \emph{EMNIST}). 

Compared with the performance of representativeness-based AL strategies on classical ML tasks \citep{zhan2021comp}, we believe that good representativeness-based DAL performance is based on good feature representations. Our analysis is consistent with the implicit analyses in \citep{munjal2020towards}. Compared with the CPU-version \textbf{KMeans}, \textbf{KMeans (GPU)} is more time-efficient (see Section E.1 in Appendix) and performs better (see Table~\ref{performance}).

Combined strategy \textbf{BADGE} shows its advantage on multiple datasets, where it consistently has relatively better performance. \textbf{BADGE} has $1\%\sim3\%$ higher AUBC performance compared with single \textbf{KMeans} and achieves comparable performance with uncertainty-based strategies and higher AUBC on \emph{CIFAR100} dataset.

For enhanced DAL methods like \textbf{LPL}, \textbf{WAAL}, \textbf{AdvBIM} and \textbf{CEAL}, we are delighted to see their potential over typical DAL methods. For instance, \textbf{LPL} improves F-acc over full training on \emph{SVHN} ($0.9452$ vs. $0.8793$), \emph{CIFAR10} ($0.9028$ vs. $0.8793$), \emph{CIFAR100} ($0.6369$ vs. $0.6062$), and \emph{CIFAR10-imb} ($0.8478$ vs. $0.8036$).
However, \textbf{LPL} is sensitive to hyper-parameters in the LossNet used to predict the target loss, \eg, the feature size determined by FC layer in LossNet. The \textbf{LPL} results on \emph{EMNIST} and \emph{TinyImageNet} indicate that it does not work on all datasets (we have tried many hyper-parameter settings on LossNet but did not work). A similar phenomenon occurs with \textbf{WAAL}. A potential explanation is that both \emph{EMNIST} and \emph{TinyImageNet} contain too many categories, which brings difficulty to the loss prediction in \textbf{LPL} and extracting diversified features in \textbf{WAAL}. This explanation is further verified in Section~\ref{pretrain-sec} -- we adopt a pre-trained \textbf{ResNet18} as the basic classifier, which obtains better feature representations for loss prediction, yielding better performance of \textbf{LPL} compared to the non-pre-trained version ($0.9923$ vs. $0.8913$). The performance comparison on \textbf{CIFAR100} also supported this explanation. \textbf{AdvBIM}, which adds adversarial samples for training, does not achieve ground-breaking performances like \textbf{LPL}.  These adversarial samples are learned by the current backbone model, thus the improvement provided by \textbf{AdvBIM} is marginal. Moreover, \textbf{AdvBIM} is extremely time-consuming since it requires calculating adversarial distance $\mathbf{r}$ for every unlabeled data sample in every iteration. \textbf{AdvBIM} on \emph{EMNIST} and \emph{TinyImageNet} could not be completed due to the prohibitive computing requirements.

\subsection{Ablation Study -- numbers of training epochs and batch size}

Compared to classical ML tasks \citep{zhan2021comp} that are typically convex optimization problems and have globally optimal solutions,
DL typically involves non-convex optimization problems with local minima. Different hyper-parameters like learning rate, optimizer, number of training epochs, and AL batch size lead to other solutions with various performances. Here we conduct ablation studies on the effect of the number of training epochs in each AL iteration and batch size $b$. 
Figure~\ref{ablation} presents the results (see Figure 6 in the Appendix for more figures). Compared with \textbf{Random}, \textbf{Entropy} achieves better performance when the model is trained using more epochs, \eg, \textbf{Entropy} boosts AUBC by around $1.5$ when the model is trained $30$ epochs. We also see that using more training epochs results in better performance, \eg, AUBC gradually increases from around $0.66$ to $0.80$ for \textbf{Random}. It is worth noting that the improvement of AUBC brought by increasing the number of epochs has diminishing returns. Some researchers prefer to use a vast number of epochs during DAL training processes, \eg, \citet{yoo2019learning} used $200$ epochs. However, others like \citep{keskar2016large} suggest that increasing the number of training epochs will not effectively improve testing performance due to generalization problems. Therefore, selecting an optimal number of training epochs is vital for reducing computation costs while maintaining good model performance. Interestingly, AL batch size has less impact on the performance, \eg, \textbf{Entropy} achieves similar performance using different AL batch sizes of $500$, $1000$, and $2000$, which is important for DAL since we can use a relatively large batch size to reduce the number of training cycles. 

\textbf{WAAL} performs consistently better than \textbf{Entropy} and \textbf{BADGE} and the number of training epochs has less impact on its performance, \eg, training with $5$ epochs, \textbf{WAAL} achieves AUBC  $0.825$ when training with 5 epochs and the AUBC remains consistent when increasing to 30 epochs. 
The possible reason is that \textbf{WAAL} considers diversity among samples and constructs a good representation through leveraging the unlabeled data information, thus reducing data bias.
Moreover, we present the accuracy-budget curves using different batch sizes and training epochs in Figure A3. We also conclude that training epochs have less impact on \textbf{WAAL}, which is important for active learning approaches since fewer training epochs will save training time. In addition, \textbf{WAAL} outperforms its counterparts (\ie, \textbf{Badge} and \textbf{Entropy}).

 \begin{figure}[!hbt]
      \centering
      \begin{minipage}{0.45\linewidth}
              \includegraphics[width=\linewidth]{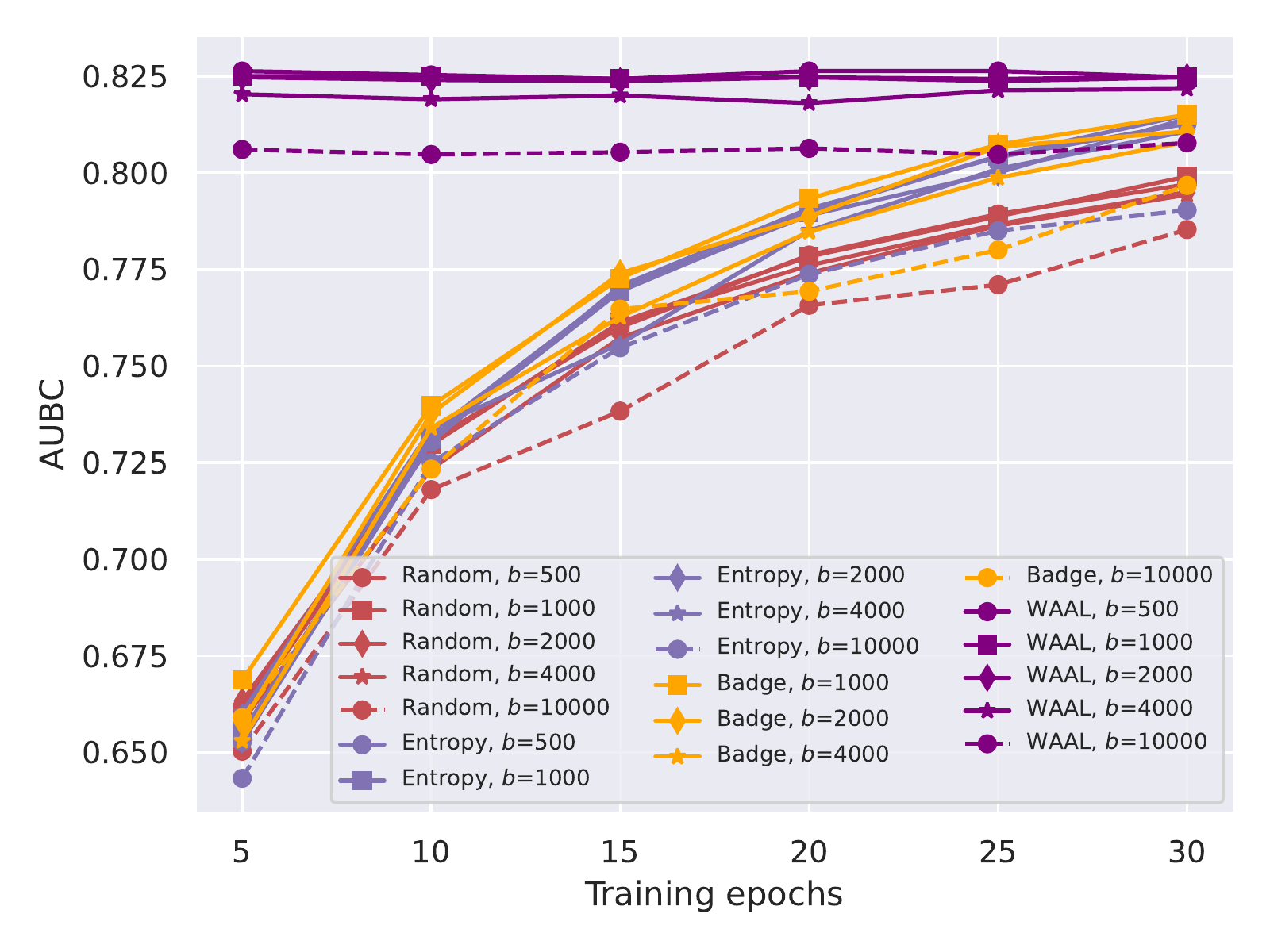}
              \caption{Ablation studies on varying number of epochs and batch size on \emph{CIFAR10}.}\label{ablation}
      \end{minipage}
      \hspace{0.05\linewidth}
      \begin{minipage}{0.45\linewidth}
              \includegraphics[width=\linewidth]{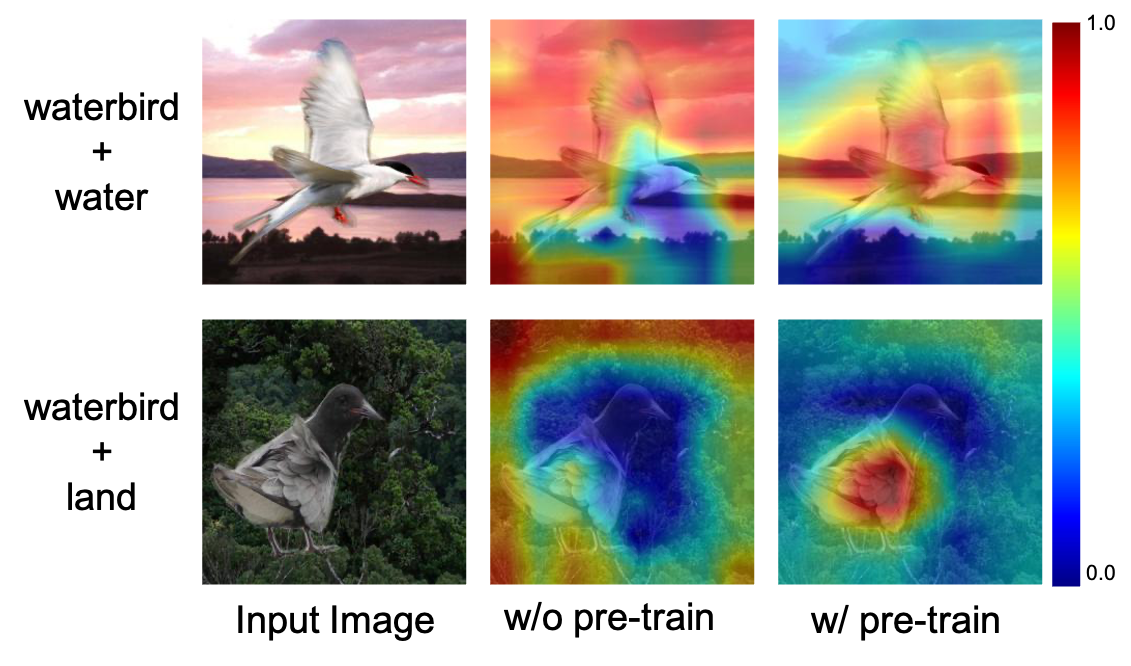}
              \caption{Activation maps of \textbf{ResNet18} w/ and w/o ImageNet1K pre-training.}\label{visualization}
      \end{minipage}
\end{figure}

\begin{figure}[!hbt]
\centering
\subfloat{\includegraphics[width=0.24\linewidth]{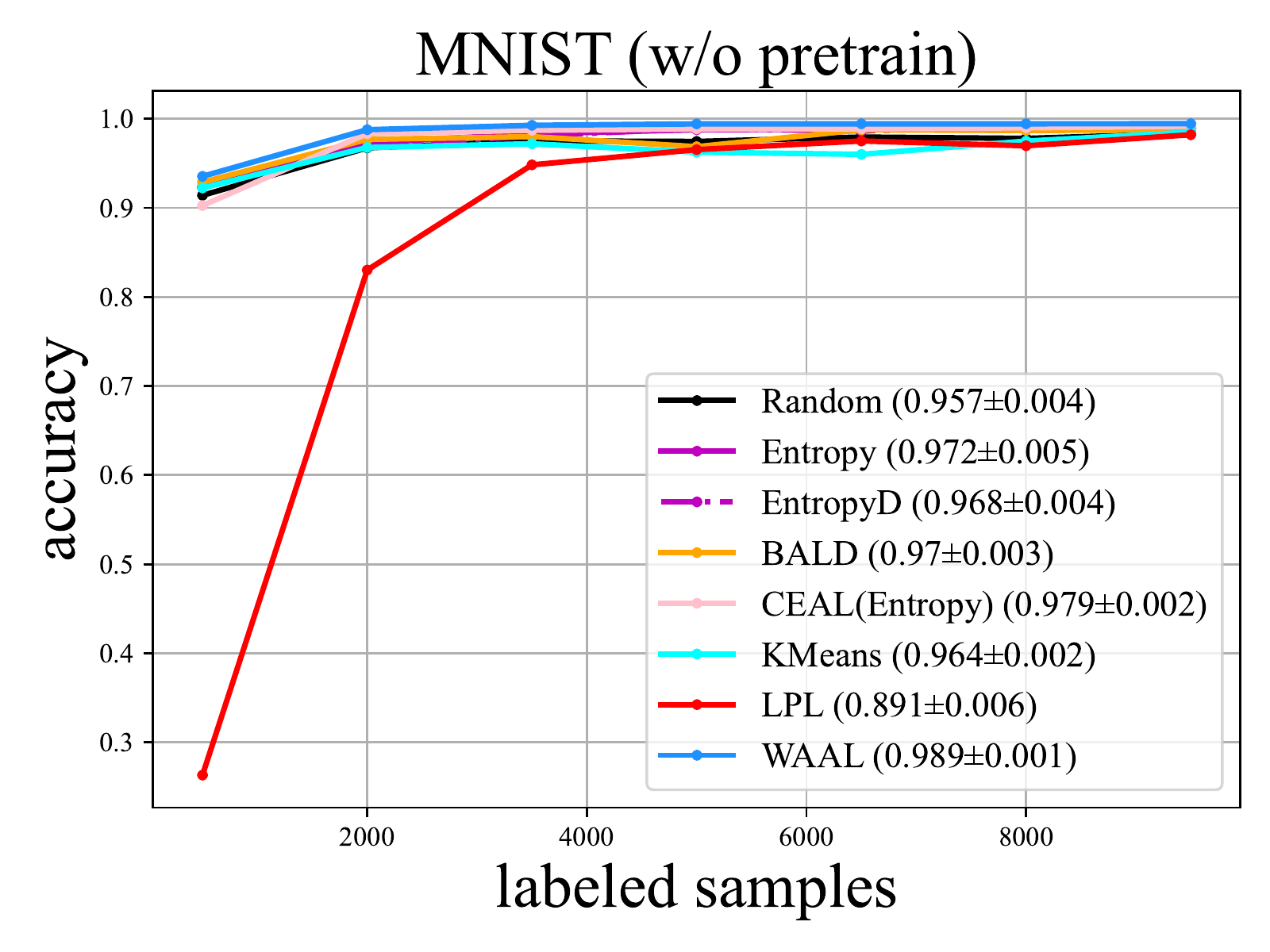}}
\subfloat{\includegraphics[width=0.24\linewidth]{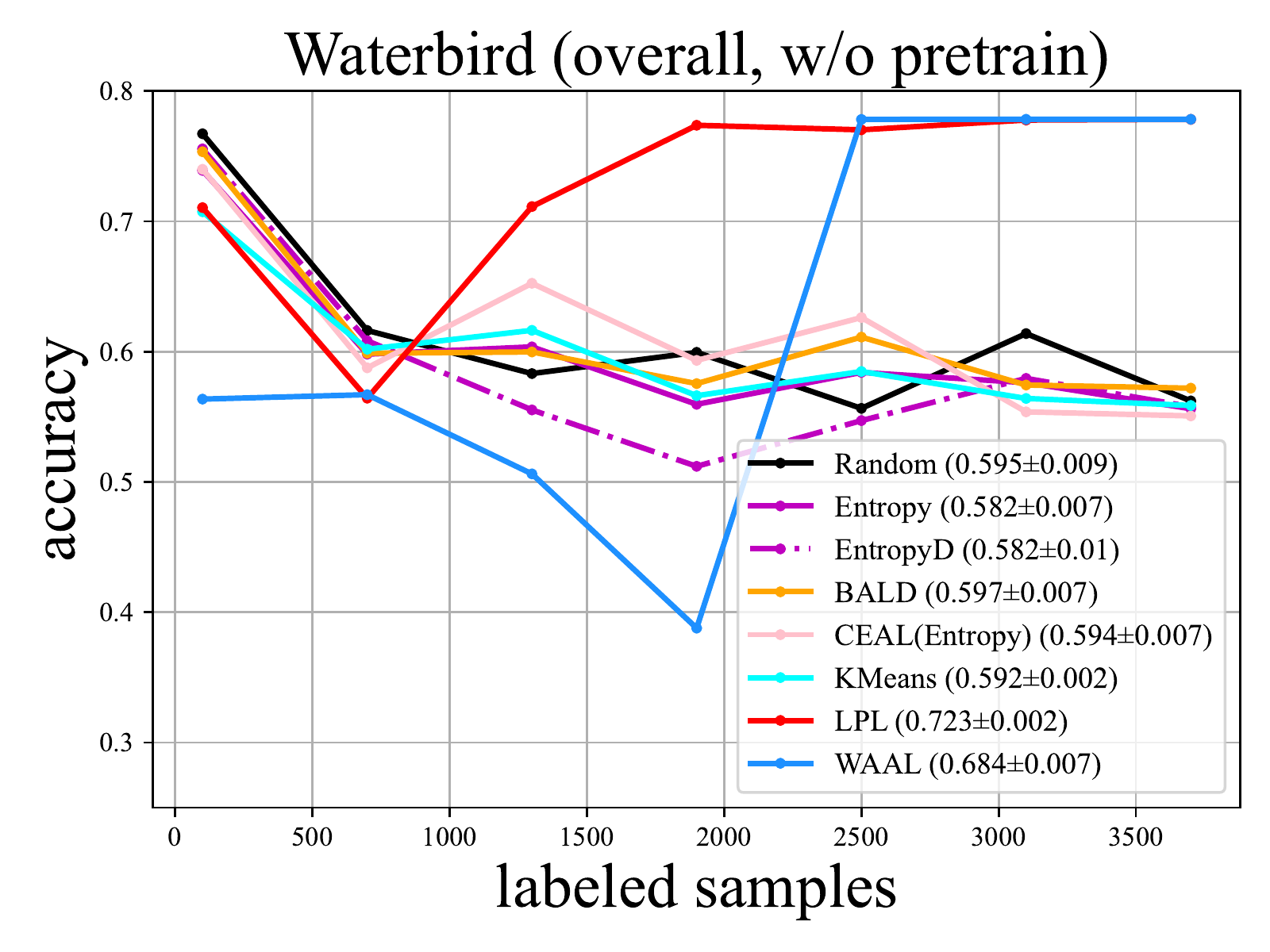}}
\subfloat{\includegraphics[width=0.24\linewidth]{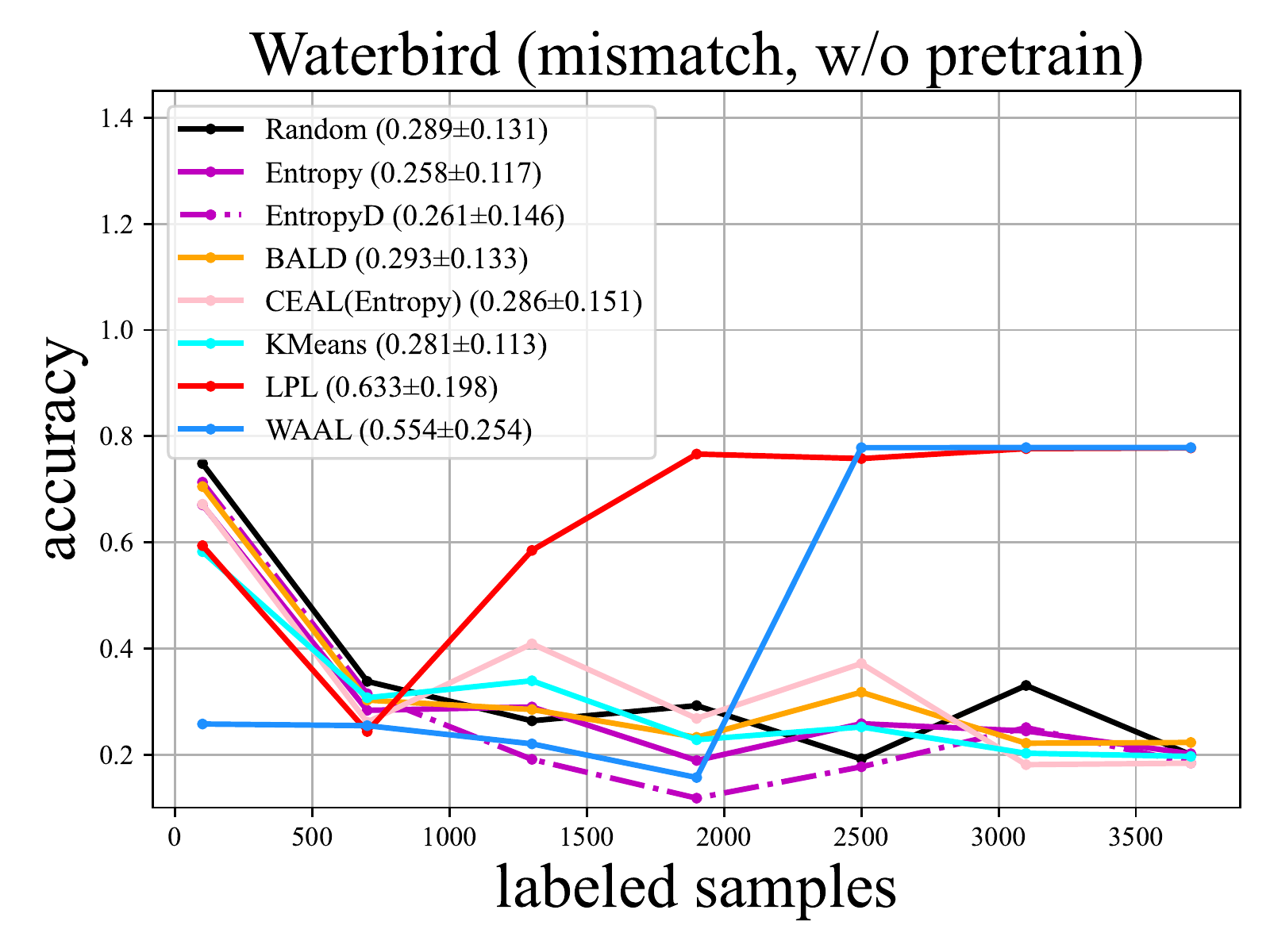}}
\subfloat{\includegraphics[width=0.24\linewidth]{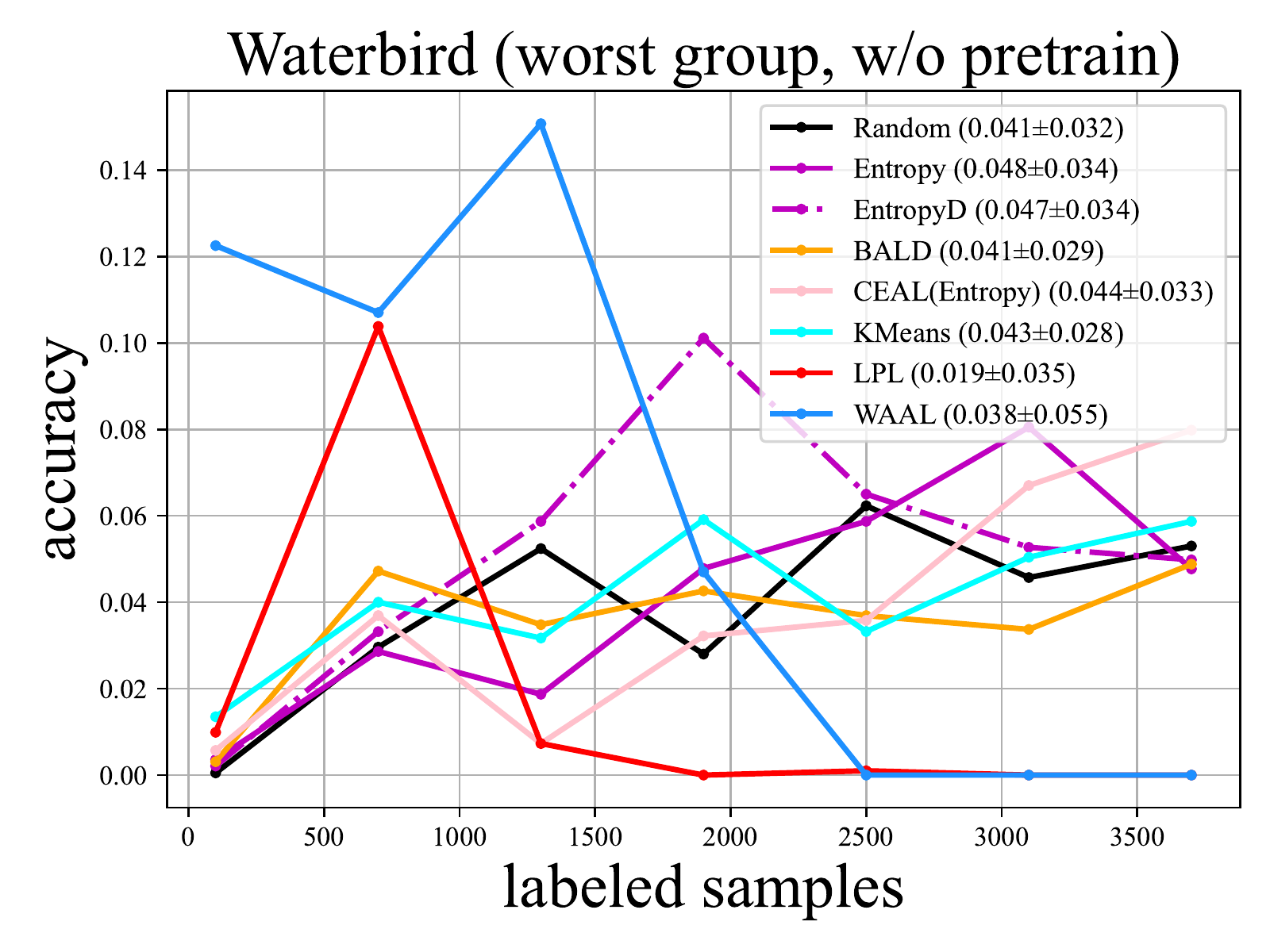}}
\\ 
\subfloat{\includegraphics[width=0.24\linewidth]{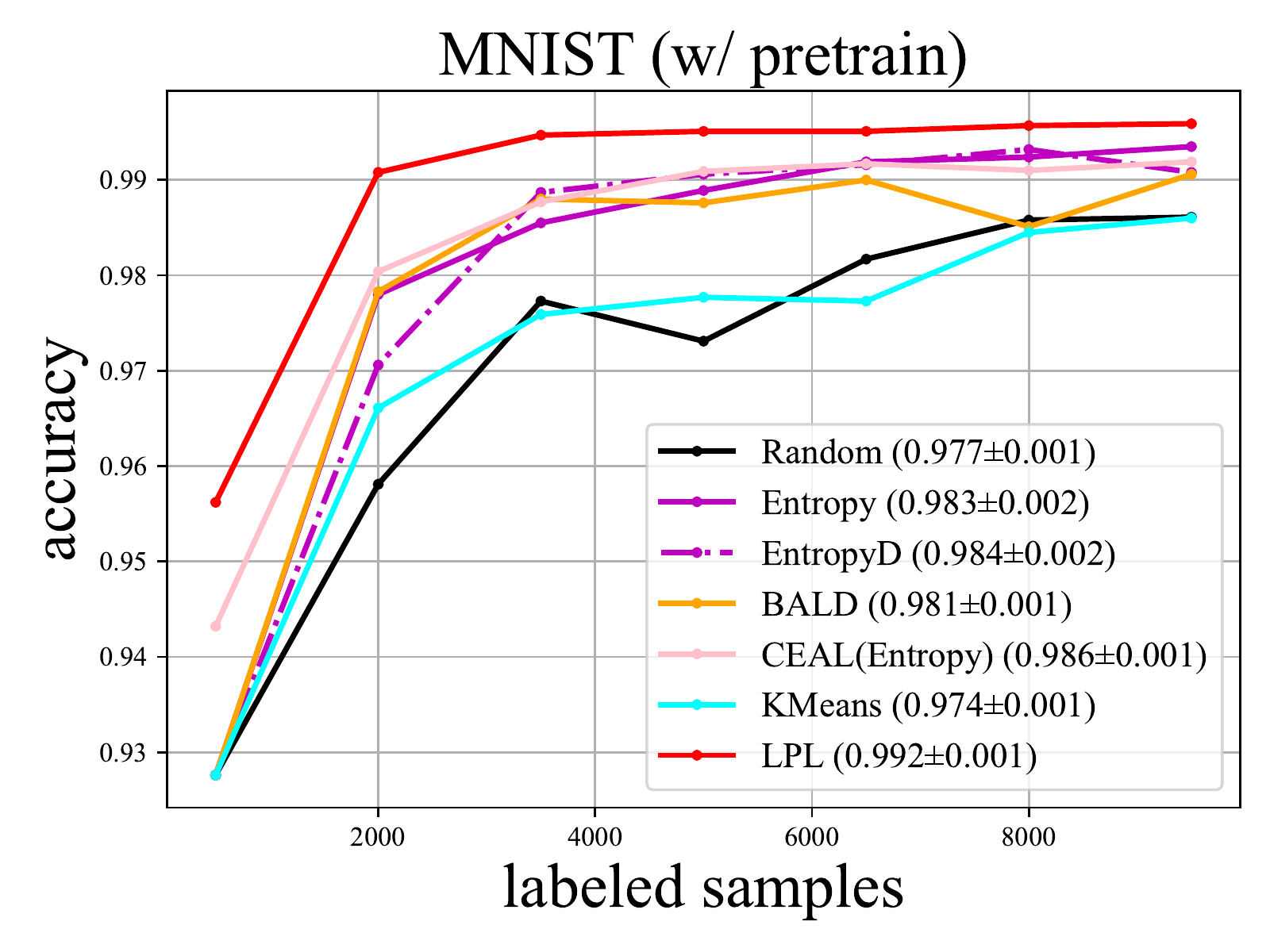}}
\subfloat{\includegraphics[width=0.24\linewidth]{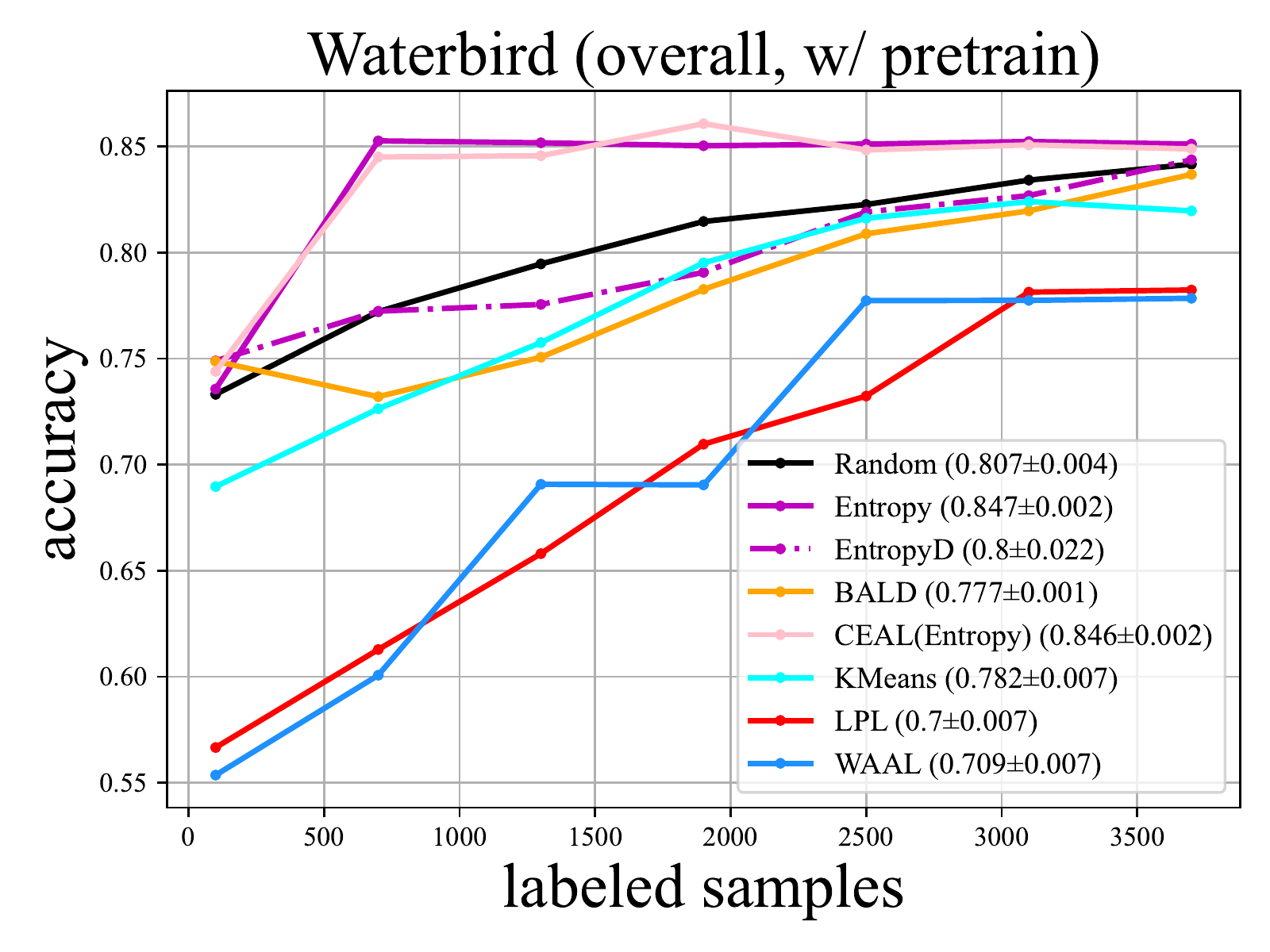}}
\subfloat{\includegraphics[width=0.24\linewidth]{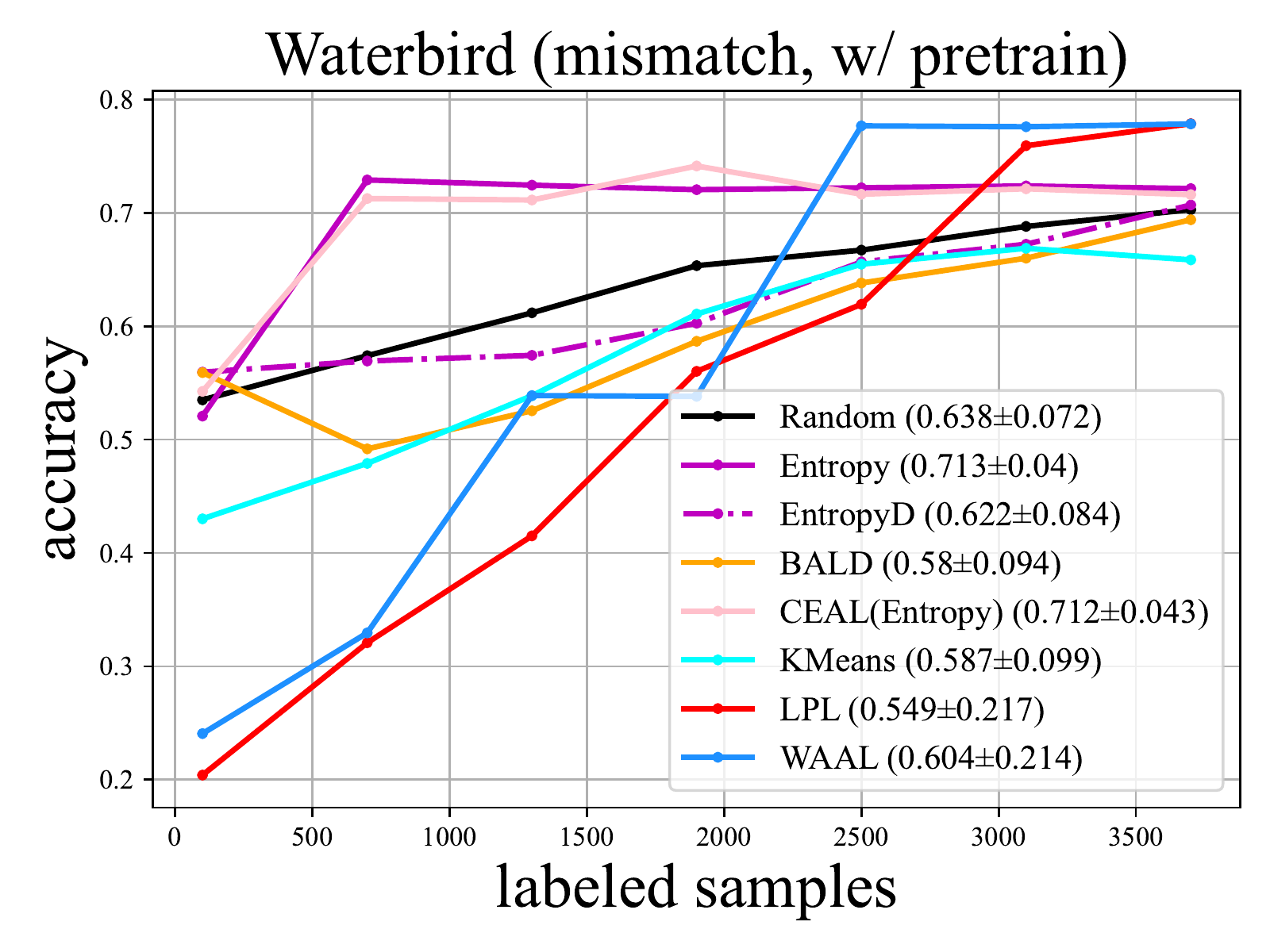}}
\subfloat{\includegraphics[width=0.24\linewidth]{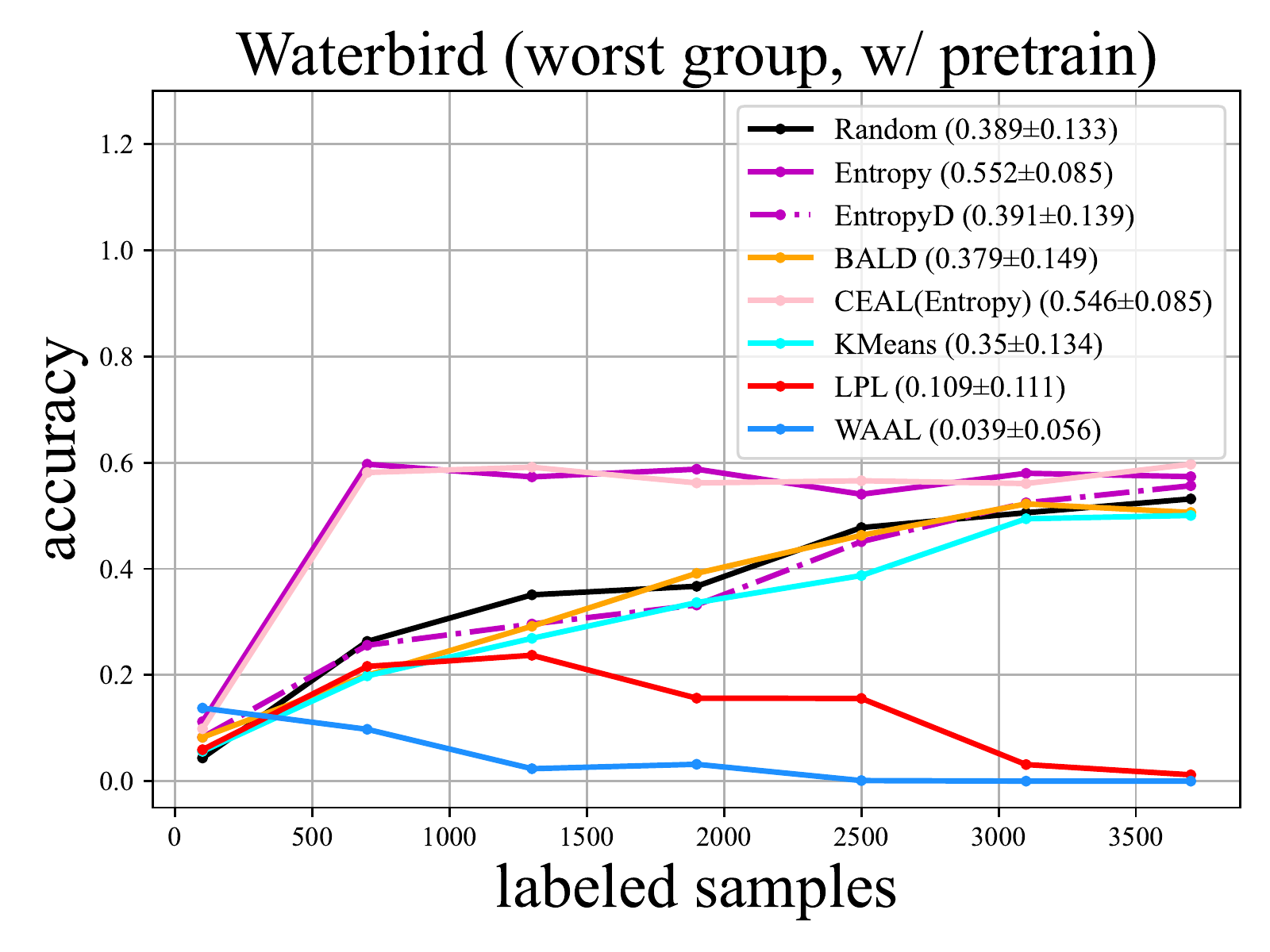}}
\caption{Overall (mismatch, worst group) accuracy vs. budget curves on \emph{MNIST} and \emph{Waterbird} datasets. 
}
\label{ablation-acc}
\end{figure}

\subsection{How pre-training influence DAL performance?}
\label{pretrain-sec}
Pre-training has become a central technique in research and applications of DNNs in recent years \citep{hendrycks2019using}. In our work, we selected \emph{Waterbird} and \emph{MNIST} datasets to conduct comparative experiments, with non pre-trained \textbf{ResNet18} and pre-trained \textbf{ResNet18} (pre-trained on ImageNet-1K data, \textbf{ResNet18P} for short). \emph{Waterbird} and \emph{MNIST} have completely different natures. \emph{Waterbird} dataset contains spurious correlations, and non pre-trained model will focus more on backgrounds, \eg, the classifier will wrongly classify a landbird as a waterbird when the background is water. Pre-trained models provide better feature representations and allow better object semantics (see Figure~\ref{visualization}).  \emph{Waterbird} is separated to four groups based on object and background: $\{(\text{waterbird, water}), \text{(waterbird, land}), \text{(landbird, land}), \text{(landbird, land})\}$. Besides overall accuracy, we also report mismatch and worst group accuracy \citep{sagawa2019distributionally, liu2022empirical}, which refers to the accuracy of groups $\{(\text{waterbird, land}), \text{(landbird, water})\}$ and the lowest accuracy among four groups respectively. On \emph{MNIST}, there is no valid background information. Therefore both models w/ and w/o pre-training would focus on semantic itself. The goal of this experiment is to observe how the pre-training technique influences DAL on hard tasks (\ie, \emph{Waterbird}) and easy/well-studied tasks (\ie, \emph{MNIST}) and how feature representations generated by basic learners influence DAL methods. 

Figure~\ref{ablation-acc} presents overall (also mismatch and worst group in \emph{Waterbird}) accuracy vs. budget curves for \emph{MNIST} and \emph{Waterbird}. On \emph{MNIST}, pre-training does enhance overall DAL performance, but the ranking across these methods is not change (except for \textbf{LPL}), \eg, \textbf{Entropy} $>$ \textbf{EntropyD} $>$ \textbf{KMeans} in both \emph{MNIST} with \textbf{ResNet18} and \textbf{ResNet18P}. \textbf{LPL} performs far better based on \textbf{ResNet18P}, since loss prediction is more accurate with better feature representations. In \emph{Waterbird}, considering \textbf{ResNet18} w/o pre-training, normal DAL methods like \textbf{Entropy}, \textbf{EntropyD}, \textbf{CEAL} and \textbf{KMeans} are affected by the quality of backbone, which influences the DAL selections. Moreover, selecting more data even induces more biases (\emph{Waterbird} is imbalanced among four groups) and cause performance reduction. These DAL methods return to normal when using the pre-trained \textbf{ResNet18P}, since it helps generate predictions with accurate directions (\ie, focus on the object itself, as shown in Figure~\ref{visualization}). On \emph{Waterbird}, \textbf{LPL} and \textbf{WAAL}  w/o pre-training has better performance, possibly because 
they acquire more information with help of enhancing techniques (\ie, loss prediction and collecting unlabeled data information).
However, \textbf{LPL} and \textbf{WAAL} could not well learn worst group under both w/ and w/o pre-training situations. A possible explanation is that they are affected by the imbalance problems of these groups, which induces bias problems in loss prediction and collecting unlabeled data information, and results in poor performance of the worst group.

\section{Challenges and Future Directions of DAL}
Since there is little room for improving DAL by only designing acquisition functions as shown in Sections~\ref{tweak_dal} and~\ref{analysis_exp}, 
researchers focus on proposing effective ways to enhance DAL methods like \textbf{LPL} and enlarging batch size in each round to reduce the time and computation cost.
However, enhancing methods might not work well on all tasks (as shown in Table~\ref{performance}). Better and more universal enhancement methods are needed in DAL. \textbf{Cluster-Margin} have scaled to batch sizes (100K-1M) several orders of magnitude larger than previous studies \citep{citovsky2021batch}, it is hard to be transcended. 

Another notable situation is the research trend shifting towards developing new methodologies to utilize better unlabeled data like semi- and self-supervised learning (S4L). \citet{chan2021marginal} integrated S4L and DAL and conducted extensive experiments with contemporary methods, demonstrating that much of the benefit of DAL is subsumed by S4L techniques. A clear direction is to better leverage the unlabeled data during training in the DAL process under the very few label regimes.

Recently, some researchers employed DAL on more complex tasks like Visual Question Answering (VQA) \citep{karamcheti2021mind} and observed that DAL methods might not perform well. 
 
Many potential reasons limit DAL performance: i) task-specific DAL may be needed for those specific tasks; ii) better feature representations are needed; and iii) various dataset properties need to be considered, like collective outliers in VQA tasks \citep{karamcheti2021mind}. These are possible research directions that demand prompt solutions. Therefore, with the increasing demand for dealing with larger and more complex data in realistic scenarios, e.g., out-of-distribution (OOD), rare classes, imbalanced data, and larger unlabeled data pool \citep{kothawade2021similar, citovsky2021batch}, DAL under different data types are becoming more popular. E.g., \citet{kothawade2021similar} let DAL work on rare classes, redundancy, imbalanced, and OOD data scenarios.

Another possible research direction is to apply DAL techniques with new data-insufficient tasks like automatic driving, medical image analysis, etc. \citet{haussmann2020scalable} have applied AL in an autonomous driving setting of nighttime detection of pedestrians and bicycles to improve nighttime detection of pedestrians and bicycles. It has shown improved detection accuracy of self-driving DNNs over manual curation -- data selected with AL yields relative improvements in mean average precision of $3\times$ on pedestrian detection and $4.4\times$ on detection of bicycles over manually-selected data. \citet{budd2021survey} presents a survey on AL adoption in the medical domain. 
Different tasks have different concerns when integrating DAL techniques. For instance, In medical imaging, there are many rare yet important diseases (\eg, various forms of cancers), while non-cancerous images are much more than compared to the cancerous ones \citep{kothawade2021similar}. Therefore, rare classes must be considered when designing AL strategies in medical image analysis. More importantly, labeling medical images require expertise, and annotation costs and effort remain significant. Task-specific DAL is also worthy of research in recent years.

\begin{ack}
Parts of experiments (including experiments on \emph{PneumoniaMNIST} and \emph{BreakHis} datasets) in this paper were carried out on Baidu Data Federation Platform (Baidu FedCube). For usages, please contact us via \texttt{\url{{fedcube, shubang}@baidu.com}}.
\end{ack}

\bibliographystyle{plainnat}
\bibliography{neurips_2022_benchmark.bib}

\newpage
\appendix

\section{Related Work: comparison with other existing DAL toolkits}
\label{comp-related}
The existing major DAL toolkits/libraries include:
\begin{itemize}
    \item Our previous work \emph{DeepAL}\footnote{\url{https://github.com/ej0cl6/deep-active-learning}} \citep{huang2021deepal}.
    \item Pytorch Active Learning (PAL) Library \footnote{\url{https://github.com/rmunro/pytorch_active_learning}}. Accompanied with the book Human-in-the-Loop Machine Learning.
    \item DISTIL library\footnote{\url{https://github.com/decile-team/distil}}.
\end{itemize}
Compared with our previous work \emph{DeepAL}, we 1) add more built-in support datasets/tasks, where \emph{DeepAL} only support \emph{CIFAR10}, \emph{MNIST}, \emph{FashionMNIST} and \emph{SVHN}, while in \emph{$\text{DeepAL}^+$}, we add \emph{EMNIST}, \emph{TinyImageNet}, \emph{PneumoniaMNIST}, \emph{BreakHis} and wilds-series tasks \citep{koh2021wilds} like \emph{waterbirds}. 2) We then optimize part of existing algorithms to make it better to be adopted on Deep Active Learning tasks like \textbf{KMeans}, we re-implemented it by using faiss-gpu libirary \citep{biyik2019batch}, it is much faster and perform better than scikit-learn library \citep{pedregosa2011scikit} based \textbf{KMeans} implementation. We conduct Principal Component Analysis (PCA) to reduce dimension on representativeness-based approach, \ie, \textbf{KCenter} since it costs too much memory for storing pair-wise similarity matrix on DAL tasks. Note that both \emph{DeepAL} and \emph{$\text{DeepAL}^+$} remove \textbf{CoreSet} approach \citep{sener2017active} since \textbf{Coreset} uses the greedy 2-OPT solution for the k-Center problem as an initialization and checks the feasibility of a mixed integer program (MIP). They adopted Gourbi optimizer\footnote{\url{https://www.gurobi.com/}} to solve MIP and it is not a free optimizer. The users can use \textbf{KCenter}, a greedy optimization of \textbf{CoreSet}. 3) We Add more independent DAL methods implementations like \textbf{MeanSTD}, \textbf{VarRatio}, \textbf{BADGE}, \textbf{LPL}, \textbf{VAAL}, \textbf{WAAL} and \textbf{CEAL}.

PAL library is a fundamental human-in-loop framework. Users need to interact with computers by inputting the ground truth labels of the instance asked by the computer. Besides typical uncertainty-/representativeness-/diversity-/adaptive-based AL approaches like Least Confidence, PAL also includes AL with transfer learning (ALTL). PAL is more likely to provide a template and tell people how to apply AL to different human-in-loop tasks. If someone is new to the AL research field and he could try to use this library to understand how AL works.

DISTIL library majorly serves for the submodular functions proposed by the authors group \citep{kothawade2021similar}, besides the implementations that are already implemented in \emph{DeepAL} and \emph{$\text{DeepAL}^+$}, they have own implementations like \textbf{FASS} \citep{wei2015submodularity}, \textbf{BatchBALD} \citep{kirsch2019batchbald} (we have also implemented this method, it is based on \textbf{BALD}, but it is really memory consuming so we finally deleted it.), \textbf{Glister} \citep{killamsetty2021glister} (for robust learning) and Submodular (Conditional) Mutual Information (\textbf{S(C)MI}) \citep{kothawade2021similar} for AL. It is a easy-to-use library especially if someone want to use their submodular functions. They have no implementations of AL with enhanced techniques like \textbf{LPL} and \textbf{WAAL}.

We made a brief comparison between our \emph{$\text{DeepAL}^+$} and existing DAL libraries, see Table~\ref{toolkit}.
\begin{table}[!htb]
\centering
\begin{tabular}{c|c|p{7.5cm}}
\hline 
Toolkit & \# of implemetations & Comparison with \emph{$\text{DeepAL}^+$} \\
\emph{DeepAL} & 11 & Our previous work, we updated it. \\
\hline
\emph{PAL} & 11 & It contains ALTL methods, \emph{$\text{DeepAL}^+$} have more concrete algorithm re-implementations \\
\hline
\emph{DISTIL} & 20 & Majorly for submodular related functions implementations, no AL with enhanced techniques implementations. \\
\hline
\emph{$\text{DeepAL}^+$} & $19$ & $-$ \\
\hline 
\end{tabular}
\caption{Comparison between our \emph{$\text{DeepAL}^+$} and existing DAL libraries. We excluded \textbf{Random} since strictly speaking, it does not belong to AL approaches.}
\label{toolkit}
\end{table}

\section{More introduction of DeepAL$+$ toolkit}
We listed some introductions of our \emph{$\text{DeepAL}^+$} in previous Section ~\ref{comp-related}. \emph{$\text{DeepAL}^+$} is user-friendly, using a single command can run experiments, we construct the framework/benchmark by easy-to-separate mode, we split the basic networks, querying strategies, dataset/task design, and parameters add-in (\eg, set numbers of training epochs, optimizer parameters). It is simple to add new AL sampling strategies, new basic backbones, and new datasets/tasks in these benchmarks. It makes the users propose new AL sampling strategies easier, test new methods on multiple basic tasks, and compare them with most SOTA DAL methods. We sincerely hope our \emph{$\text{DeepAL}^+$} would help researchers in the DAL research field reduce unnecessary workload and focus on designing new DAL approaches more. This work is ongoing; we would continually add the latest and well-perform DAL approaches and incorporate more datasets/tasks. Moreover, if newly proposed DAL methods are designed based on \emph{DeepAL} or \emph{$\text{DeepAL}^+$}, it would be easier to be further incorporated into our toolkit like \textbf{BADGE} and \textbf{WAAL}.

\section{Licences}
\paragraph{Datasets.}
We listed the licence of datasets we used in our experiments, all datasets employed in our comparative experiments are public datasets:
\begin{itemize}
\item CIFAR10 and CIFAR100 \citep{krizhevsky2009learning}: MIT Licence.  
\item MNIST \citep{deng2012mnist}, EMNIST \citep{cohen2017emnist}: Creative Commons Attribution-Share Alike 3.0 license.
\item FashionMNIST \citep{xiao2017fashion}: MIT Licence.
\item PneumoniaMNIST \citep{kermany2018identifying}: CC BY 4.0 License.
\item BreakHis \citep{spanhol2015dataset}: Creative Commons Attribution 4.0 International License.
\item Waterbird \citep{sagawa2019distributionally, koh2021wilds}: MIT License.
\end{itemize}

\paragraph{Methods.}
We listed all related license of the original implementations of DAL methods that we re-implemented and basic backbone models in our \emph{DeepAL$+$} toolkit:
\begin{itemize}
\item PyTorch \citep{paszke2019pytorch}: Modified BSD.
\item Scikit-Learn \citep{pedregosa2011scikit}: BSD License.
\item BADGE \citep{ash2019deep}: Not listed.
\item LPL \citep{yoo2019learning}: Not listed.
\item VAAL \citep{sinha2019variational}: BSD 2-Clause ``Simplified'' License.
\item WAAL \citep{shui2020deep}: Not listed.
\item CEAL \citep{wang2016cost}: Not listed.
\item Methods originated from DeepAL \citep{huang2021deepal} library implementation: MIT Licence.
\item KMeans (faiss library \citep{johnson2019billion} implementation): MIT Licence.
\item ResNet18 \citep{he2016deep}: MIT License.
\end{itemize}

\section{Experimental Settings}
\subsection{Datasets}
Considering some DAL approaches currently only support computer vision tasks like \textbf{VAAL}, for consistency and fairness of our experiments, we adopt 1) the image classification tasks, similar to most DAL papers. We use the following datasets (details in Table \ref{dataset}): 
\emph{MNIST} \citep{deng2012mnist}, \emph{FashionMNIST} \citep{xiao2017fashion}, \emph{EMNIST} \citep{cohen2017emnist}, \emph{SVHN} \citep{netzer2011reading}, \emph{CIFAR10} and \emph{CIFAR100} \citep{krizhevsky2009learning} and \emph{Tiny ImageNet} \citep{le2015tiny}. Additionally, to explore DAL performance on imbalanced data, we construct an imbalanced dataset based on \emph{CIFAR10}, called \emph{CIFAR10-imb}, which sub-samples the training set 
with ratios of 1:2:$\cdots$:10 for classes 0 through 9. 2) The medical image analysis tasks, including Breast Cancer Histopathological Image Classification (\emph{BreakHis}) \citep{spanhol2015dataset} and Pneumonia-MNIST (pediatric chest X-ray) (\emph{PneumoniaMNIST}) \citep{kermany2018identifying}. Additionally, we adopted an object recognition with correlated backgrounds dataset (\emph{Waterbird}) \citep{sagawa2019distributionally}, it considers two classes: waterbird and landbird. These objected were manually mixed to water and land background, and waterbirds (landbirds) more frequently appearing against a water (land) background. It is a challenging task since DNNs might spuriously rely on background instead of learning to recognize semantic/object.

\begin{table}[!htb]
\centering
\begin{tabular}{lccccccc}
\hline 
Dataset & $\#i$ & $\#u$ & $\#t$ & $b$ & $Q$ & $\#k$ & $\#e$\\
\hline  
\emph{MNIST} & $500$ & $59,500$ & $10,000$ & $250$ & $10,000$ & $10$ & $20$\\ 
\emph{FashionMNIST} & $500$ & $59,500$ & $10,000$ & $250$ & $10,000$ & $10$ & $20$  \\
\emph{EMNIST} & $1,000$ & $696,932$ & $116,323$ & $500$ & $50,000$ & $62$ & $20$ \\
\emph{SVHN} & $500$  & $72,757$ & $26,032$ & $250$ & $20,000$ & $10$ & $20$ \\
\emph{CIFAR10} & $1,000$ & $49,000$ & $10,000$ & $500$ & $40,000$ & $10$ & $30$ \\
\emph{CIFAR100} & $1,000$ & $49,000$ & $10,000$ & $500$ & $40,000$ & $100$ & $40$ \\
\emph{Tiny ImageNet} & $1,000$ & $99,000$ &$10,000$ & $500$ & $40,000$ & $200$ & $40$ \\
\emph{CIFAR10-imb} & $1,000$ & $26,499$ & $10,000$ & $500$ & $20,000$ & $10$ & $30$ \\
\emph{BreakHis} & $100$ & $5,436$ & $2,373$ & $100$ & $5,000$ & $2$ & $10$\\
\emph{PneumoniaMNIST} & $100$ & $5,132$ & $5,232$ & $100$ & $5,000$ & $2$ & $10$ \\
\emph{Waterbird} & $100$ & $4,695$ & $5,794$ & $100$ & $4,000$ & $2$ & $10$ \\
\hline 
\end{tabular}
\caption{Datasets used in comparative experiments. $\#i$ is the size of initial labeled pool, $\#u$ is the size of unlabeled data pool, $\#t$ is the size of testing set, $\#k$ is number of categories and $\#e$ is number of epochs used to train the basic classifier in each AL round.}
\label{dataset}
\end{table}

\subsection{Implementation details}

We employed \textbf{ResNet18}\footnote{\url{https://pytorch.org/vision/stable/models.html\#id10}} \citep{he2016deep} as the basic learner. On \emph{MNIST}, \emph{EMNIST}, \emph{FashionMNIST}, \emph{TinyImagenet}, \emph{CIFAR10} and \emph{CIFAR100}, we adopted \emph{Adam} optimizer (learning rate: $1e-3$) . On \emph{PneumoniaMNIST}, \emph{BreakHis} and \emph{Waterbird}, since Adam would cause overfitting, we use SGD optimizer with learning rate: 1e-2 on \emph{BreakHis} and \emph{PneumoniaMNIST}, learning rate: $0.0005$, weight decay: 1e-5, momentum: $0.9$ on \emph{Waterbird}.

For a fair comparison, consistent experimental settings of the basic classifier are used across all DAL methods. The dataset-specific implementation details are discussed as follows. 
\begin{itemize}
    \item \emph{MNIST}, \emph{FashionMNIST} and \emph{EMNIST}: number of training epochs is $20$, the kernel size of the first convolutional layer in \textbf{ResNet18} is $7 \times 7$ (consistent with the original PyTorch implementation), input pre-processing step include normalization. 
    \item \emph{CIFAR10}, \emph{CIFAR100}: number of training epochs is $30$, the kernel size of the first convolutional layer in \textbf{ResNet18} is $3 \times 3$ (consistent with PyTorch-CIFAR implementation\footnote{\url{https://github.com/kuangliu/pytorch-cifar/blob/master/models/resnet.py}}), input pre-processing steps include random crop (pad=4), random horizontal flip ($p=0.5$)  and normalization. 
    \item \emph{TinyImageNet}: number of training epochs is $40$, the same implementation of \textbf{ResNet18} as \emph{CIFAR}, input pre-processing steps include random rotation (degree=20), random horizontal flip ($p=0.5$) and normalization.
    \item \emph{SVHN}: number of training epochs is $20$, the same implementation of \textbf{ResNet18} as \emph{MNIST}, input pre-processing steps include normalization. 
    \item \emph{BreakHis}: number of training epochs is $10$, the same implementation of \textbf{ResNet10} as \emph{CIFAR}, input pre-proccessing steps include random rotation (degree=90), random horizontal flip ($p=0.8$), random resize crop (scale=224), randomly change the brightness, contrast, saturation and hue of image -- ColorJitter (brightness=0.4, contrast=0.4, saturation=0.4, hue=0.1) and normalization. 
    \item \emph{PneumoniaMNIST}: number of training epochs is $10$, the same implementation o \textbf{ResNet18} as \emph{CIFAR}, input pre-processing steps include resize (shape=255), center crop (shape = 224), random horizontal flip ($p=0.5$), random rotation (degree=10), random gray scale, random affine (translate=(0.05, 0.05), degree=0).
    \item \emph{Waterbirds}: number of training epochs is $30$, the same implementation of \textbf{ResNet18} as \emph{MNIST}, input pre-processing steps include random horizontal flip ($p=0.5$). 
\end{itemize}
 
The model-specific implementation details are discussed as follows. For MC Dropout implementation, we employed $10$ forward passes. For \textbf{CEAL(Entropy)}, we set threshold of confidence/entropy score for assigning pseudo labels as $1e-5$. For \textbf{KCenter}, since using the full feature vector would take too much memory in pair-wise distance calculation, we employ Principal components analysis (PCA) to reduce feature dimension to $32$ according to \citep{karamcheti2021mind}. For VAE in \textbf{VAAL}, we followed the same architecture in \citep{sinha2019variational} and train VAE with $30$ epochs with \emph{Adam} optimizer (learning rate: $1e-3$). For \textbf{LPL}, we train LossNet with \emph{Adam} optimizer (learning rate: $1e-2$); since LossNet is co-trained with basic classifier, we firstly co-trained LossNet and basic classifier followed by normal training processes, then we detached feature updating (the same as stop training basic classifier) and assign $20$ extra epochs for training LossNet.

\subsection{Experimental environments in our comparative experiments}
We conduct experiments on a single Tesla V100-SXM2 GPU with 16GB memory except for running experiments on \emph{PneumoniaMNIST} and \emph{BreakHis}, since running them need $>16GB$ and $<32GB$ memories. We run experiments of \emph{PneumoniaMNIST} and \emph{BreakHis} on another single Tesla V100-SXM2 GPU with 32GB memory. We only use a single GPU for each experiment.

\section{Completed Experimental Results}

\subsection{Overall experiments}

\subsubsection{Performance of Standard Image Classification tasks.}
Tables~\ref{performance-1},~\ref{performance-2},~\ref{performance-3} record the overall performances of standard image classification tasks group, including \emph{MNIST}, \emph{FashionMNIST}, \emph{EMNIST}, \emph{SVHN}, \emph{CIFAR10}, \emph{CIFAR1O-imb}, \emph{CIFAR100} and \emph{TinyImageNet} datasets. Including AUBC (acc) performance with mean and standard deviation over 3 trials, the average running time that takes the running time of \textbf{Random} as unit and the F-acc score. 

Note that \textbf{KMeans(GPU)} performs better than \textbf{KMeans} on major tasks, \ie, Table~\ref{performance-2}. However, from the average running time, \textbf{KMeans(GPU)} seems to have more time than \textbf{KMeans}, it does not mean \textbf{KMeans(GPU)} run slower than \textbf{KMeans}, since the running time calculation does not count the waiting time, \eg, wait for memory allocation, the time for data load from GPU to CPU or from CPU to GPU. In \textbf{KMeans}, in every AL iteration, we need to load data (feature embedding) from GPU to CPU and use scikit-learn library to calculate. At this step, the program must waste time waiting for the operating system to allocate memory for calculation. Nevertheless, these waiting times could be saved in \textbf{KMeans(GPU)}. So actually \textbf{KMeans(GPU)} run faster than \textbf{KMeans} on DAL tasks that use GPU for calculation.

\subsubsection{Performance of Medical Image Analysis tasks.}

Table~\ref{performance-4} records the overall performances of medical image analysis group, including \emph{PneumoniaMNIST} and \emph{BreakHis} datasets. Both \textbf{LPL}, \textbf{WAAL} and \textbf{BADGE} perform well on these medical image analysis tasks. Another thing that worth to pay attention is: all MC dropout based versions (\ie, \textbf{LeastConfD}, \textbf{MarginD}. \textbf{EntropyD}, as well as \textbf{BALD}), perform worse than original versions (\ie, \textbf{LeastConf}, \textbf{Margin} and \textbf{Entropy}), especially on \emph{PneumonialMNIST}. For example, on \emph{PneumonialMNIST}, the AUBC value of \textbf{LeastConf} is 0.852, while \textbf{LeastConfD} only have 0.8243 AUBC value. A potential reason is in \emph{PneumoniaMNIST}, to justify whether an image -- a chest X-ray report pneumonia, one needs to check the local lesions and observe the lung's overall condition. The basic learner needs both local and global features to make an accurate prediction. While MC dropout reduces the model capacity and might ignore some feature information, making less convincing predictions \citep{beluch2018power} and hurt DAL performance. Another phenomenon is, considering F-acc, we noticed that many DAL approaches' F-acc are higher than the accuracy trained on full dataset (0.9039), \eg, 0.9149 on \textbf{MarginD}, 0.9204 on \textbf{BALD}, 0.9179 on \textbf{CEAL}, 0.9197 on \textbf{AdvBIM}. These results can be summarized as one phenomenon: the subset selected from the full subset would contribute to better performance. This is because \emph{PheumoniaMNIST} contains distribution/dataset shift between training and testing set. Also, this dataset might contain redundant data samples and confusing information. That is, these medical images are not one-to-one. They are many-to-one. One patient would correspond to several chest X-ray images, which causes redundancy. Additionally, some patients may have more than one disease \eg, we can see on some X-ray images that there is a posterior spinal fixator that the patient used to fix his spine. These features also might influence the predictions.

Compared with standard tasks (\ie, standard image classification tasks in our comparative survey), real-life applications would encounter more unexpected problems like we discussed aforementioned \emph{PneumoniaMNIST} dataset. That is why we are working on adding more different kinds of tasks for testing DAL approaches. We also encourage DAL researchers to try DAL approaches on various data scenarios and tasks.

\begin{table*}[!htb]
\centering
\resizebox{14.0cm}{!}{
\begin{tabular}{l|ccr|ccr|ccr|ccr}
\hline 
& \multicolumn{3}{c|}{\emph{MNIST}}  & \multicolumn{3}{c|}{\emph{MNIST (w/ pre-train)}} & \multicolumn{3}{c|}{\emph{Waterbird}} & \multicolumn{3}{c}{\emph{Waterbird (w/ pre-train)}} \\
\hline
Model & AUBC & F-acc & Time & AUBC & F-acc & Time & AUBC & F-acc & Time & AUBC & F-acc & Time  \\
\hline 
Full & $-$ & $0.9916$ & $-$ & $-$ & $0.9931$ & $-$ & $-$  & $0.5678$ & $-$&  $-$ & $0.8459$ & $-$ \\
\hline
\textbf{Random}  & 0.9570 $\pm$ 0.0036 & 0.9738 & 1.00 & 0.9767 $\pm$ 0.0005 & 0.9822 & 1.00 & 0.5950 $\pm$ 0.0092 & 0.5657 & 1.00 & 0.8070 $\pm$ 0.0043 & 0.8511  & Time  \\ \hline
\textbf{LeastConf}  & 0.9677 $\pm$ 0.0041 & 0.9892 & 1.14 & 0.9833 $\pm$ 0.0012 & 0.9794 & 1.38 & 0.5843 $\pm$ 0.0054 & 0.5612 & 2.78 & \topb{0.8473 $\pm$ 0.0021} & \topb{0.8605}  & 1.00  \\ \hline
\textbf{LeastConfD}  & \topc{0.9745 $\pm$ 0.0008} & \topc{0.9915} & 2.17 & 0.9840 $\pm$ 0.0029 & \topc{0.9926} & 2.37 & 0.5847 $\pm$ 0.0009 & 0.5947 & 2.87 & 0.7947 $\pm$ 0.0173 & 0.8516  & 1.00  \\ \hline
\textbf{Margin}  & 0.9733 $\pm$ 0.0012 & 0.9881 & 1.25 & 0.9813 $\pm$ 0.0029 & 0.9875 & 1.35 & 0.5840 $\pm$ 0.0029 & 0.6064 & 0.99 & 0.8460 $\pm$ 0.0029 & \topa{0.8629}  & 2.01  \\ \hline
\textbf{MarginD}  & 0.9703 $\pm$ 0.0025 & 0.9899 & 2.84 & 0.9843 $\pm$ 0.0005 & 0.9883 & 2.41 & 0.5960 $\pm$ 0.0049 & 0.5629 & 1.19 & 0.7983 $\pm$ 0.0166 & 0.8458  & 1.00  \\ \hline
\textbf{Entropy}  & 0.9723 $\pm$ 0.0052 & 0.9883 & 1.54 & 0.9830 $\pm$ 0.0024 & 0.9907 & 1.09 & 0.5823 $\pm$ 0.0074 & 0.6204 & 1.02 & \topa{0.8473 $\pm$ 0.0019} & 0.8557  & 1.19  \\ \hline
\textbf{EntropyD}  & 0.9643 $\pm$ 0.0045 & 0.9887 & 3.14 & 0.9840 $\pm$ 0.0022 & 0.9912 & 2.08 & 0.5817 $\pm$ 0.0101 & 0.6321 & 1.19 & 0.8000 $\pm$ 0.0222 & 0.8477  & 1.03  \\ \hline
\textbf{BALD}  & 0.9697 $\pm$ 0.0034 & 0.9885 & 3.12 & 0.9807 $\pm$ 0.0009 & 0.9834 & 2.16 & 0.5970 $\pm$ 0.0070 & 0.6136 & 2.01 & 0.7773 $\pm$ 0.0012 & 0.8452  & 2.32  \\ \hline
\textbf{MeanSTD}  & 0.9713 $\pm$ 0.0034 & 0.9735 & 2.50 & \topc{0.9847 $\pm$ 0.0005} & 0.9907 & 2.20 & 0.5890 $\pm$ 0.0063 & 0.5758 & 2.20 & 0.7890 $\pm$ 0.0107 & 0.8401  & 1.64  \\ \hline
\textbf{VarRatio}  & 0.9717 $\pm$ 0.0083 & 0.9841 & 1.77 & \topc{0.9847 $\pm$ 0.0005} & 0.9902 & 1.26 & 0.5803 $\pm$ 0.0026 & 0.5570 & 1.87 & 0.8460 $\pm$ 0.0029 & \topc{0.8577}  & 2.32  \\ \hline
\textbf{CEAL(Entropy)}  & \topb{0.9787 $\pm$ 0.0019} & 0.9889 & 3.33 & \topb{0.9863 $\pm$ 0.0005} & 0.9872 & 2.27 & 0.5943 $\pm$ 0.0071 & 0.5811 & 1.93 & 0.8460 $\pm$ 0.0022 & 0.8518  & 1.06  \\ \hline
\textbf{KMeans}  & 0.9640 $\pm$ 0.0016 & 0.9813 & 8.78 & \# & \# & \# & 0.5920 $\pm$ 0.0022 & 0.5846 & 2.31 & 0.7823 $\pm$ 0.0066 & 0.8410  & 1.43  \\ \hline
\textbf{Kmeans(GPU)}  & 0.9637 $\pm$ 0.0021 & 0.9747 & 10.14 & 0.9743 $\pm$ 0.0005 & 0.98 & 3.34 & 0.5663 $\pm$ 0.0054 & 0.5987 & 1.30 & 0.7937 $\pm$ 0.0041 & 0.8365  & 2.33  \\ \hline
\textbf{KCenter}  & 0.9740 $\pm$ 0.0014 & 0.9877 & 7.10 & 0.9523 $\pm$ 0.0039 & 0.9659 & 8.60 & 0.6097 $\pm$ 0.0108 & \topc{0.6373} & 2.13 & 0.8297 $\pm$ 0.0031 & 0.8555  & 1.31  \\ \hline
\textbf{VAAL}  & 0.9623 $\pm$ 0.0024 & 0.9573 & 19.20 & 0.9737 $\pm$ 0.0005 & 0.9718 & 6.93 & \topc{0.6217 $\pm$ 0.0025} & 0.5758 & 9.78 & 0.8070 $\pm$ 0.0029 & 0.8546  & 9.69  \\ \hline
\textbf{BADGE(KMeans++)}  & 0.9707 $\pm$ 0.0062 & \topb{0.9904} & 32.51 & 0.9647 $\pm$ 0.0026 & 0.9841 & 7.04 & 0.5837 $\pm$ 0.0074 & 0.6194 & 2.47 & \topc{0.8460 $\pm$ 0.0014} & 0.8538  & 1.94  \\ \hline
 \textbf{AdvBIM}  & 0.9680 $\pm$ 0.0037 & 0.9840 & 20.74 & \# & \# & \# & \# & \# & \# & 0.8033 $\pm$ 0.0082 & 0.8380  & 10.64  \\ \hline
\textbf{LPL}  & 0.8913 $\pm$ 0.0062 & 0.9732 & 5.44 & \topa{0.9923 $\pm$ 0.0005} & \topa{0.9955} & 2.29 & \topa{0.7277 $\pm$ 0.0017} & \topb{0.7783} & 4.50 & 0.7803 $\pm$ 0.0073 & 0.7817  & 4.51  \\ \hline
\textbf{WAAL} & \topa{0.9890 $\pm$ 0.0014} & \topa{0.9946} & 36.10 & 0.9780 $\pm$ 0.0008 & \topb{0.9943} & 2.39 & \topb{0.6837 $\pm$ 0.0073} & \topa{0.7784} & 6.87 & 0.7009 $\pm$ 0.0067 & 0.7783 & 6.81  \\ \hline
\hline 

\end{tabular}
}
\caption{Results of DAL comparative experiments with \emph{MNIST} w/ and w/o pre-train and \emph{Waterbird} w/ and w/o pre-train. We report the AUBC for overall accuracy, final accuracy (F-acc)  after quota $Q$ is exhausted, and the average running time of the whole AL processes (including training and querying processes) relative to \textbf{Random}. 
We rank F-acc and AUBC of each task with top \topa{1st}, \topb{2nd} and \topc{3rd} with \topa{red}, \topb{teal} and \topc{blue} respectively.
``\#'' indicates that the experiment has not been completed yet.} \label{performance-1}
\end{table*}

\begin{table*}[!htb]
\centering
\resizebox{14.0cm}{!}{
\begin{tabular}{l|ccr|ccr|ccr|ccr}
\hline 
& \multicolumn{3}{c|}{\emph{CIFAR10}}  & \multicolumn{3}{c|}{\emph{CIFAR10-imb}} & \multicolumn{3}{c|}{\emph{CIFAR100}} & \multicolumn{3}{c}{\emph{SVHN}} \\
\hline 
Model & AUBC & F-acc & Time & AUBC & F-acc & Time & AUBC & F-acc & Time & AUBC & F-acc & Time  \\
\hline
Full & $-$ & $0.8793$ & $-$ & $-$ &  $0.8036$ & $-$ & $-$ &  $0.6062$ & $-$& $-$ &  $0.9190$ & $-$ \\
\textbf{Random} & 0.7967 $\pm$ 0.0005 & 0.8679 & 1.00 & 0.7103 $\pm$ 0.0017 & 0.8015 & 1.00 & 0.4667 $\pm$ 0.0009 & 0.5903 & 1.00 & 0.8110 $\pm$ 0.0008 & 0.8806 & 1.00  \\ \hline
\textbf{LeastConf} & 0.8150 $\pm$ 0.0000 & 0.8785 & 1.04 & 0.7330 $\pm$ 0.0022 & 0.8022 & 1.04 & 0.4747 $\pm$ 0.0009 & \topb{0.6072} & 1.02 & 0.8350 $\pm$ 0.0028 & 0.9094 & 1.05  \\ \hline
\textbf{LeastConfD} & 0.8137 $\pm$ 0.0012 & 0.8825 & 1.10 & 0.7323 $\pm$ 0.0033 & 0.8065 & 1.18 & 0.4730 $\pm$ 0.0008 & 0.5997 & 1.13 & 0.8320 $\pm$ 0.0008 & 0.9083 & 1.68  \\ \hline
\textbf{Margin} & \topc{0.8153 $\pm$ 0.0005} & \topc{0.8834} & 1.01 & \topc{0.7367 $\pm$ 0.0033} & 0.8029 & 0.80 & \topb{0.4790 $\pm$ 0.0008} & 0.6010 & 0.93 & 0.8373 $\pm$ 0.0005 & \topb{0.9138} & 1.35  \\ \hline
\textbf{MarginD} & 0.8140 $\pm$ 0.0008 & \topb{0.8837} & 1.17 & 0.7260 $\pm$ 0.0014 & 0.8128 & 0.86 & \topc{0.4777 $\pm$ 0.0005} & 0.6000 & 1.06 & 0.8357 $\pm$ 0.0034 & 0.9104 & 1.46  \\ \hline
\textbf{Entropy} & 0.8130 $\pm$ 0.0008 & 0.8784 & 1.07 & 0.7320 $\pm$ 0.0019 & \topb{0.8187} & 0.73 & 0.4693 $\pm$ 0.0017 & 0.6048 & 0.78 & 0.8297 $\pm$ 0.0009 & 0.9099 & 1.33  \\ \hline
\textbf{EntropyD} & 0.8140 $\pm$ 0.0000 & 0.8787 & 1.12 & 0.7317 $\pm$ 0.0021 & 0.7963 & 0.78 & 0.4677 $\pm$ 0.0005 & 0.6004 & 0.87 & 0.8290 $\pm$ 0.0008 & 0.9091 & 1.43  \\ \hline
\textbf{BALD} & 0.8103 $\pm$ 0.0009 & 0.8762 & 1.18 & 0.7210 $\pm$ 0.0024 & 0.7927 & 1.20 & 0.4760 $\pm$ 0.0008 & 0.5942 & 1.03 & 0.8333 $\pm$ 0.0005 & 0.9020 & 1.51  \\ \hline
\textbf{MeanSTD} & 0.8087 $\pm$ 0.0009 & 0.8821 & 1.11 & 0.7203 $\pm$ 0.0017 & 0.7996 & 0.78 & 0.4717 $\pm$ 0.0012 & 0.5963 & 1.11 & 0.8323 $\pm$ 0.0026 & 0.9087 & 2.52  \\ \hline
\textbf{VarRatio} & 0.8150 $\pm$ 0.0008 & 0.8780 & 1.00 & 0.7353 $\pm$ 0.0024 & 0.8165 & 1.03 & 0.4747 $\pm$ 0.0012 & 0.5959 & 0.97 & 0.8357 $\pm$ 0.0009 & 0.9079 & 1.45  \\ \hline
\textbf{CEAL(Entropy)} & 0.8150 $\pm$ 0.0016 & 0.8794 & 1.00 & 0.7327 $\pm$ 0.0050 & \topb{0.8187} & 0.75 & 0.4693 $\pm$ 0.0005 & \topc{0.6043} & 0.94 & \topc{0.8430 $\pm$ 0.0028} & \topc{0.9142} & 1.16  \\ \hline
\textbf{KMeans} & 0.7910 $\pm$ 0.0016 & 0.8713 & 0.50 & 0.7070 $\pm$ 0.0029 & 0.7908 & 3.06 & 0.4570 $\pm$ 0.0008 & 0.5834 & 1.01 & 0.8027 $\pm$ 0.0012 & 0.8671 & 5.22  \\ \hline
\textbf{KMeans(GPU)} & 0.7977 $\pm$ 0.0009 & 0.8718 & 1.64 & 0.7140 $\pm$ 0.0008 & 0.7921 & 1.54 & 0.4687 $\pm$ 0.0005 & 0.5842 & 1.28 & 0.8120 $\pm$ 0.0008 & 0.8688 & 5.76  \\ \hline
\textbf{KCenter} & 0.8047 $\pm$ 0.0012 & 0.8741 & 0.98 & 0.7233 $\pm$ 0.0009 & 0.7826 & 2.87 & 0.4770 $\pm$ 0.0016 & 0.5993 & 1.02 & 0.8283 $\pm$ 0.0017 & 0.9000 & 5.66  \\ \hline
\textbf{VAAL} & 0.7973 $\pm$ 0.0009 & 0.8679 & 1.26 & 0.7113 $\pm$ 0.0012 & 0.7950 & 4.58 & 0.4693 $\pm$ 0.0005 & 0.5870 & 1.20 & 0.8117 $\pm$ 0.0012 & 0.8813 & 9.84  \\ \hline
\textbf{BADGE(KMeans++)} & 0.8143 $\pm$ 0.0005 & 0.8794 & 2.08 & 0.7347 $\pm$ 0.0019 & 0.8126 & 5.91 & \topa{0.4803 $\pm$ 0.0005} & 0.6028 & 1.12 & 0.8377 $\pm$ 0.0017 & 0.9057 & 10.27  \\ \hline
\textbf{AdvBIM} & 0.7997 $\pm$ 0.0005 & 0.8750 & 2.59 & \# & \# & \# & \# & \# & \# & \# & \# & \# \\ \hline
\textbf{LPL} & \topb{0.8220 $\pm$ 0.0014} & \topa{0.9028} & 2.19 & \topb{0.7477 $\pm$ 0.0060} & \topa{0.8478} & 3.14 & 0.4640 $\pm$ 0.0024 & \topa{0.6369} & 0.71 & \topa{0.8737 $\pm$ 0.0061} & \topa{0.9452} & 2.26  \\ \hline
\textbf{WAAL} & \topa{0.8253 $\pm$ 0.0005} & 0.8717 & 1.65 & \topa{0.7523 $\pm$ 0.0021} & 0.7993 & 4.00 & 0.4277 $\pm$ 0.0005 & 0.5560 & 1.13 & \topb{0.8603 $\pm$ 0.0017} & 0.9139 & 9.88 \\ \hline
\end{tabular}
}
\caption{Results of DAL comparative experiments, including \emph{CIFAR10}, \emph{CIFAR10-imb}, \emph{CIFAR100} and \emph{SVHN}. We report the AUBC for overall accuracy, final accuracy (F-acc)  after quota $Q$ is exhausted, and the average running time of the whole AL processes (including training and querying processes) relative to \textbf{Random}. We rank F-acc and AUBC of each task with top \topa{1st}, \topb{2nd} and \topc{3rd} with \topa{red}, \topb{teal} and \topc{blue} respectively. ``\#'' indicates that the experiment has not been completed yet.}
\label{performance-2}
\end{table*}

\begin{table*}[!htb]
\centering
\resizebox{14.0cm}{!}{
\begin{tabular}{l|ccr|ccr|ccr}
\hline 
& \multicolumn{3}{c|}{\emph{EMNIST}}  & \multicolumn{3}{c|}{\emph{FashionMNIST}} & \multicolumn{3}{c}{\emph{TinyImageNet}} \\
\hline 
Model & AUBC & F-acc & Time & AUBC & F-acc & Time & AUBC & F-acc & Time \\
\hline 
Full & $-$ & $0.8684$ &  $-$ & $-$ & $0.9120$ & $-$ & $-$ &  $0.4583$ & $-$ \\
\textbf{Random} & 0.8057 $\pm$ 0.0026 & 0.8377 & 1.00 & 0.8313 $\pm$ 0.0034 & 0.8434 & 1.00 & 0.2577 $\pm$ 0.0017 & 0.3544 & 1.00  \\ \hline
\textbf{LeastConf} & 0.8113 $\pm$ 0.0068 & 0.8479 & 1.11 & 0.8377 $\pm$ 0.0029 & 0.8820 & 1.10 & 0.2417 $\pm$ 0.0009 & 0.3470 & 1.00  \\ \hline
\textbf{LeastConfD} & 0.8177 $\pm$ 0.0005 & \topc{0.8483} & 1.90 & 0.8450 $\pm$ 0.0036 & 0.8744 & 1.80 & \topc{0.2620 $\pm$ 0.0000} & \topb{0.3698} & 0.93  \\ \hline
\textbf{Margin} & 0.8103 $\pm$ 0.0041 & 0.8468 & 0.85 & 0.8427 $\pm$ 0.0040 & 0.8772 & 1.21 & 0.2557 $\pm$ 0.0012 & 0.3611 & 1.02  \\ \hline
\textbf{MarginD} & \topb{0.8197 $\pm$ 0.0017} & 0.8472 & 0.72 & 0.8417 $\pm$ 0.0012 & 0.8756 & 2.07 & 0.2607 $\pm$ 0.0017 & 0.3541 & 1.32  \\ \hline
\textbf{Entropy} & 0.8090 $\pm$ 0.0057 & 0.8458 & 0.97 & 0.8397 $\pm$ 0.0029 & 0.8660 & 1.22 & 0.2343 $\pm$ 0.0005 & 0.3346 & 1.02  \\ \hline
\textbf{EntropyD} & 0.8167 $\pm$ 0.0019 & \topa{0.8507} & 0.15 & 0.8417 $\pm$ 0.0033 & 0.8784 & 2.02 & \topa{0.2627 $\pm$ 0.0005} & \topa{0.3716} & 0.09  \\ \hline
\textbf{BALD} & \topc{0.8197} $\pm$ 0.0024 & 0.8448 & 1.87 & 0.8423 $\pm$ 0.0095 & \topb{0.8888} & 2.07 & \topb{0.2623 $\pm$ 0.0017} & \topc{0.3648} & 0.91  \\ \hline
\textbf{MeanSTD} & 0.8110 $\pm$ 0.0014 & 0.8426 & 0.68 & \topc{0.8457 $\pm$ 0.0017} & 0.8766 & 2.18 & 0.2510 $\pm$ 0.0008 & 0.3551 & 0.17  \\ \hline
\textbf{VarRatio} & 0.8107 $\pm$ 0.0060 & \topb{0.8497} & 1.14 & 0.8410 $\pm$ 0.0037 & 0.8754 & 1.27 & 0.2407 $\pm$ 0.0005 & 0.3426 & 1.06  \\ \hline
\textbf{CEAL(Entropy)} & 0.8167 $\pm$ 0.0039 & 0.8459 & 2.05 & \topb{0.8477 $\pm$ 0.0026} & \topc{0.8826} & 1.29 & 0.2347 $\pm$ 0.0009 & 0.3400 & 1.03  \\ \hline
\textbf{KMeans} & 0.7863 $\pm$ 0.0068 & 0.8222 & 1.56 & 0.8260 $\pm$ 0.0036 & 0.8525 & 5.93 & 0.2447 $\pm$ 0.0009 & 0.3385 & 0.55  \\ \hline
\textbf{KMeans(GPU)} & 0.7990 $\pm$ 0.0022 & 0.8362 & 1.90 & 0.8343 $\pm$ 0.0012 & 0.8657 & 7.58 & 0.1340 $\pm$ 0.0008 & 0.2288 & 0.79  \\ \hline
\textbf{KCenter} & * & * & * & 0.8353 $\pm$ 0.0019 & 0.8466 & 0.76 & 0.2540 $\pm$ 0.0000 & 0.3460 & 0.43  \\ \hline
\textbf{VAAL} & 0.8027 $\pm$ 0.0019 & 0.8363 & 1.78 & 0.8297 $\pm$ 0.0012 & 0.8535 & 12.78 & 0.1313 $\pm$ 0.0005 & 0.2191 & 0.15  \\ \hline
\textbf{BADGE(KMeans++)} & * & * & * & 0.8437 $\pm$ 0.0019 & 0.8662 & 19.36 & \# & \# & \#  \\ \hline
\textbf{AdvBIM} & \# & \# & \# & 0.8390 $\pm$ 0.0016 & 0.8737 & 22.86 & \# & \# & \#  \\ \hline
\textbf{LPL} & 0.5447 $\pm$ 0.0023 & 0.6555 & 1.09 & 0.7600 $\pm$ 0.0094 & 0.8471 & 4.00 & 0.0090 $\pm$ 0.0000 & 0.0051 & 0.40  \\ \hline
\textbf{WAAL} & \topa{0.8293 $\pm$ 0.0005} & 0.8423 & 1.59 & \topa{0.8703 $\pm$ 0.0012} & \topa{0.8984} & 18.42 & 0.0157 $\pm$ 0.0005 & 0.0050 & 0.58 \\ \hline
\end{tabular}
}
\caption{Results of DAL comparative experiments, including \emph{EMNIST}, \emph{FashionMNIST} and \emph{TinyImageNet}. We report the AUBC for accuracy, final accuracy (F-acc)  after quota $Q$ is exhausted, and the average running time of the whole AL processes (including training and querying processes) relative to \textbf{Random}.  We rank F-acc and AUBC of each task with top \topa{1st}, \topb{2nd} and \topc{3rd} with \topa{red}, \topb{teal} and \topc{blue} respectively. 
``$*$'' indicates that the experiment needed too much memory, \eg, \textbf{KCenter} on \emph{EMNIST}, while ``\#'' indicates that the experiment has not been completed yet.} \label{performance-3}
\end{table*}

\begin{table*}[!htb]

\centering
\resizebox{14.0cm}{!}{
\begin{tabular}{l|ccr|ccr}
\hline 
& \multicolumn{3}{c|}{\emph{PneumoniaMNIST}}  & \multicolumn{3}{c}{\emph{BreakHis}} \\
\hline 
Model & AUBC & F-acc & Time & AUBC & F-acc & Time \\
\hline 
Full & $-$ & $0.9039$ &  $-$ & $-$ & $0.8306$ & $-$  \\
\textbf{Random} & 0.8283 $\pm$ 0.0073 & 0.9077 & 1.00 & 0.8010 $\pm$ 0.0014 & 0.8150 & 1.00  \\ \hline
\textbf{LeastConf} & 0.8520 $\pm$ 0.0022 & 0.9097 & 0.62 & 0.8213 $\pm$ 0.0017 & 0.8302 & 0.91  \\ \hline
\textbf{LeastConfD} & 0.8243 $\pm$ 0.0127 & 0.8654 & 1.10 & 0.8130 $\pm$ 0.0022 & 0.8269 & 1.02  \\ \hline
\textbf{Margin} & \topc{0.8580 $\pm$ 0.0045} & 0.8859 & 0.96 & 0.8217 $\pm$ 0.0009 & 0.8289 & 1.12  \\ \hline
\textbf{MarginD} & 0.8230 $\pm$ 0.0054 & 0.9149 & 1.27 & 0.8257 $\pm$ 0.0012 & 0.8364 & 1.16  \\ \hline
\textbf{Entropy} & 0.8570 $\pm$ 0.0028 & 0.9132 & 0.95 & 0.8213 $\pm$ 0.0005 & 0.8251 & 1.30  \\ \hline
\textbf{EntropyD} & 0.8177 $\pm$ 0.0045 & 0.8710 & 1.26 & 0.8017 $\pm$ 0.0009 & 0.8115 & 1.49  \\ \hline
\textbf{BALD} & 0.8270 $\pm$ 0.0014 & \topc{0.9204} & 0.87 & 0.8150 $\pm$ 0.0016 & 0.8334 & 0.89  \\ \hline
\textbf{MeanSTD} & 0.7827 $\pm$ 0.0041 & 0.8802 & 1.18 & 0.7960 $\pm$ 0.0016 & 0.8076 & 0.98  \\ \hline
\textbf{VarRatio} & 0.8530 $\pm$ 0.0065 & 0.8672 & 0.72 & 0.8270 $\pm$ 0.0008 & 0.8365 & 0.74  \\ \hline
\textbf{CEAL(Entropy)} & 0.8543 $\pm$ 0.0102 & 0.9179 & 0.75 & 0.8143 $\pm$ 0.0025 & 0.8206 & 0.80  \\ \hline
\textbf{KMeans} & 0.8243 $\pm$ 0.0042 & 0.9044 & 0.64 & 0.8203 $\pm$ 0.0024 & \topc{0.8394} & 0.68  \\ \hline
\textbf{KMeans(GPU)} & 0.8333 $\pm$ 0.0053 & 0.9155 & 1.04 & 0.8140 $\pm$ 0.0016 & 0.8323 & 1.38  \\ \hline
\textbf{KCenter} & 0.8130 $\pm$ 0.0057 & 0.9189 & 1.01 & 0.8027 $\pm$ 0.0012 & 0.8289 & 1.45  \\ \hline
\textbf{VAAL} & 0.8393 $\pm$ 0.0063 & 0.9064 & 5.33 & 0.8197 $\pm$ 0.0021 & 0.8344 & 2.81  \\ \hline
\textbf{BADGE(KMeans++)} & 0.8340 $\pm$ 0.0022 & 0.9066 & 0.56 & \topb{0.8343 $\pm$ 0.0012} & \topb{0.8470} & 0.68  \\ \hline
\textbf{AdvBIM} & 0.8297 $\pm$ 0.0087 & 0.9197 & 3.61 & 0.8240 $\pm$ 0.0008 & 0.8337 & 2.52  \\ \hline
\textbf{LPL} & \topb{0.8593 $\pm$ 0.0087} & \topb{0.9346} & 4.53 & \topc{0.8277 $\pm$ 0.0009} & 0.8316 & 2.66  \\ \hline
\textbf{WAAL} & \topa{0.9663 $\pm$ 0.0012} & \topa{0.9564} & 5.24 & \topa{0.8620 $\pm$ 0.0036} & \topa{0.8698} & 2.78 \\ \hline
\end{tabular}
}
\caption{Results of DAL comparative experiments, including \emph{PneumoniaMNIST} and \emph{BreakHis}. We report the AUBC for overall accuracy, final accuracy (F-acc)  after quota $Q$ is exhausted, and the average running time of the whole AL processes (including training and querying processes) relative to \textbf{Random}.We rank F-acc and AUBC of each task with top \topa{1st}, \topb{2nd} and \topc{3rd} with \topa{red}, \topb{teal} and \topc{blue} respectively. 
} \label{performance-4}
\end{table*}

\begin{table*}[!htb]

\centering
\resizebox{14.0cm}{!}{
\begin{tabular}{l|ccr|ccr|ccr|ccr}
\hline 
& \multicolumn{3}{c|}{\emph{\makecell{Waterbird \\ (mismatch)}}}  & \multicolumn{3}{c|}{\emph{\makecell{Waterbird \\ (mismatch w/ pre-train)}}} & \multicolumn{3}{c|}{\emph{\makecell{Waterbird \\ (worst group)}}}  & \multicolumn{3}{c}{\emph{\makecell{Waterbird \\ (worst group w/ pre-train)}}} \\
\hline 
Model & AUBC & F-acc & Time & AUBC & F-acc & Time & AUBC & F-acc & Time & AUBC & F-acc & Time  \\
\textbf{Random} & 0.2892 $\pm$ 0.1309 & 0.2713 & 1.00 & 0.6378 $\pm$ 0.0720 & 0.7209 & 1.00 & 0.0405 $\pm$ 0.0321 & 0.0535 & 1.00 & 0.3893 $\pm$ 0.1329 & 0.5177 & 1.00  \\ \hline
\textbf{LeastConf} & 0.2587 $\pm$ 0.1158 & 0.2224 & 2.78 & 0.7139 $\pm$ 0.0418 & 0.7402 & 1.00 & 0.0452 $\pm$ 0.0310 & 0.0467 & 2.78 & 0.5508 $\pm$ 0.0872 & 0.5348 & 1.00  \\ \hline
\textbf{LeastConfD} & 0.2691 $\pm$ 0.1525 & 0.2635 & 2.88 & 0.6115 $\pm$ 0.0795 & 0.7229 & 2.01 & 0.0458 $\pm$ 0.0334 & 0.0621 & 2.88 & 0.3869 $\pm$ 0.1383 & 0.5576 & 2.01  \\ \hline
\textbf{Margin} & 0.2592 $\pm$ 0.1167 & 0.3034 & 0.99 & 0.7117 $\pm$ 0.0408 & 0.7450 & 1.00 & 0.0449 $\pm$ 0.0308 & 0.0389 & 0.99 & 0.5509 $\pm$ 0.0851 & 0.5475 & 1.00  \\ \hline
\textbf{MarginD} & 0.2923 $\pm$ 0.1512 & 0.3089 & 1.19 & 0.6190 $\pm$ 0.0779 & 0.7109 & 1.19 & 0.0381 $\pm$ 0.0303 & 0.0581 & 1.19 & 0.3943 $\pm$ 0.1358 & 0.5675 & 1.19  \\ \hline
\textbf{Entropy} & 0.2578 $\pm$ 0.1173 & 0.3255 & 1.03 & 0.7133 $\pm$ 0.0396 & 0.7289 & 1.03 & 0.0475 $\pm$ 0.0337 & 0.0498 & 1.03 & 0.5516 $\pm$ 0.0825 & 0.5540 & 1.03  \\ \hline
\textbf{EntropyD} & 0.2612 $\pm$ 0.1463 & 0.3608 & 1.19 & 0.6221 $\pm$ 0.0842 & 0.7163 & 2.32 & 0.0466 $\pm$ 0.0342 & 0.1158 & 1.19 & 0.3910 $\pm$ 0.1393 & 0.5582 & 2.32  \\ \hline
\textbf{BALD} & 0.2929 $\pm$ 0.1333 & 0.3421 & 2.02 & 0.5798 $\pm$ 0.0943 & 0.7107 & 1.64 & 0.0405 $\pm$ 0.0287 & 0.1022 & 2.02 & 0.3788 $\pm$ 0.1488 & 0.5680 & 1.64  \\ \hline
\textbf{MeanSTD} & 0.2789 $\pm$ 0.1386 & 0.2237 & 2.20 & 0.5999 $\pm$ 0.0925 & 0.7003 & 2.32 & 0.0461 $\pm$ 0.0298 & 0.0472 & 2.20 & 0.3782 $\pm$ 0.1473 & 0.5665 & 2.32  \\ \hline
\textbf{VarRatio} & 0.2512 $\pm$ 0.1063 & 0.2307 & 1.88 & 0.7111 $\pm$ 0.0413 & 0.7333 & 1.06 & 0.0465 $\pm$ 0.0305 & 0.0696 & 1.88 & 0.5509 $\pm$ 0.0854 & 0.5488 & 1.06  \\ \hline
\textbf{CEAL(Entropy)} & 0.2863 $\pm$ 0.1510 & 0.3436 & 1.93 & 0.7119 $\pm$ 0.0432 & 0.7224 & 1.43 & 0.0440 $\pm$ 0.0333 & 0.0363 & 1.93 & 0.5458 $\pm$ 0.0852 & 0.5680 & 1.43  \\ \hline
\textbf{KMeans} & 0.2809 $\pm$ 0.1130 & 0.2984 & 2.31 & 0.5867 $\pm$ 0.0992 & 0.7005 & 2.33 & 0.0428 $\pm$ 0.0275 & 0.0457 & 2.31 & 0.3499 $\pm$ 0.1335 & 0.4943 & 2.33  \\ \hline
\textbf{KMeans(GPU)} & 0.2361 $\pm$ 0.1354 & 0.2831 & 1.31 & 0.6131 $\pm$ 0.0811 & 0.6922 & 1.31 & 0.0544 $\pm$ 0.0384 & 0.0524 & 1.31 & 0.4669 $\pm$ 0.1173 & 0.5774 & 1.31  \\ \hline
\textbf{KCenter} & 0.3272 $\pm$ 0.1520 & 0.3870 & 2.13 & 0.6810 $\pm$ 0.0665 & 0.7156 & 2.12 & 0.0315 $\pm$ 0.0257 & 0.0363 & 2.13 & 0.4872 $\pm$ 0.1188 & 0.5696 & 2.12  \\ \hline
\textbf{VAAL} & 0.3521 $\pm$ 0.1716 & 0.3500 & 9.79 & 0.6469 $\pm$ 0.0517 & 0.7301 & 9.69 & 0.0323 $\pm$ 0.0306 & 0.0633 & 9.79 & 0.3688 $\pm$ 0.1656 & 0.5135 & 9.69  \\ \hline
\textbf{BADGE(KMeans++)} & 0.2637 $\pm$ 0.1347 & 0.3215 & 2.48 & 0.7118 $\pm$ 0.0449 & 0.7260 & 1.94 & 0.0458 $\pm$ 0.0328 & 0.0639 & 2.48 & 0.5391 $\pm$ 0.1005 & 0.5587 & 1.94  \\ \hline
\textbf{AdvBIM} & \# & \# & \# & 0.6309 $\pm$ 0.0836 & 0.6965 & 10.64 & \# & \# & \# & 0.4164 $\pm$ 0.1499 & 0.5992 & 10.64  \\ \hline
\textbf{LPL} & 0.6332 $\pm$ 0.1979 & 0.7783 & 4.50 & 0.5494 $\pm$ 0.2175 & 0.7786 & 4.51 & 0.0187 $\pm$ 0.0035 & 0.0005 & 4.50 & 0.1094 $\pm$ 0.0112 & 0.0099 & 4.51  \\ \hline
\textbf{WAAL} & 0.5543 $\pm$ 0.2544 & 0.7784 & 6.88 & 0.6036 $\pm$ 0.2138 & 0.7782 & 6.81 & 0.0382 $\pm$ 0.0054 & 0.0005 & 6.88 & 0.0390 $\pm$ 0.0055 & 0.0010 & 6.81 \\ \hline
\hline 

\end{tabular}
}
\caption{Results of \emph{Waterbird}. We report the AUBC for mismatch and worst group accuracy, final accuracy (F-acc)  after quota $Q$ is exhausted, and the average running time of the whole AL processes (including training and querying processes) relative to \textbf{Random}. We \textbf{bold} F-acc values that are higher than full performance. We did not rank the top three methods since labeling them is not of great reference value.
 ``\#'' indicates that the experiment has not been completed yet.} \label{performance-5}
\end{table*}

\subsection{DAL method ranking and summarizing}

We next consider the overall performance of DAL methods on the eight standard image classification datasets and two medical image analysis datasets using win-tie-loss counts, respectively, as shown in Table \ref{wintieloss}. We use a margin of $0.5\%$, \eg, 
 a ``win'' is counted for method A if it outperforms method B by $0.5\%$ in pairwise comparison. Table \ref{wintieloss} shows the advantage of  uncertainty-based DAL methods like\textbf{LeastConfD} (3rd), and pseudo labeling for enhancing uncertainty-based DAL methods, \ie, \textbf{CEAL} (2nd). \textbf{WAAL} perform the best. Additionally, Dropout method also can improve DAL methods, \eg, \textbf{LeastConfD} ranks 3nd while \textbf{LeastConf} only ranks 9th. \textbf{LPL} only 13th. Although they achieve the best performances on some datasets (\eg, \emph{SVHN}, \emph{CIFAR10}) and have high win counts, they also perform extremely poorly on other datasets (e.g., \emph{Tiny ImageNet}), which contributes to their low ranking. \textbf{VAAL} and \textbf{KMeans} perform even worse than \textbf{Random}. \textbf{AdvBIM} ranks far behind due to many incomplete tasks. This is a drawback of \textbf{AdvBIM} (also of \textbf{AdvDeepFool}), that is, these methods spend too much for re-calculating adversarial distance $\mathbf{r}$ for each unlabeled data sample per AL round. 

On medical image analysis tasks, both \textbf{WAAL} and \textbf{LPL} outperform other DAL approaches, which constantly shows the advantage of DAL with enhancing techniques. Combined strategy, \textbf{BADGE} also obtained a good ranking (4th). More noticeably, \textbf{BADGE} always obtains comparable performances on various tasks. Therefore, for new/unseen tasks/data, we recommend first trying combined DAL approaches. In medical image analysis tasks, \textbf{VAAL} perform better than standard image classification tasks.  

\begin{table*}[!htb]
\small
\centering
\begin{tabular}{l|ll|ll}
\hline
 & \multicolumn{2}{c|}{\emph{Standard Image Classification (8 datasets)}}  & \multicolumn{2}{c}{\emph{Medical Image Analysis (2 datasets)}} \\
Rank & Method & $\text{win}-\text{tie}-\text{loss}$ & Method & $\text{win}-\text{tie}-\text{loss}$ \\
\hline
1 & \textbf{WAAL} & $103-2-31$ & \textbf{WAAL} & $34-0-0$ \\
2 & \textbf{CEAL} & $74-35-27$ &  \textbf{LPL} & $25-6-3$\\
3 & \textbf{LeastConfD} & $63-59-14$  & \textbf{VarRatio} & $23-7-4$ \\
4 & \textbf{MarginD} & $61-55-20$ & \textbf{BADGE} & $24-1-9$ \\
5 & \textbf{Margin} & $60-57-19$ &  \textbf{Margin}& $19-10-5$  \\
6 & \textbf{BALD} & $56-59-21$ & \textbf{Entropy} & $18-11-5$ \\
7 & \textbf{EntropyD} & $54-55-27$ & \textbf{LeastConf} & $18-9-7$ \\
8 & \textbf{VarRatio} & $52-58-26$  &  \textbf{VAAL} & $16-9-12$ \\
9 & \textbf{LeastConf} & $51-53-32$ &\textbf{CEAL} & $15-7-12$ \\
10 & \textbf{Badge} & $46-49-41$ &\textbf{AdvBIM} & $13-11-10$ \\
11 & \textbf{MeanSTD} & $44-50-42$ &\textbf{MarginD} & $12-9-13$ \\
12 & \textbf{Entropy} & $40-54-42$ &\textbf{KMeans} & $10-9-15$ \\
13 & \textbf{LPL} & $57-4-75$ &\textbf{BALD} & $7-8-19$ \\
14 & \textbf{KCenter}& $41-34-61$ &\textbf{LeastConfD} & $7-6-21$ \\
15 & \textbf{Random} & $26-23-87$ &\textbf{Random} & $4-7-23$ \\
16 & \textbf{VAAL} & $20-15-101$ &\textbf{EntropyD} & $2-3-29$ \\
17 & \textbf{KMeans} & $18-9-109$ &\textbf{KCenter} & $2-3-29$ \\
18 & \textbf{AdvBIM} & $10-25-101$ &\textbf{MeanSTD} & $0-1-33$ \\
\hline
\end{tabular}
\caption{Comparison of DAL methods using win-tie-loss across 8
datasets on standard image classification tasks and 2 medical image analysis tasks with AUBC (acc). Methods are ranked by $2 \times win + tie$.}
\label{wintieloss}
\end{table*}

\begin{figure}[!hbt]
\centering
\subfloat[MNIST]{\includegraphics[width=0.33\linewidth]{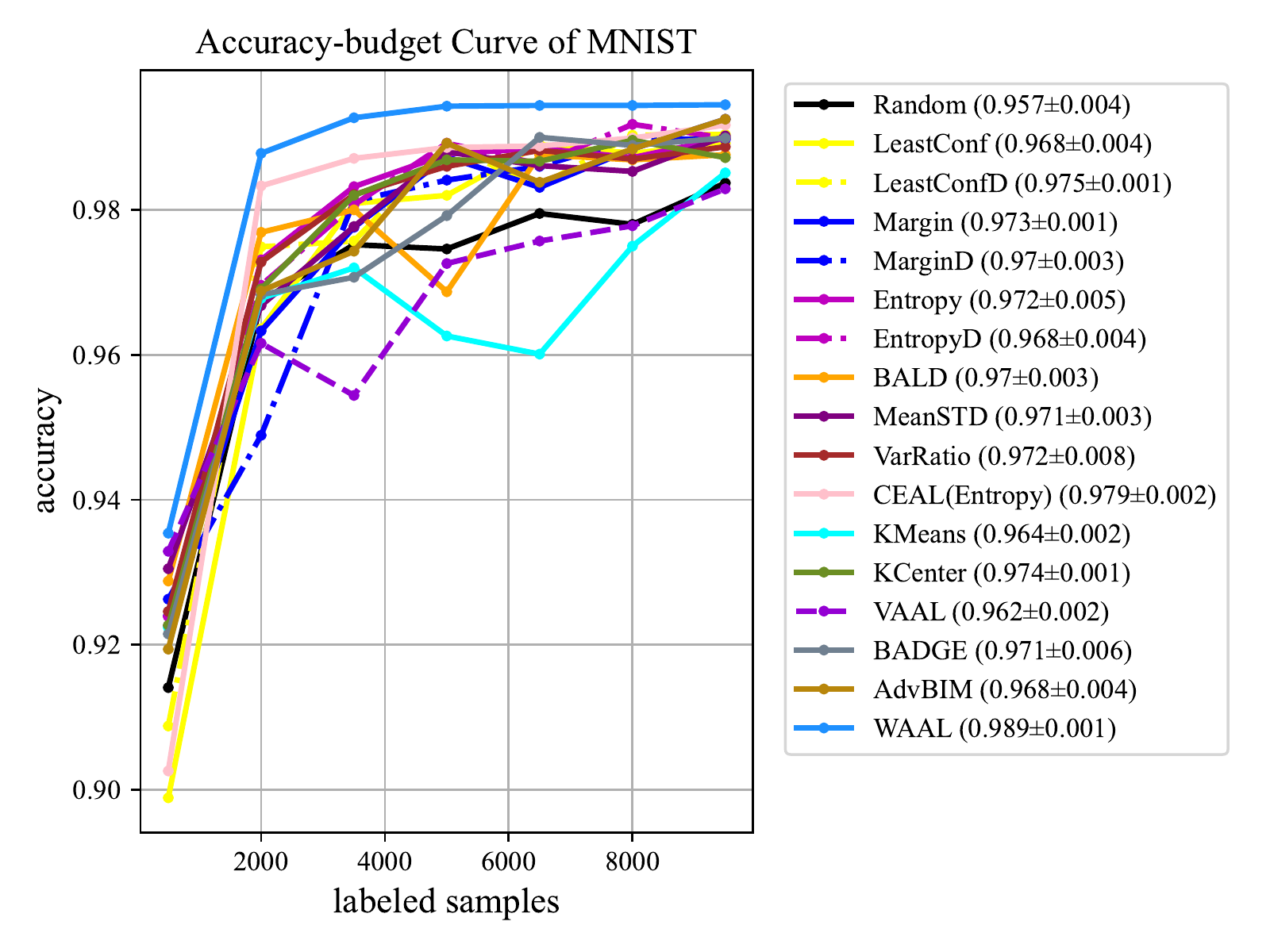}}
\subfloat[FashionMNIST]{\includegraphics[width=0.33\linewidth]{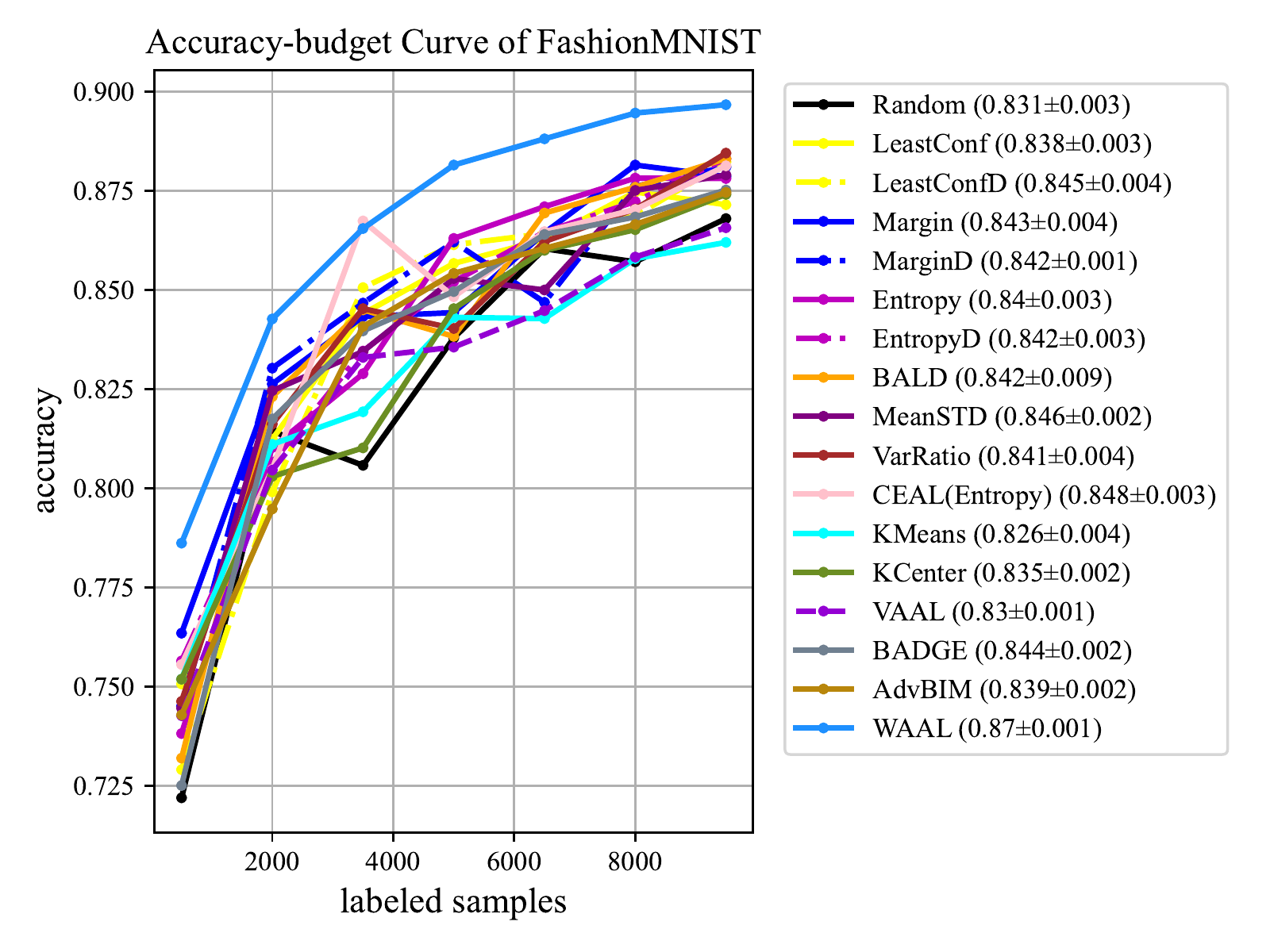}}
\subfloat[SVHN]{\includegraphics[width=0.33\linewidth]{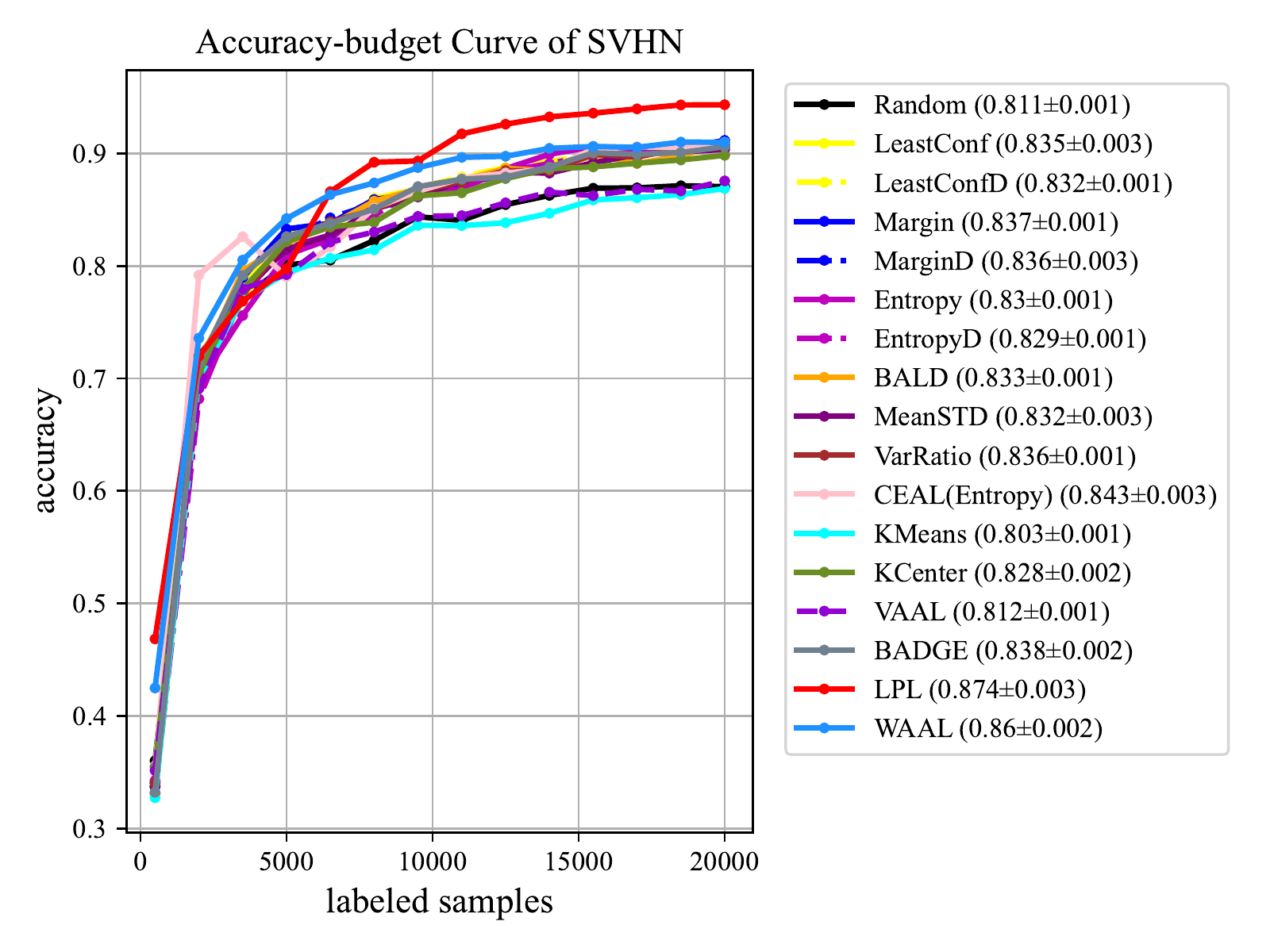}}
\\
\subfloat[CIFAR10]{\includegraphics[width=0.33\linewidth]{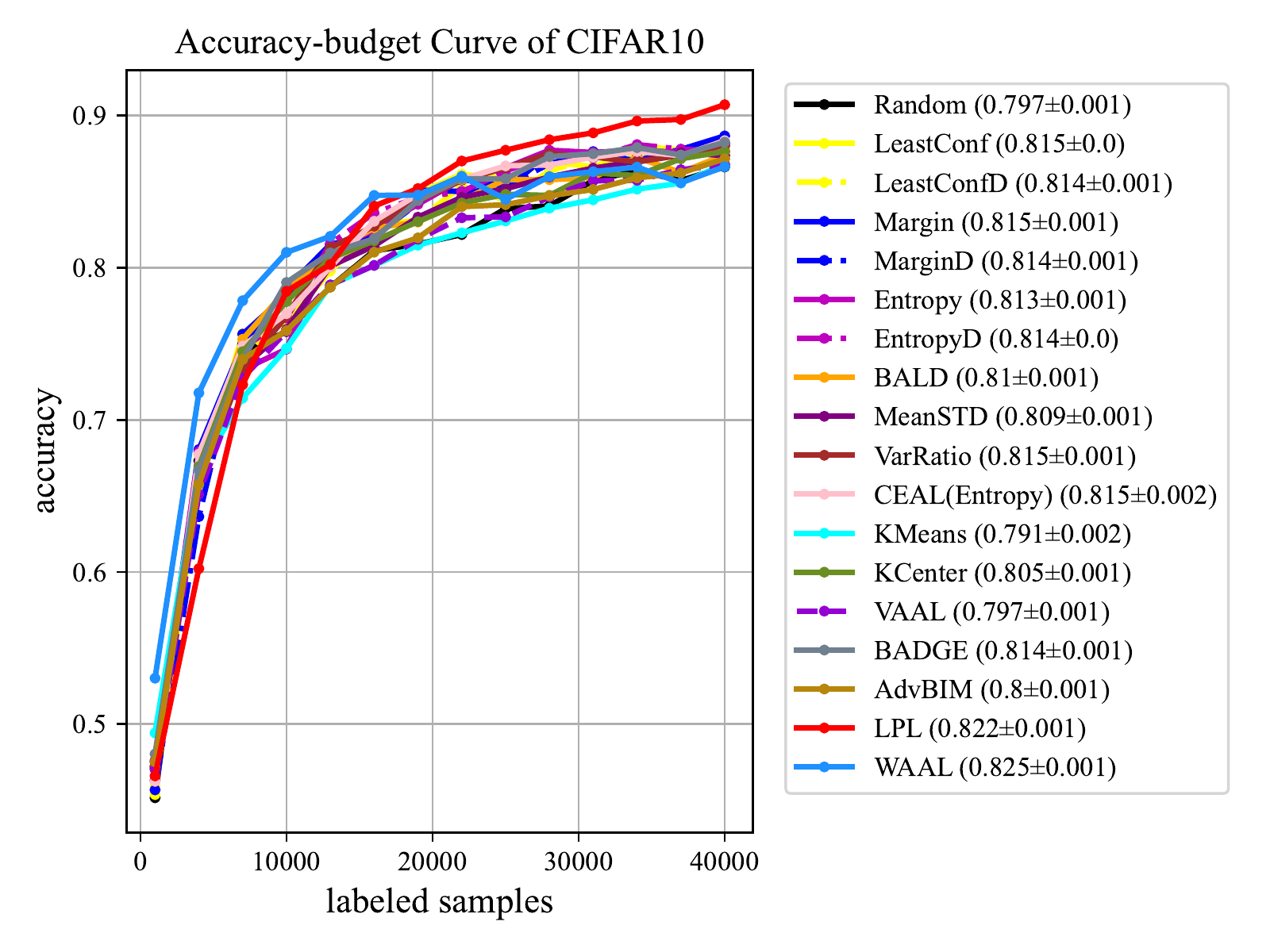}}
\subfloat[CIFAR10-imb]{\includegraphics[width=0.33\linewidth]{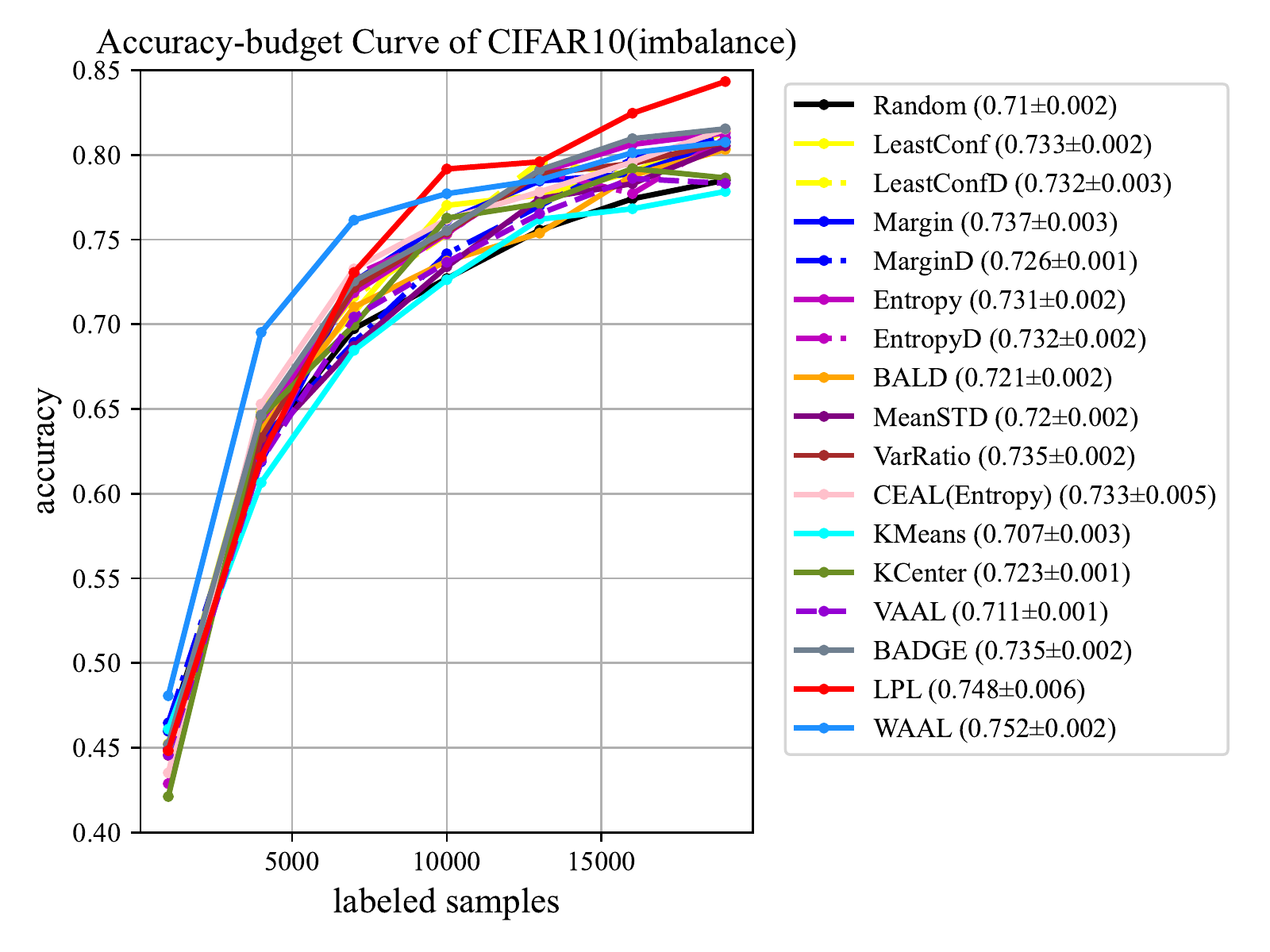}}
\subfloat[CIFAR100]{\includegraphics[width=0.33\linewidth]{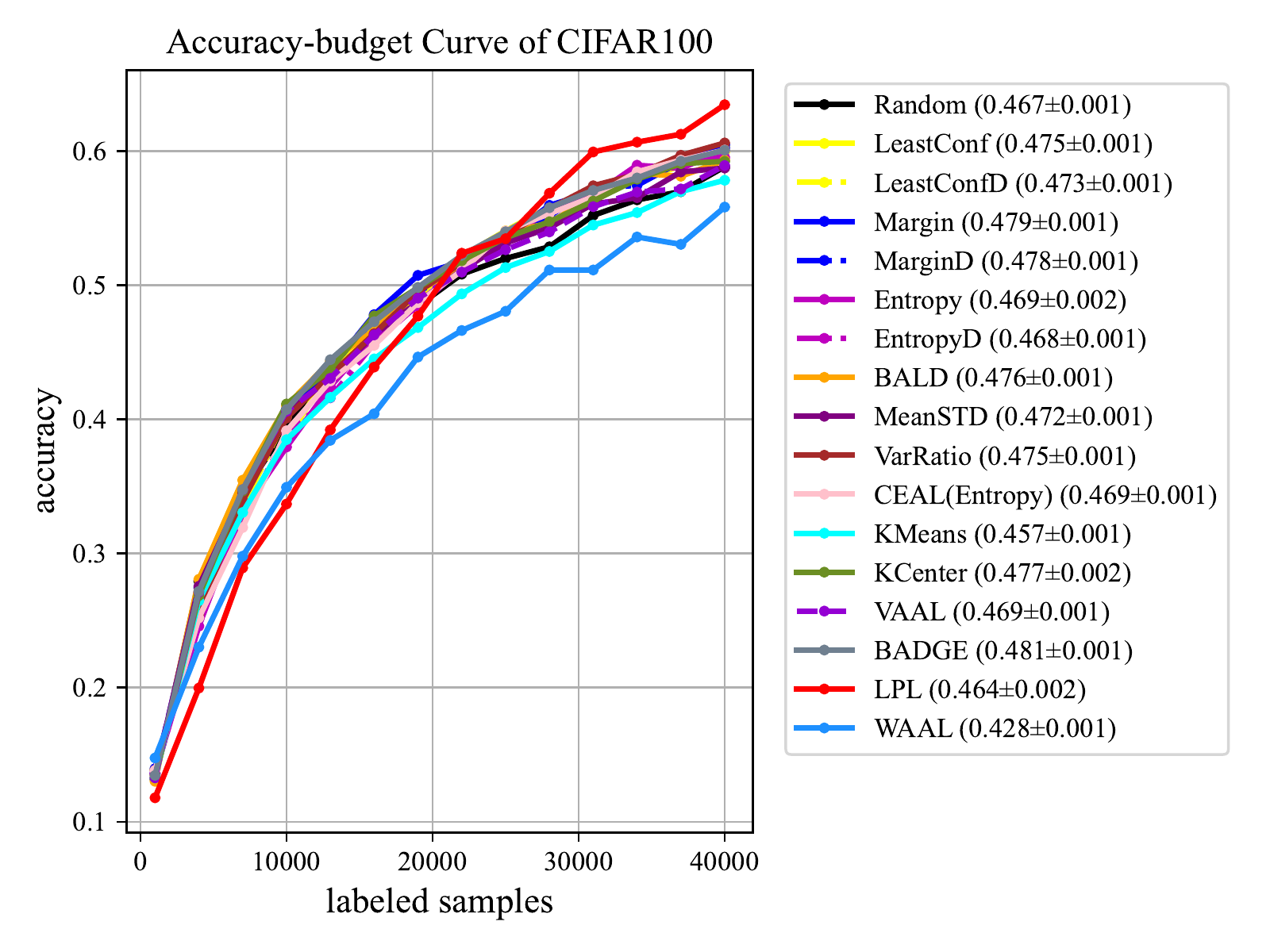}}
\caption{Overall accuracy-budget curve of \emph{MNIST}, \emph{FashionMNIST}, \emph{CIFAR10}, \emph{CIFAR10 (imb)} and \emph{CIFAR100} datasets. The mean and standard deviation of the AUBC (acc) performance over $3$ trials is shown in parentheses in the legend.}
\label{acc-budget}
\end{figure}

To observe the performance differences on various DAL methods in varying AL stages, we provide overall accuracy-budget curves on multiple datasets, as shown in Figure~\ref{acc-budget}. From this figure, it could be observed 
that \textbf{LPL} is weak in the early stage of AL processes due to the inaccurate loss prediction trained on insufficient labeled data. In later stages, by co-training LossNet and the basic classifier on more labeled data, LossNet has demonstrated its ability to enhance the basic classifier. In contrast, \textbf{WAAL} performs better in the early stage of AL processes due to the design of the loss function that is more suitable for AL. It helps distinguish labeled and unlabeled samples and select more representative data samples in the early stage. Therefore, \textbf{WAAL} brings more benefits when limiting the budget for labeling costs.

\subsection{Ablation study: numbers of training epochs and batch size}

\begin{figure}[!htb]
\centering
\subfloat[$b=1000$, $\#e=5$]{\includegraphics[width=0.25\linewidth]{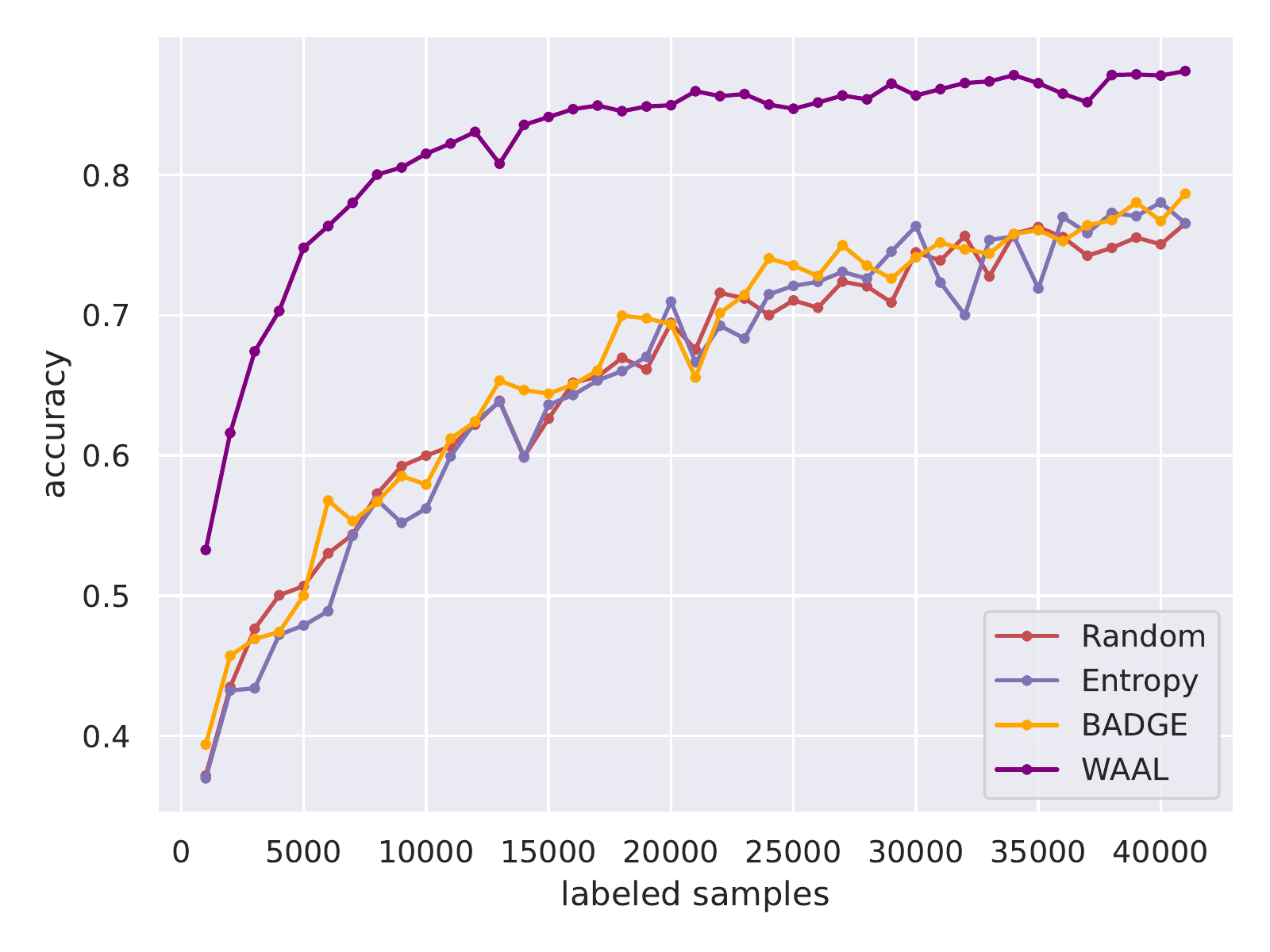}}
\subfloat[$b=2000$, $\#e=5$]{\includegraphics[width=0.25\linewidth]{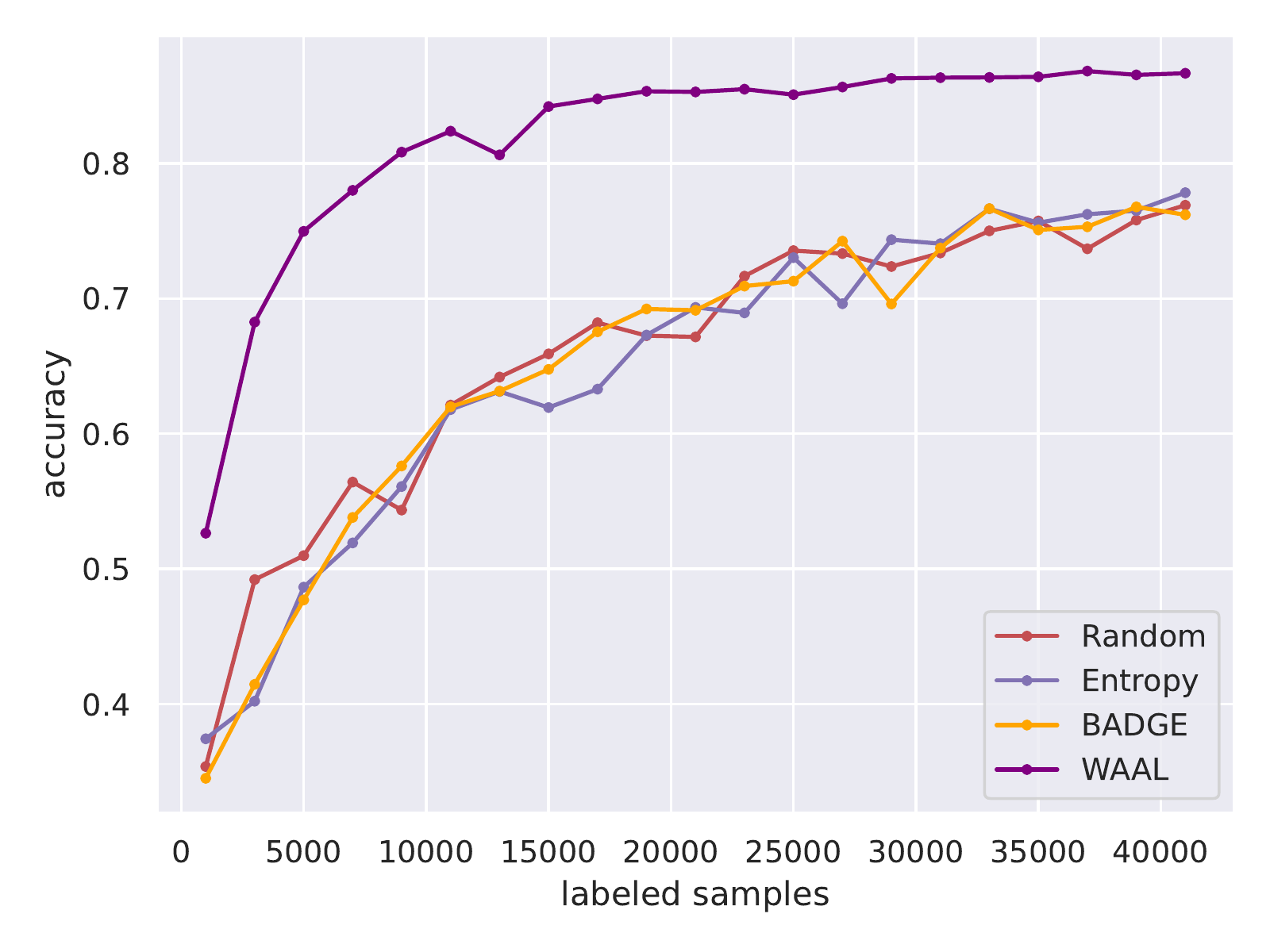}}
\subfloat[$b=4000$, $\#e=5$]{\includegraphics[width=0.25\linewidth]{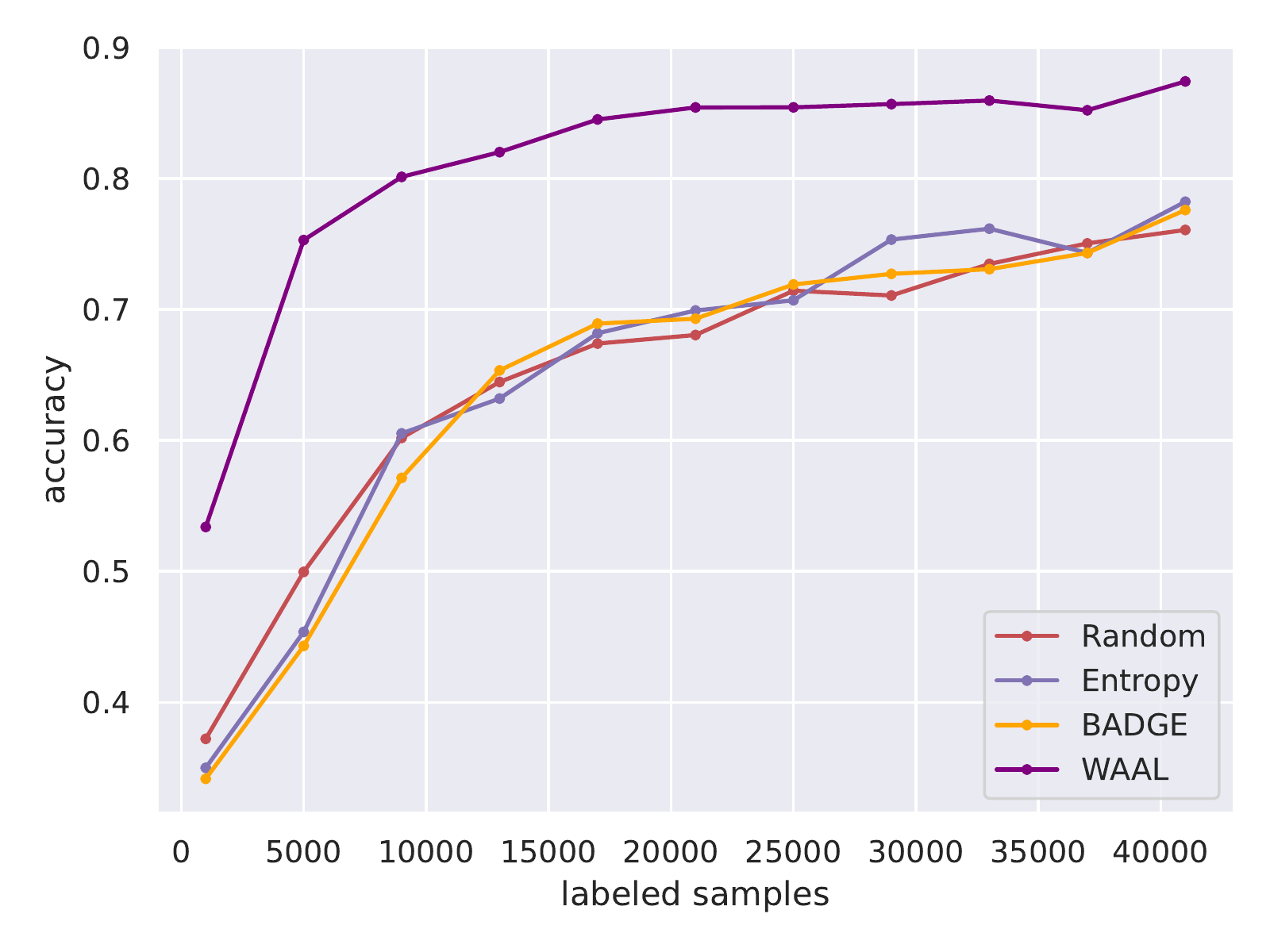}}
\subfloat[$b=10000$, $\#e=5$]{\includegraphics[width=0.25\linewidth]{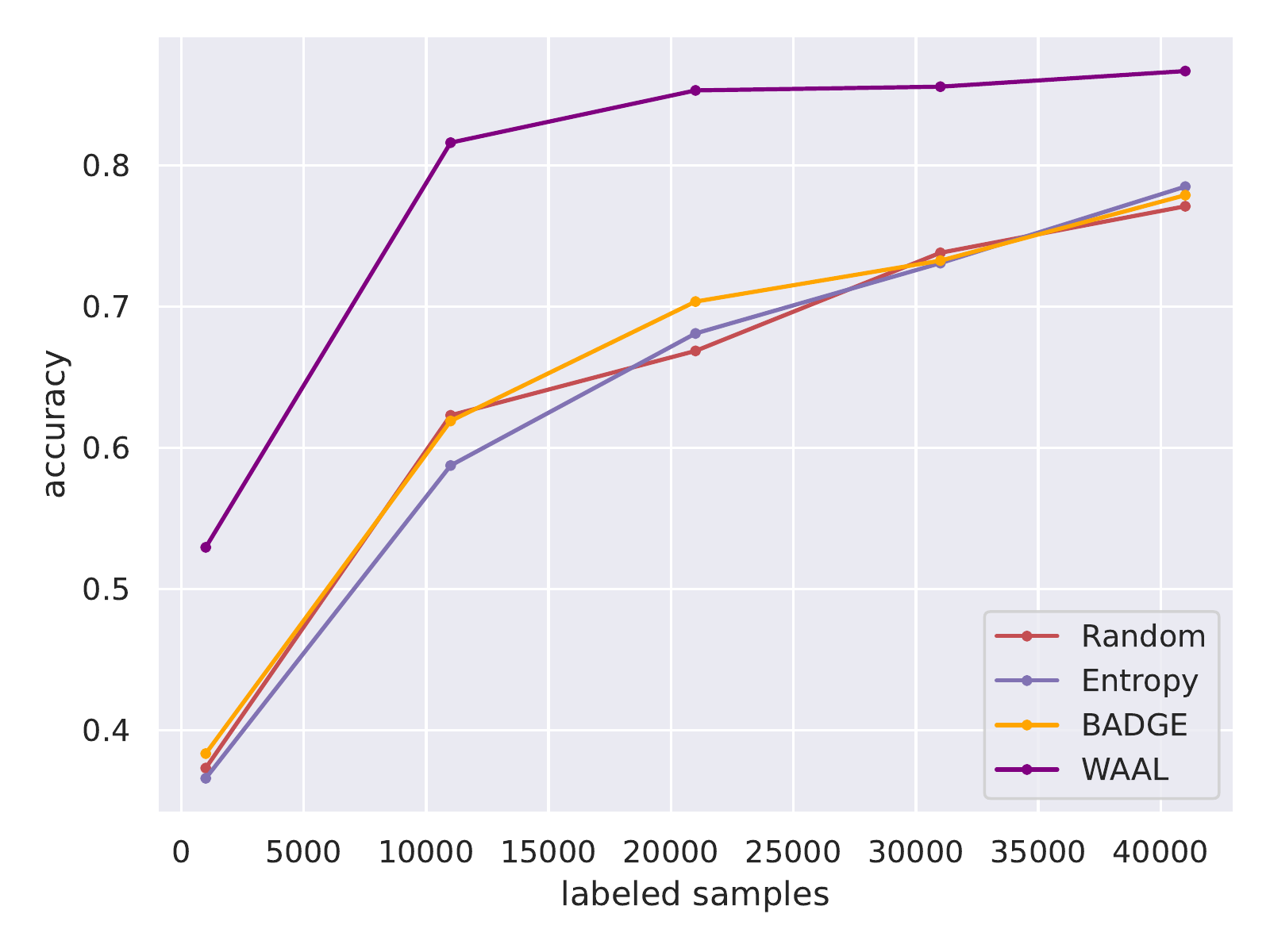}} \\ 
\subfloat[$b=1000$, $\#e=10$]{\includegraphics[width=0.25\linewidth]{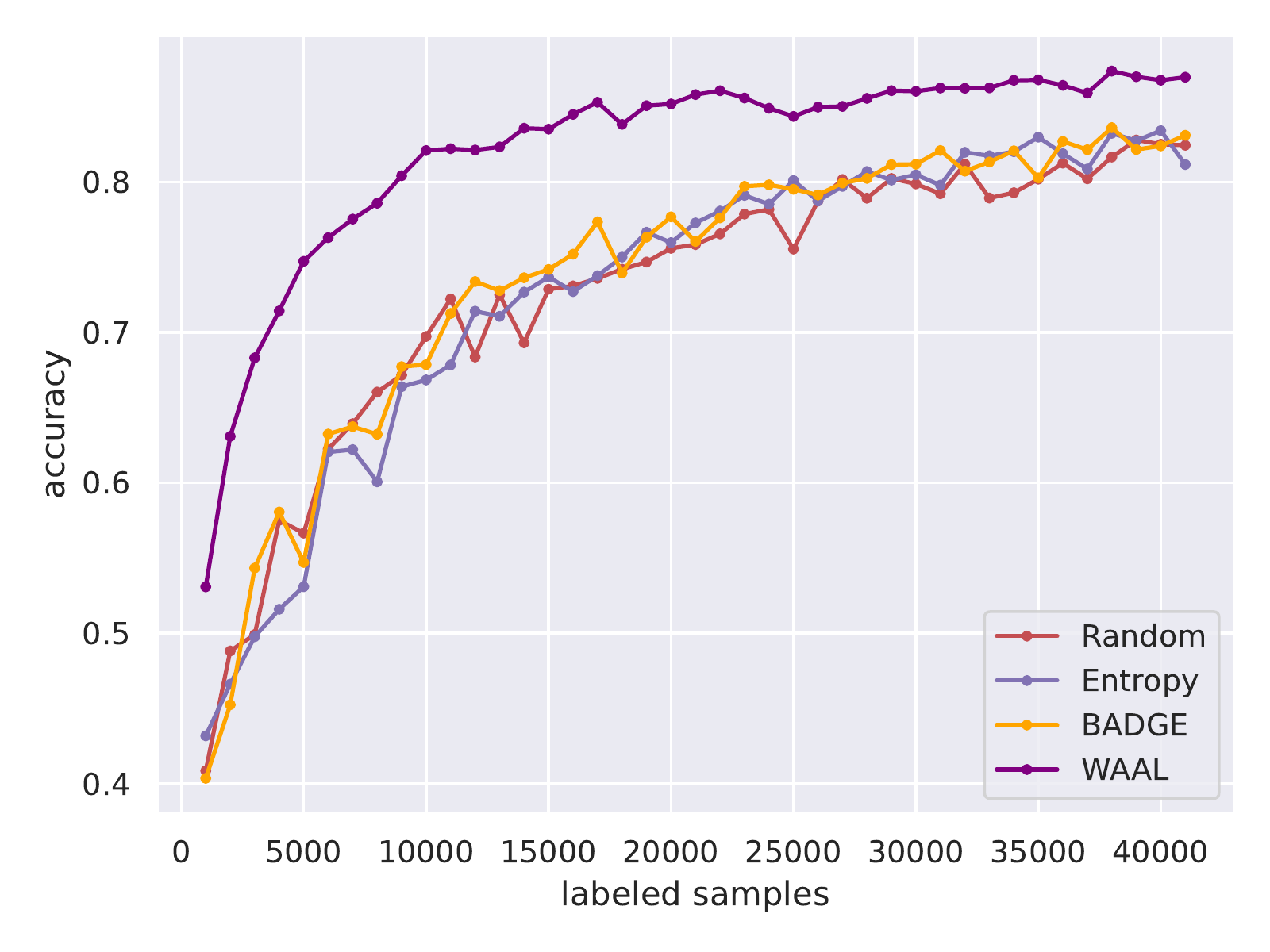}}
\subfloat[$b=2000$, $\#e=10$]{\includegraphics[width=0.25\linewidth]{img/exp/2000-5.pdf}}
\subfloat[$b=4000$, $\#e=10$]{\includegraphics[width=0.25\linewidth]{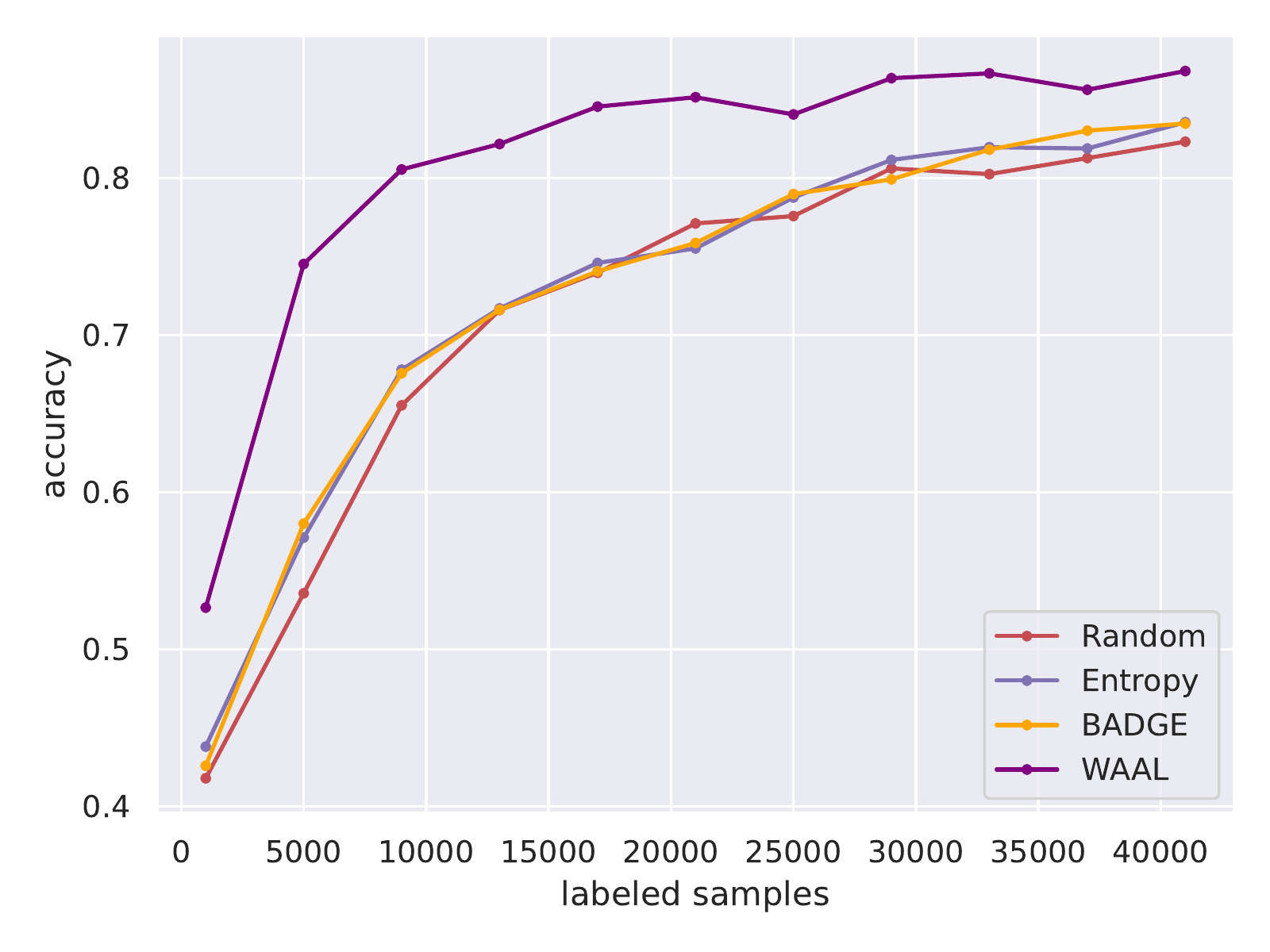}}
\subfloat[$b=10000$, $\#e=10$]{\includegraphics[width=0.25\linewidth]{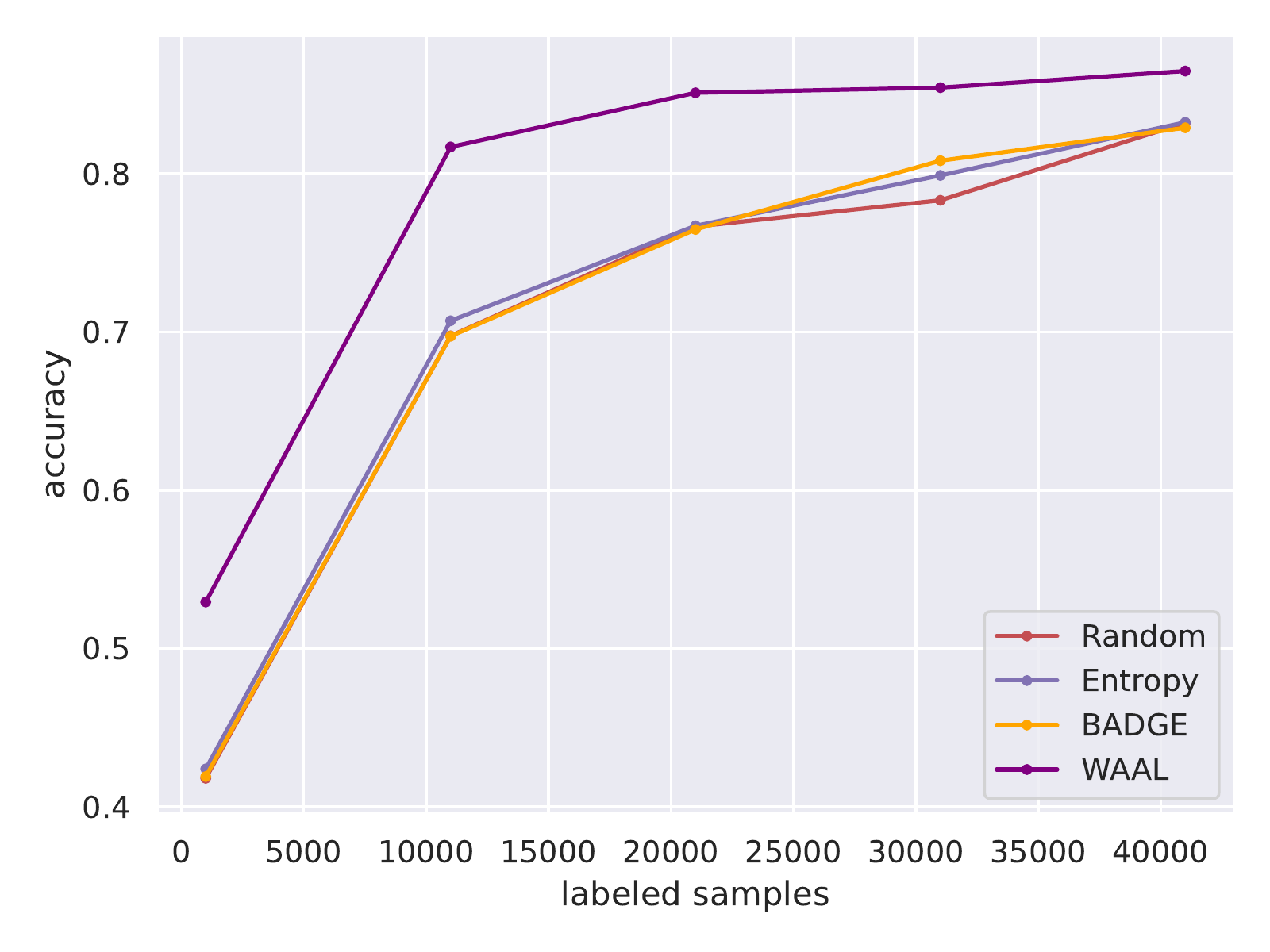}}\\
\subfloat[$b=1000$, $\#e=15$]{\includegraphics[width=0.25\linewidth]{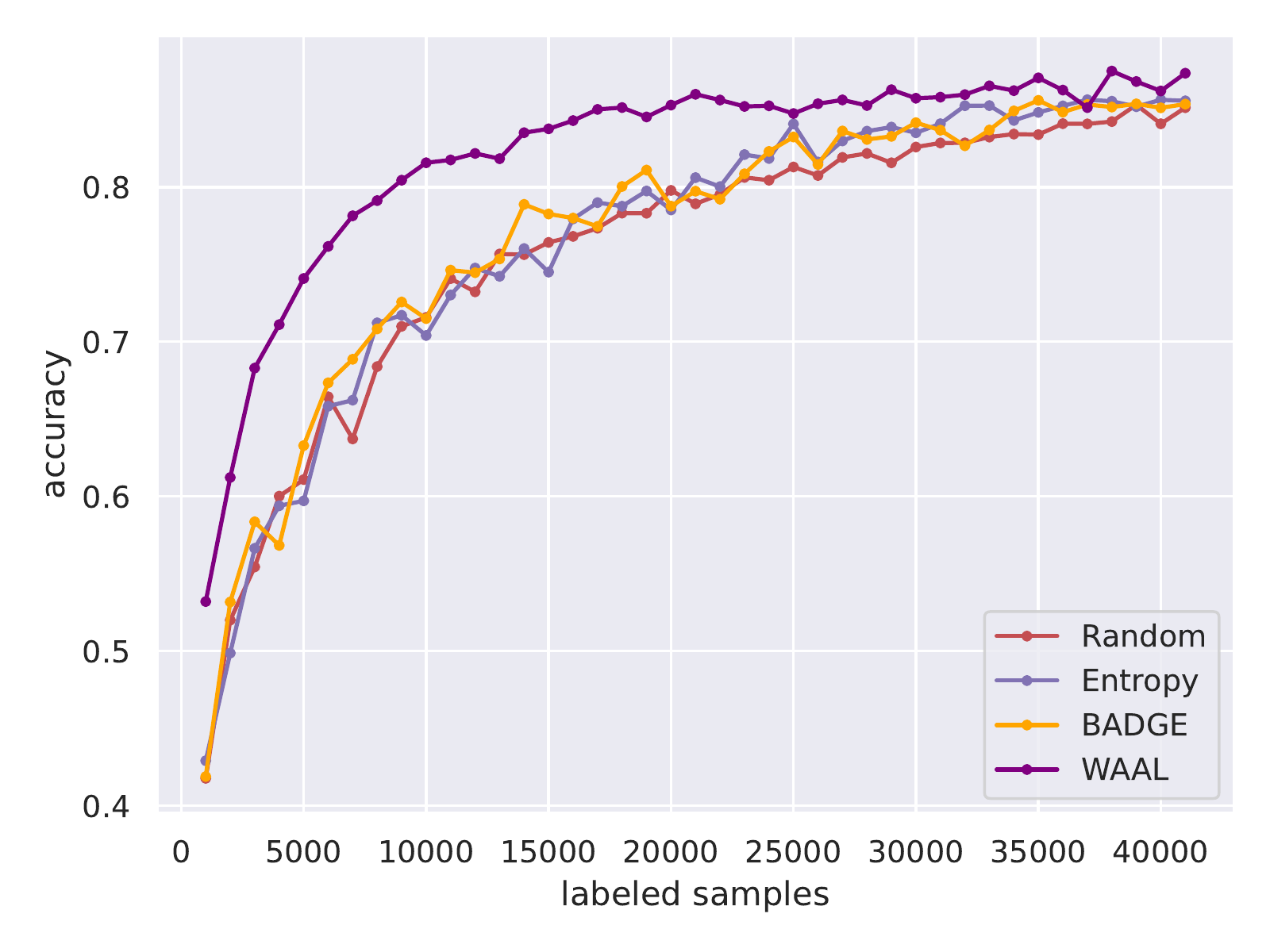}}
\subfloat[$b=2000$, $\#e=15$]{\includegraphics[width=0.25\linewidth]{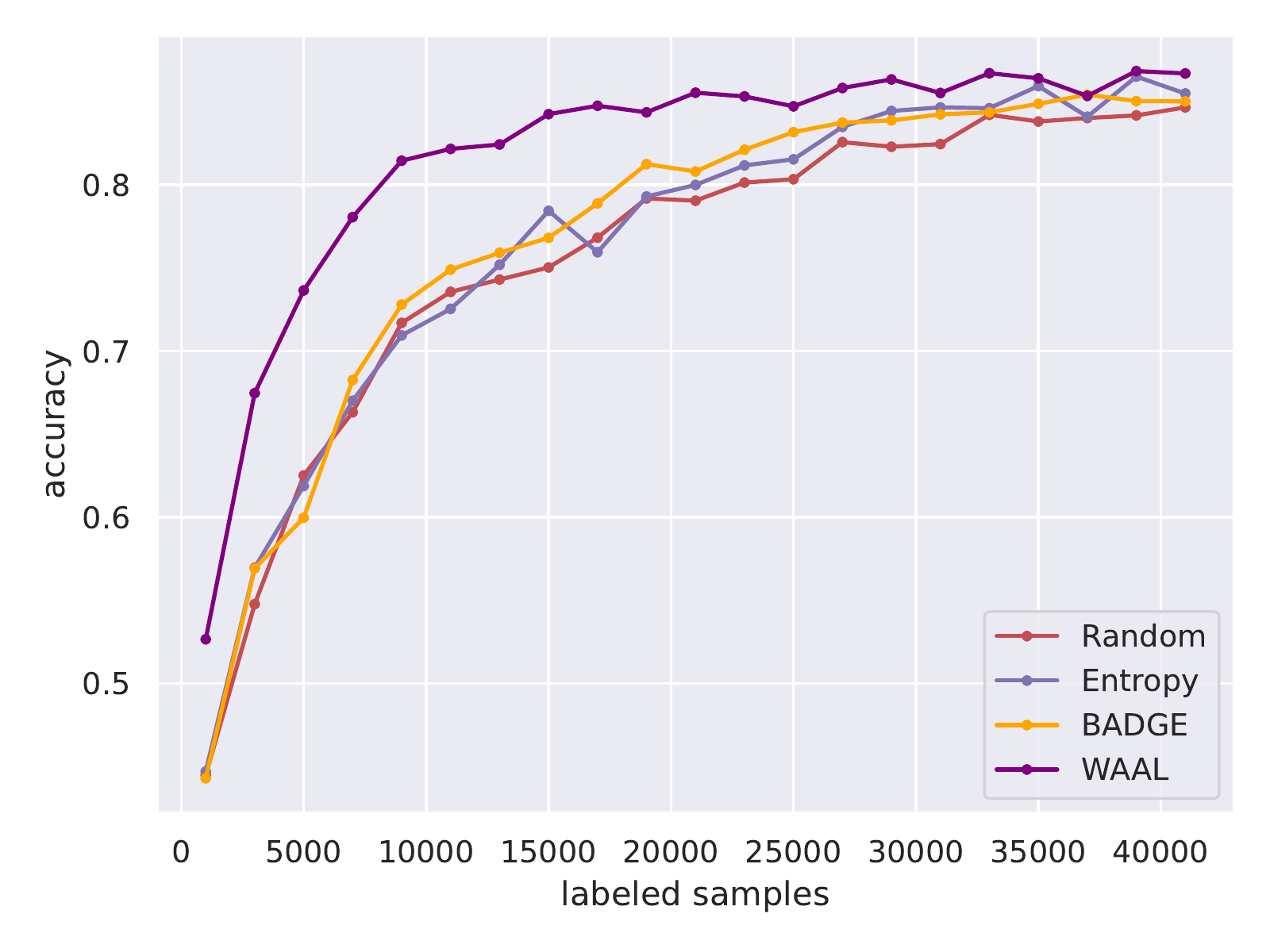}}
\subfloat[$b=4000$, $\#e=15$]{\includegraphics[width=0.25\linewidth]{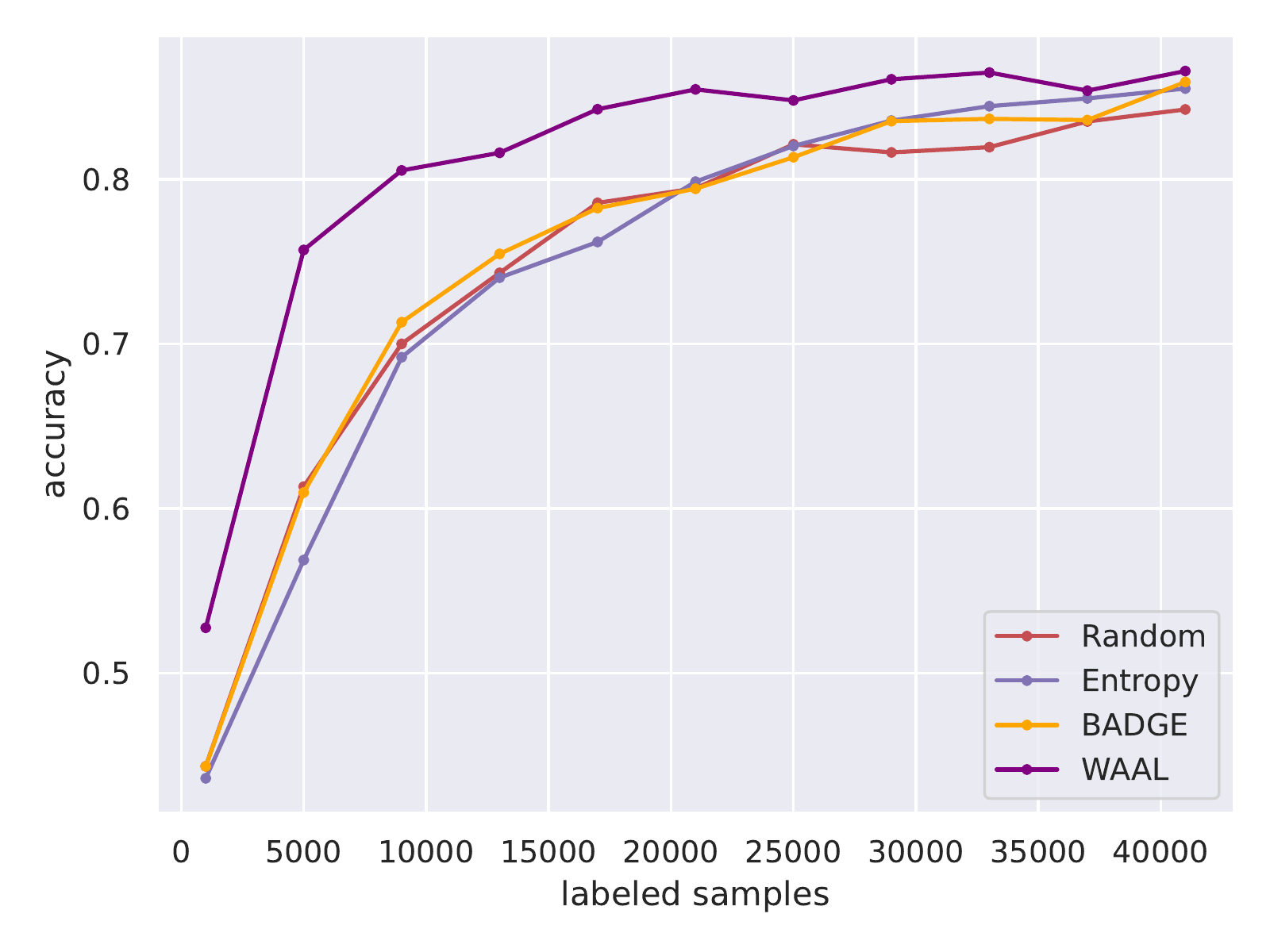}}
\subfloat[$b=10000$, $\#e=15$]{\includegraphics[width=0.25\linewidth]{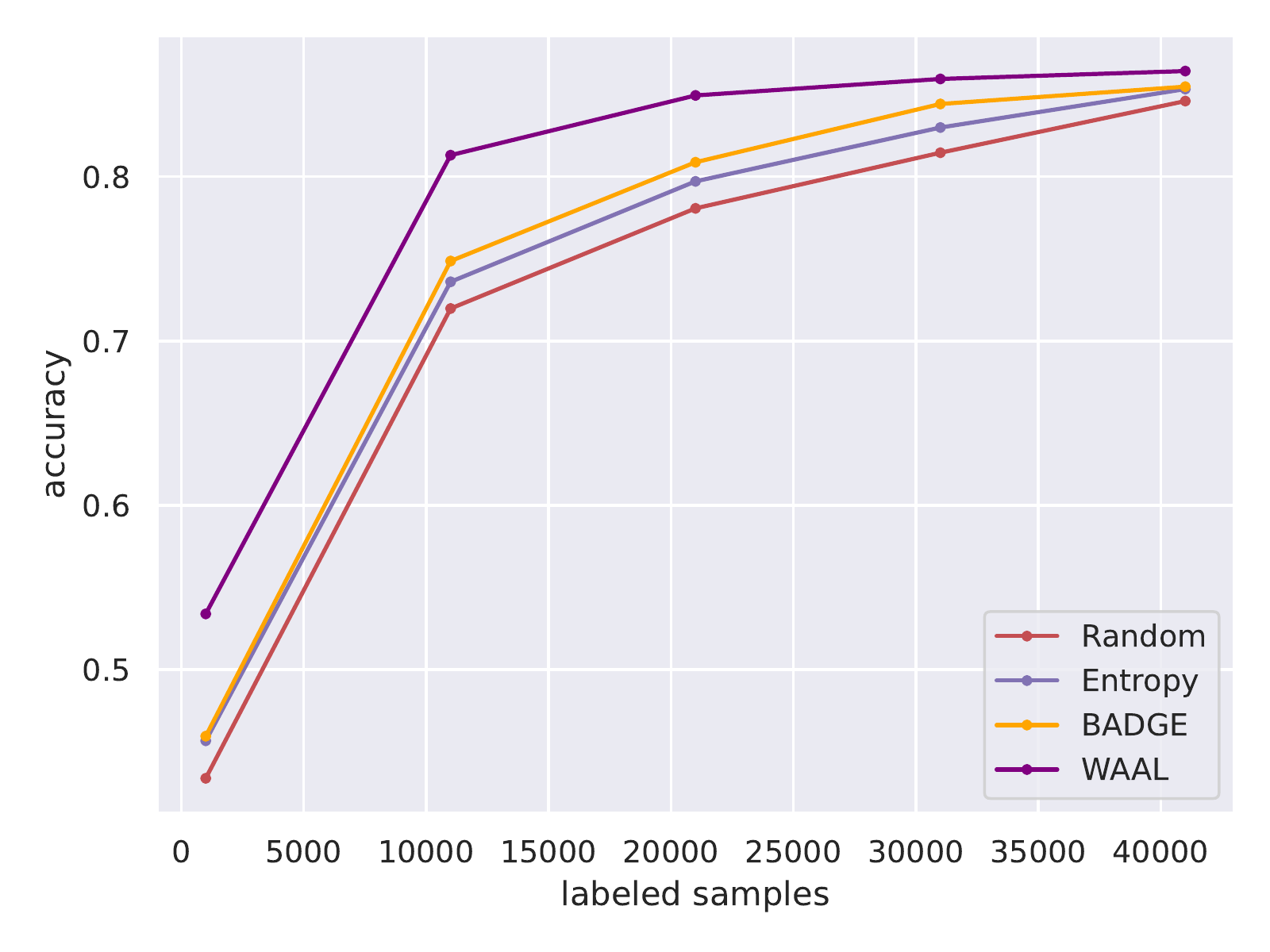}}\\
\subfloat[$b=1000$, $\#e=20$]{\includegraphics[width=0.25\linewidth]{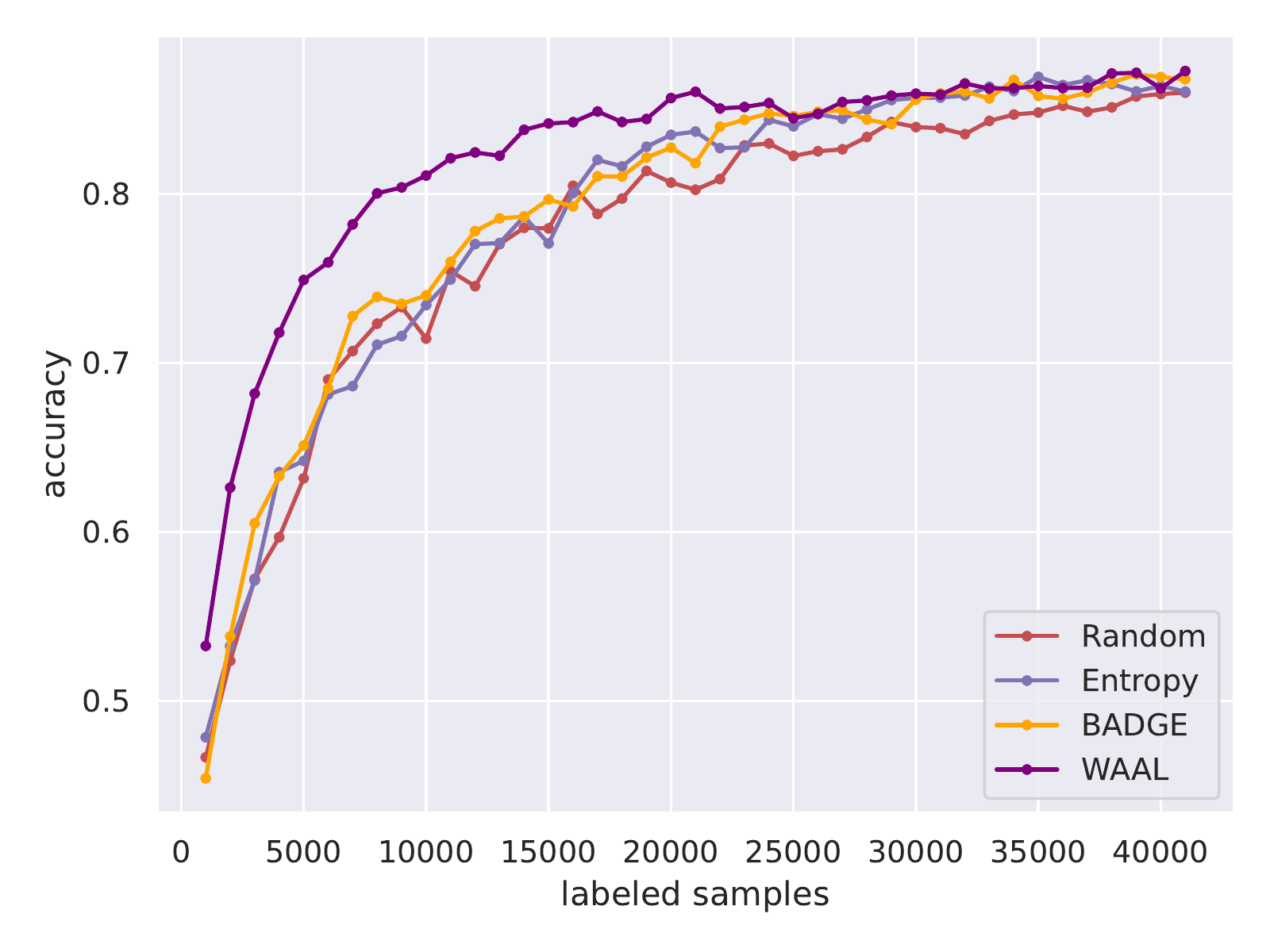}}
\subfloat[$b=2000$, $\#e=20$]{\includegraphics[width=0.25\linewidth]{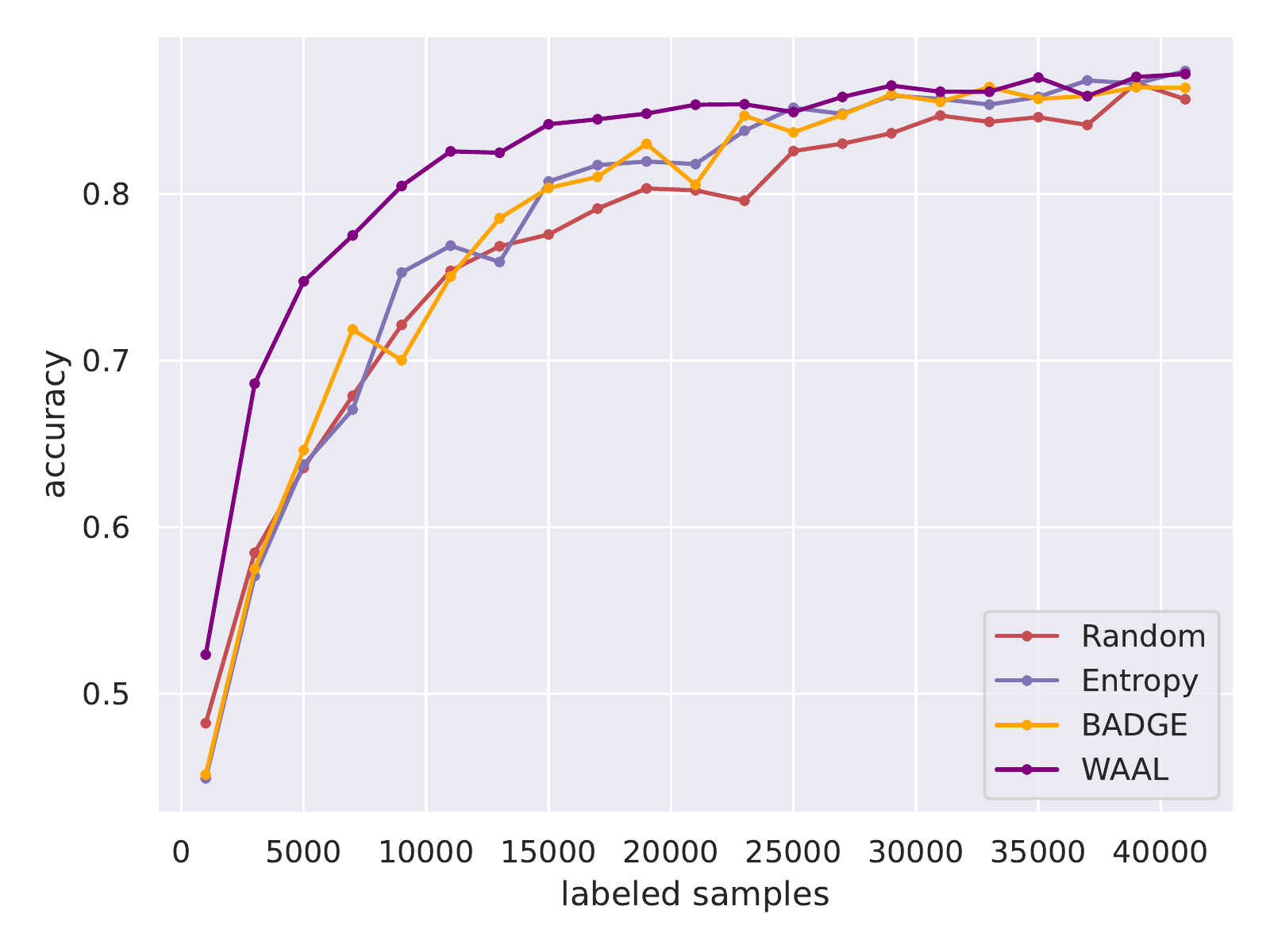}}
\subfloat[$b=4000$, $\#e=20$]{\includegraphics[width=0.25\linewidth]{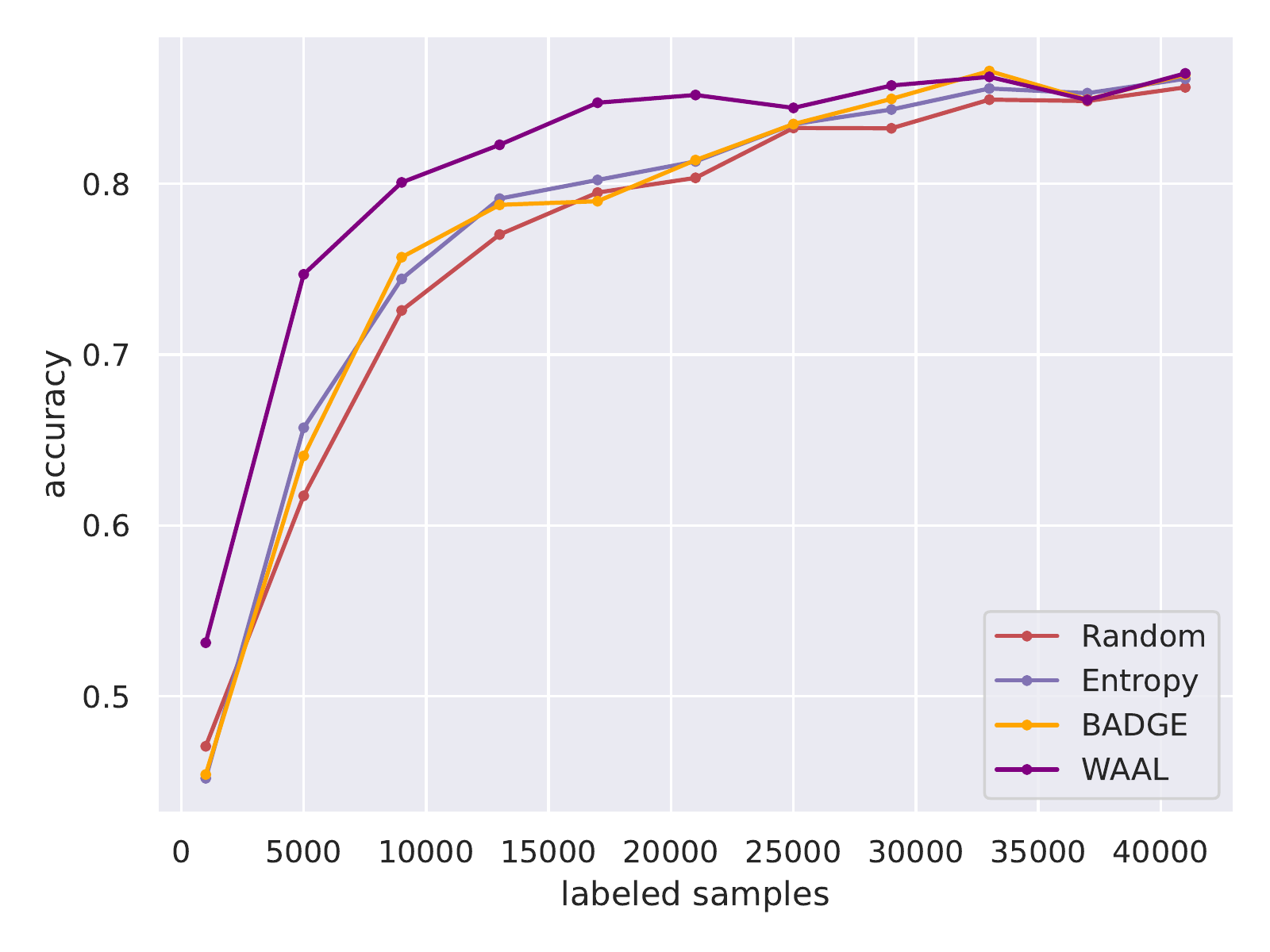}}
\subfloat[$b=10000$, $\#e=20$]{\includegraphics[width=0.25\linewidth]{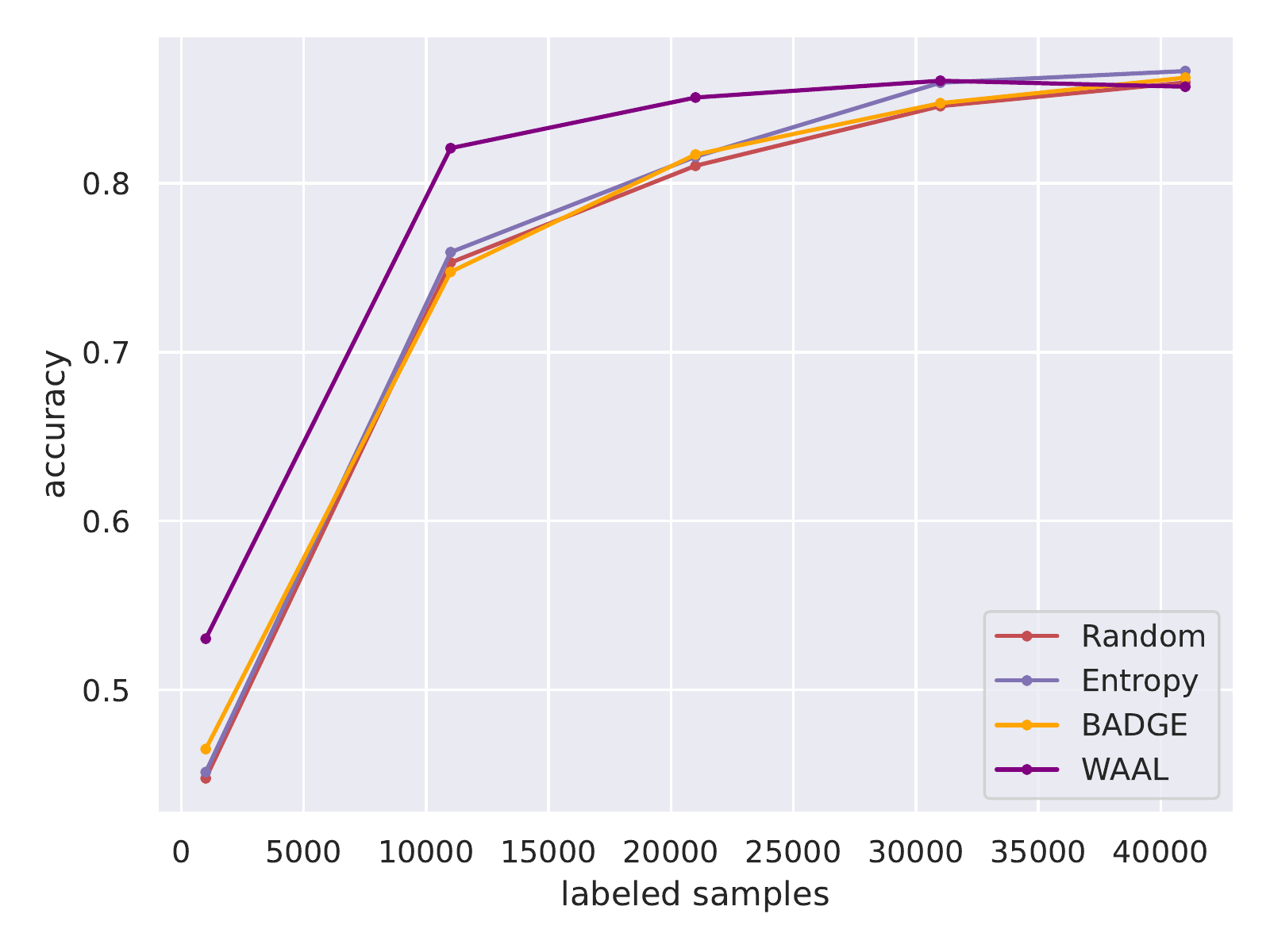}}\\
\subfloat[$b=1000$, $\#e=25$]{\includegraphics[width=0.25\linewidth]{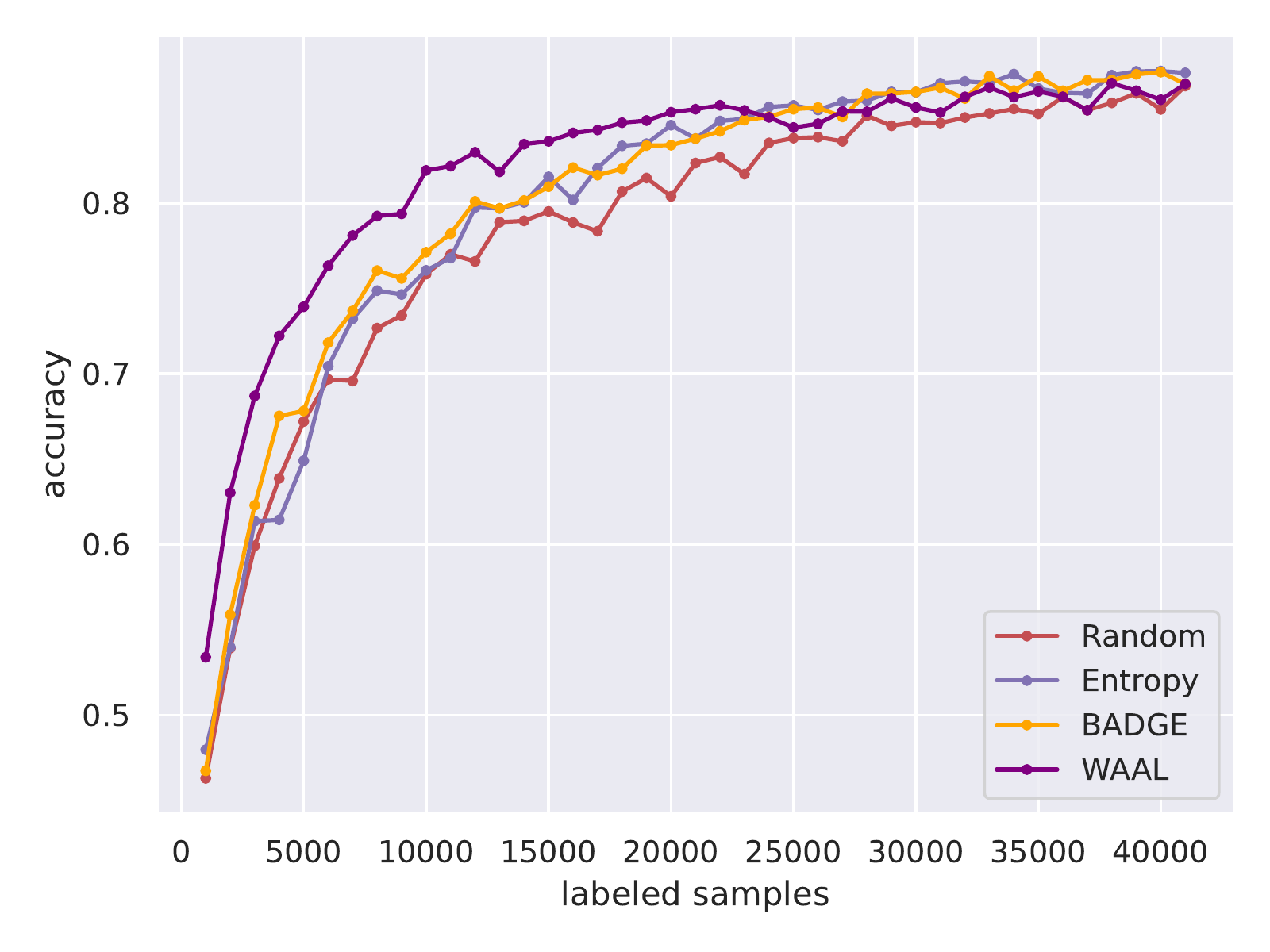}}
\subfloat[$b=2000$, $\#e=25$]{\includegraphics[width=0.25\linewidth]{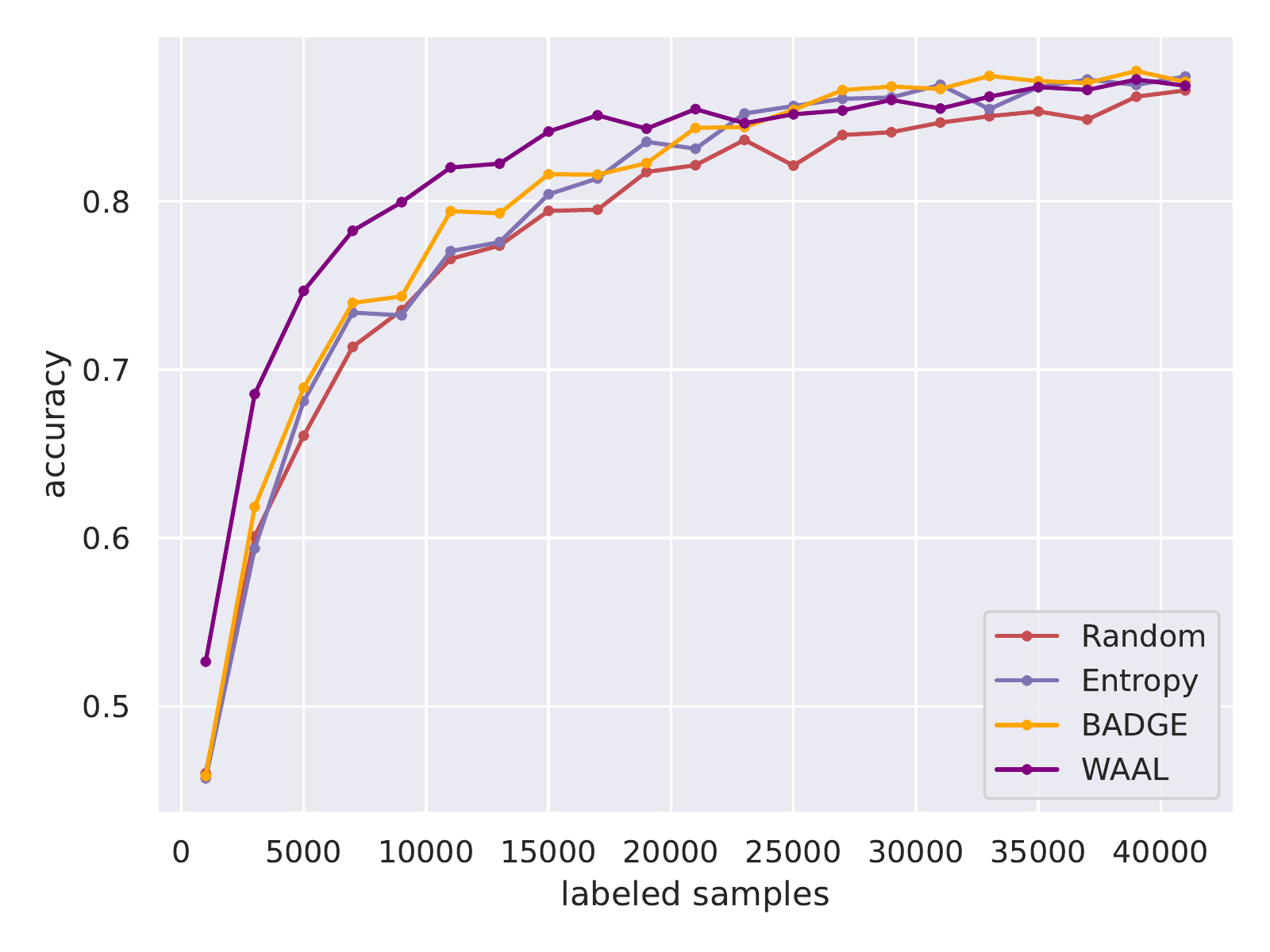}}
\subfloat[$b=4000$, $\#e=25$]{\includegraphics[width=0.25\linewidth]{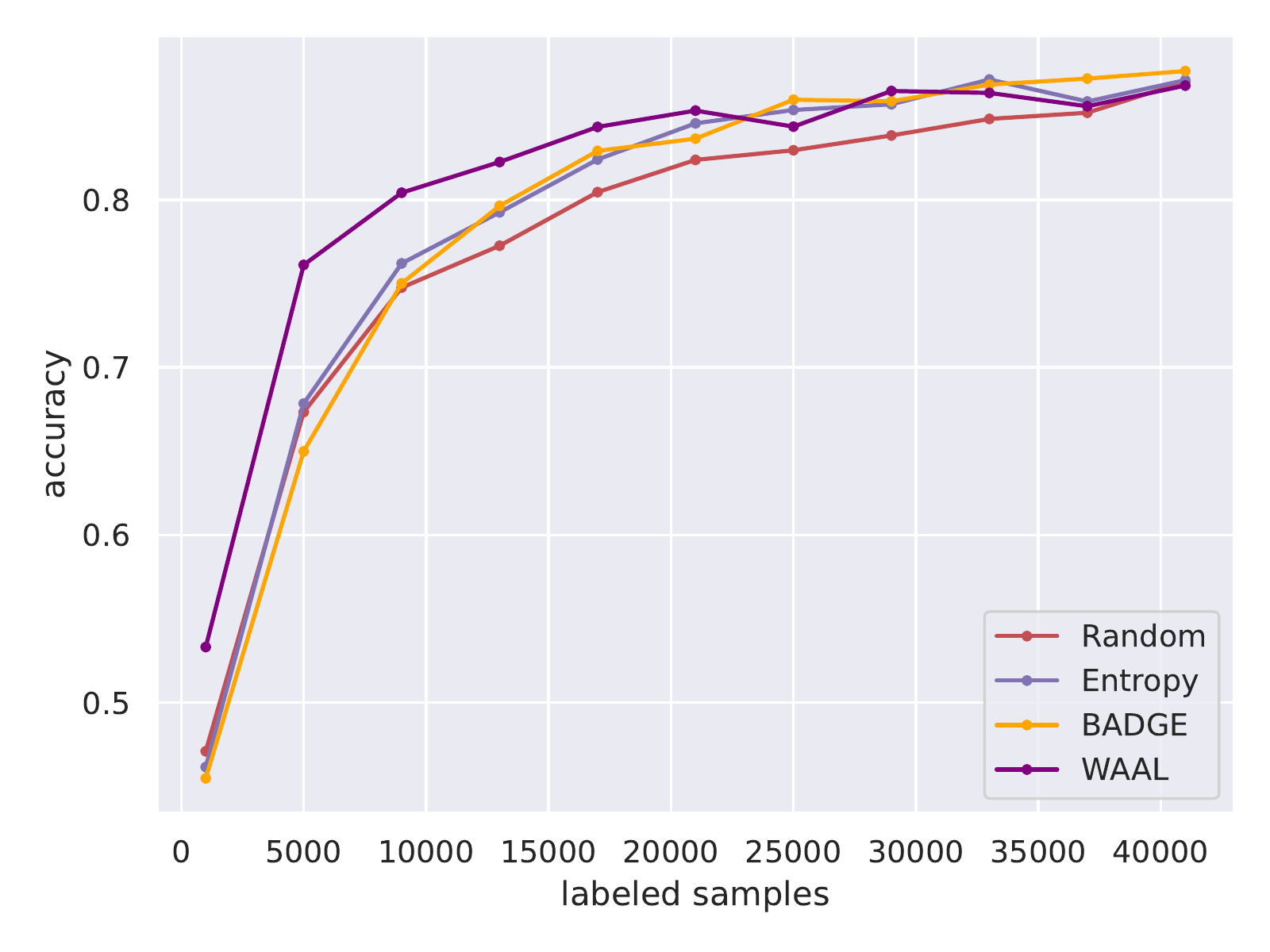}}
\subfloat[$b=10000$, $\#e=25$]{\includegraphics[width=0.25\linewidth]{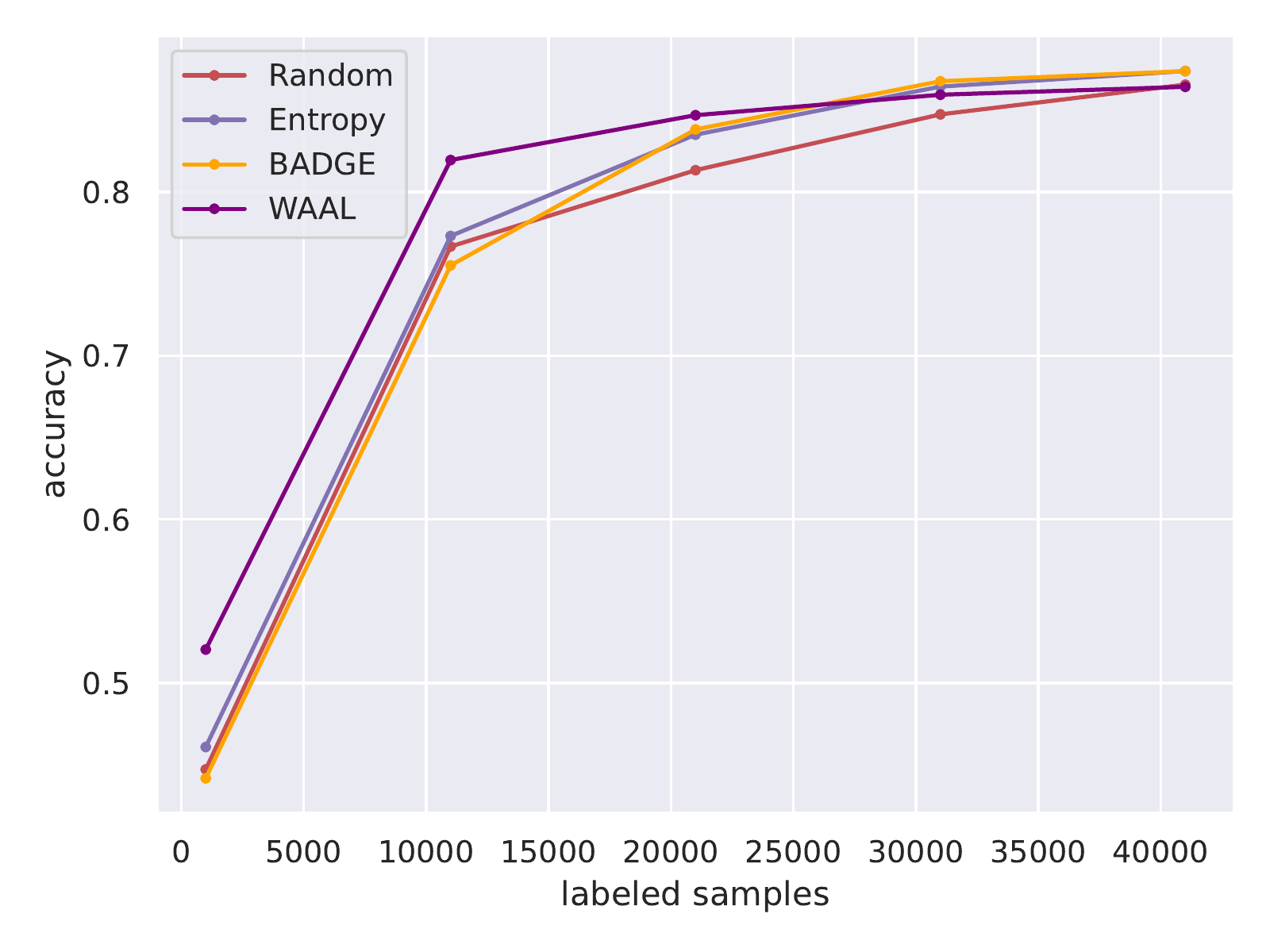}}\\
\subfloat[$b=1000$, $\#e=30$]{\includegraphics[width=0.25\linewidth]{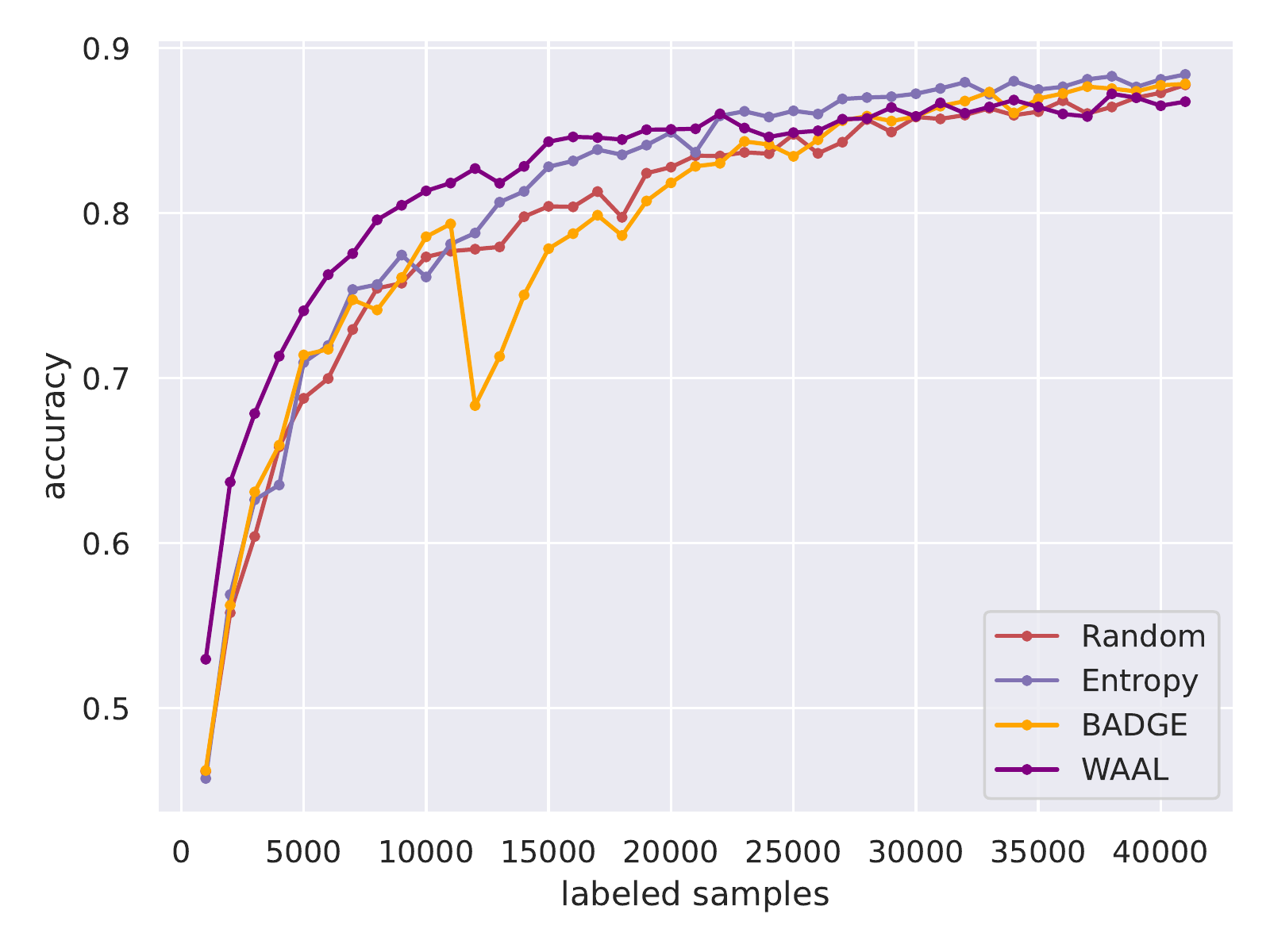}}
\subfloat[$b=2000$, $\#e=30$]{\includegraphics[width=0.25\linewidth]{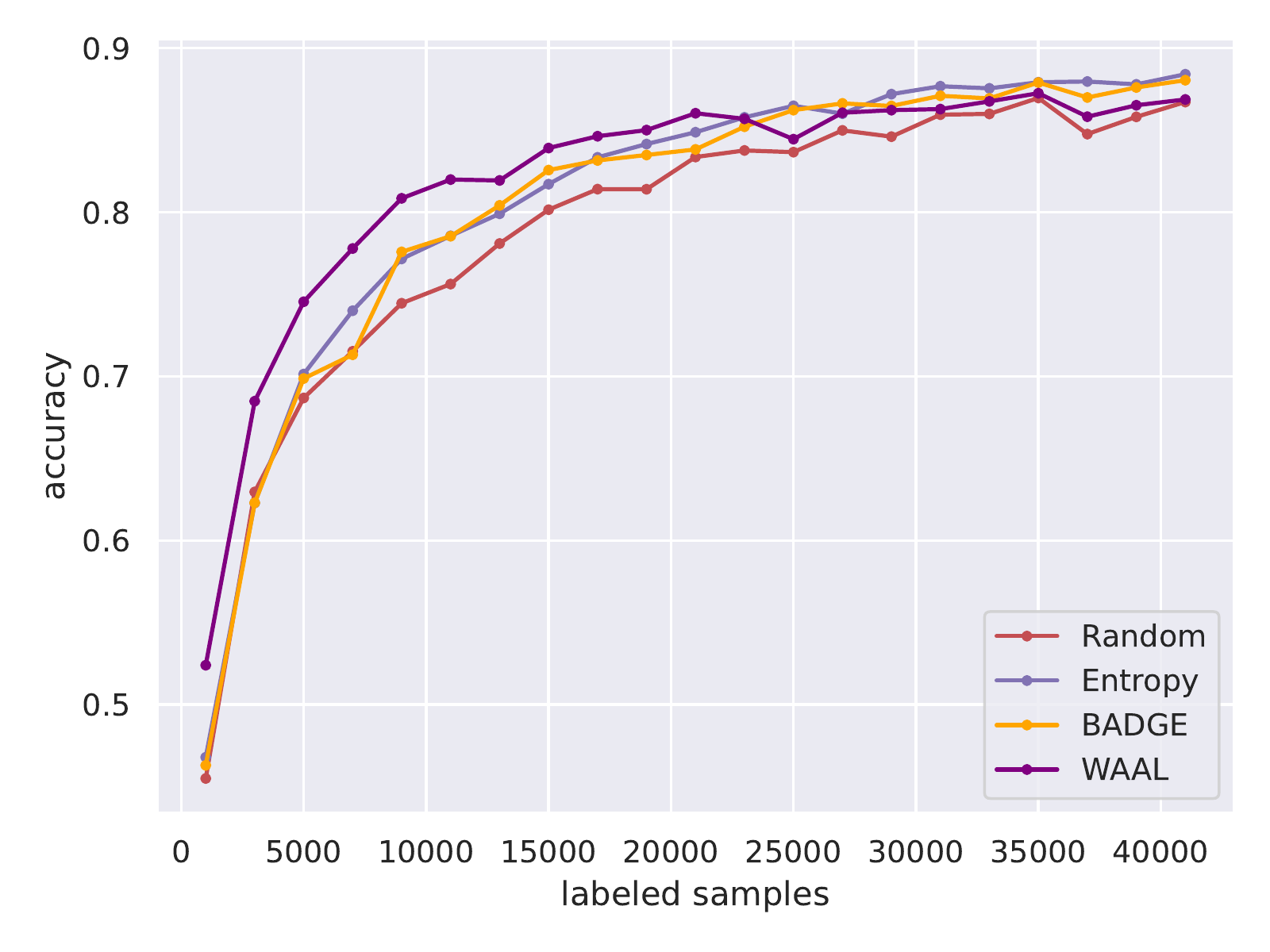}}
\subfloat[$b=4000$, $\#e=30$]{\includegraphics[width=0.25\linewidth]{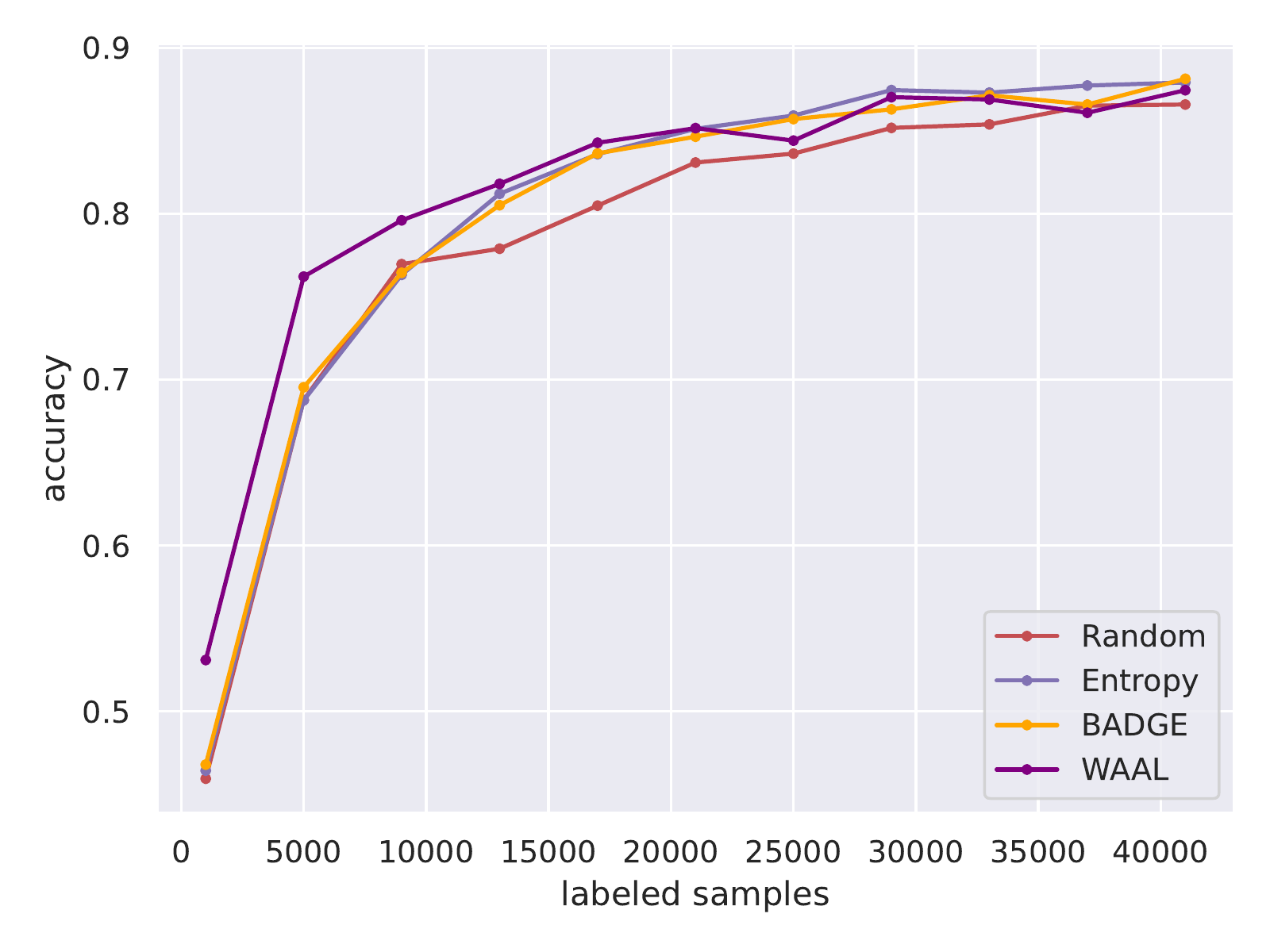}}
\subfloat[$b=10000$, $\#e=30$]{\includegraphics[width=0.25\linewidth]{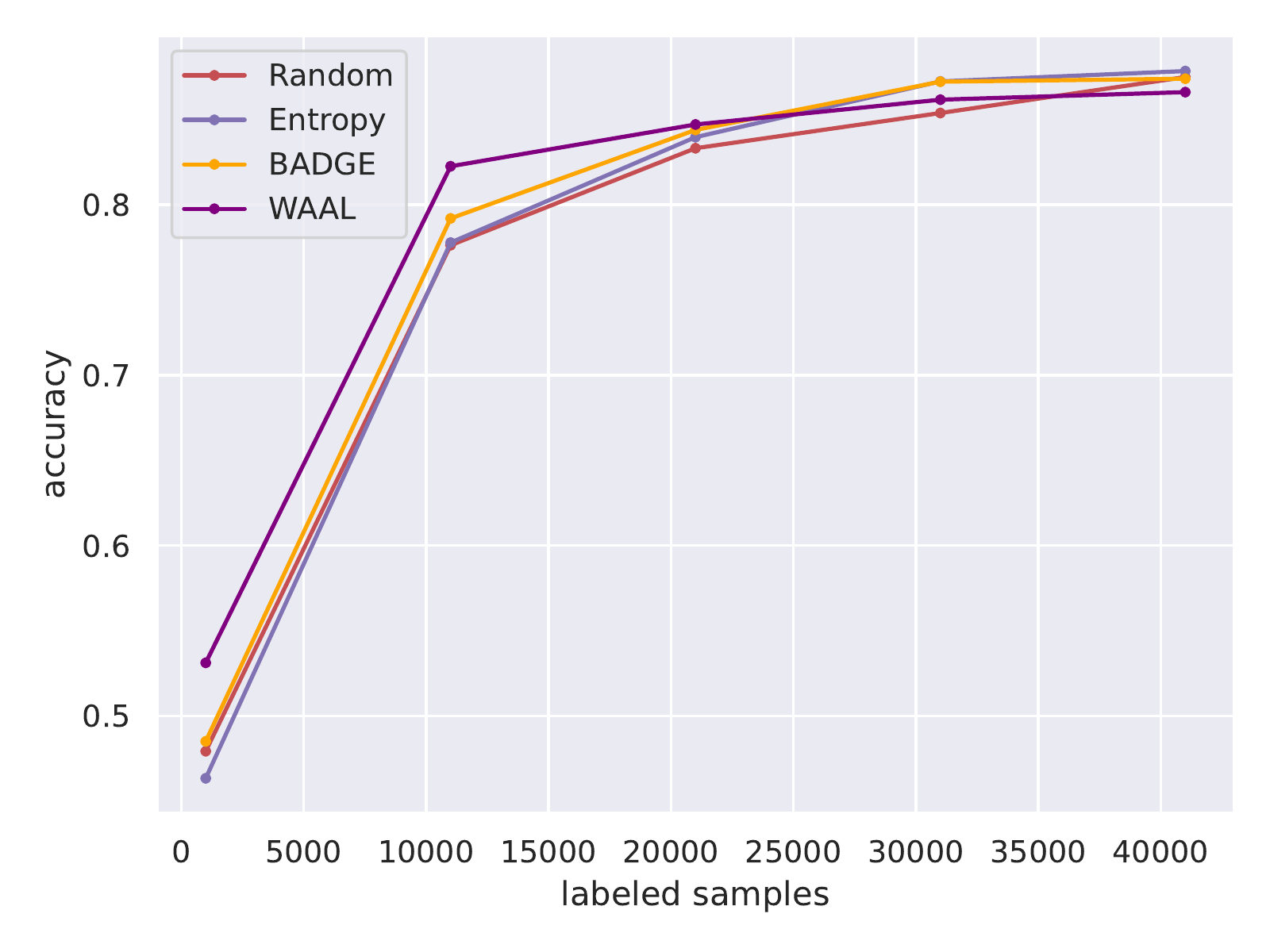}}
\caption{The accuracy-budget curve of different DAL methods on \emph{CIFAR10} dataset with different batch sizes and numbers of training epochs in each active learning round. From left to right, the value of $b$ equals to $1,000$, $2,000$, $4,000$ and $10,000$. From top to bottom, the value of training epoch equals to $5$, $10$, $15$, $20$, $25$ and $30$.}\label{fig:curves}
\end{figure}

We present the accuracy-budget curves using different batch sizes and training epochs, as shown in Figure~\ref{fig:curves}. A detailed analysis of this ablation study is in the main paper.

\subsection{Ablation study: w/ and w/o pre-training techniques}
The detailed results of \textbf{MNIST}, \textbf{Waterbird} w/ and w/o pre-training techniques are shown in Tables~\ref{performance-1} and \ref{performance-5}, including the overall accuracy, mismatch group accuracy and worst group accuracy. We record AUBC (acc) with mean and standard deviation over 3 trials, F-acc and average running time. We have detailed analysed this experiment in main paper. From Tables~\ref{performance-1} and \ref{performance-5}, we could observe that on \emph{Waterbird} w/o pre-train, most typical DAL sampling methods like \textbf{LeastConf}, \textbf{Margin}, \textbf{Entropy}, \textbf{KCenter}, etc., perform even worse than \textbf{Random}.

\end{document}